\documentclass[10pt,twocolumn,letterpaper]{article}

\usepackage{iccv}
\usepackage{times}
\usepackage{epsfig}
\usepackage{graphicx}
\usepackage{amsmath}
\usepackage{amssymb}
\usepackage{graphicx}
\usepackage{bbm}
\usepackage{amsmath}
\usepackage{amssymb}
\usepackage{amsthm}
\usepackage{float}
\usepackage{subcaption}
\usepackage{csquotes}
\usepackage[final]{listings}
\usepackage{color, soul}
\usepackage{xcolor}
\usepackage{microtype}
\usepackage{booktabs} 
\usepackage[shortlabels]{enumitem}

\lstdefinestyle{mystyle}{
    backgroundcolor=\color{backcolour},   
    commentstyle=\color{codegreen},
    keywordstyle=\color{codeblue},
    numberstyle=\tiny\color{codegray},
    stringstyle=\color{codeturquoise},
    basicstyle=\footnotesize\ttfamily,
    breakatwhitespace=false,         
    breaklines=true,                 
    captionpos=b,                    
    keepspaces=true,                 
    numbers=left,                    
    numbersep=5pt,                  
    showspaces=false,                
    showstringspaces=false,
    showtabs=false,                  
    tabsize=2
}
\newtheorem{proposition}{Claim}

\usepackage[toc,page]{appendix}

\definecolor{codeblue}{rgb}{0,0,0.6}
\definecolor{codegreen}{rgb}{0,0.6,0}
\definecolor{codegray}{rgb}{0.5,0.5,0.5}
\definecolor{codeturquoise}{rgb}{0.0,0.5,0.5}
\definecolor{backcolour}{rgb}{0.98,0.98,0.98}

\usepackage[pagebackref=true,breaklinks=true,letterpaper=true,colorlinks,bookmarks=false]{hyperref}
\iccvfinalcopy 

\def\iccvPaperID{****} 

\begin{document}

\title{Why are Saliency Maps Noisy? Cause of and Solution to Noisy Saliency Maps}

\author{
    \begin{tabular}[t]{c@{\extracolsep{4.0em}}c@{\extracolsep{4.0em}}c}
      Beomsu Kim$^{1}$ & Junghoon Seo$^{2}$ & Seunghyeon Jeon$^{2}$ \\ Jamyoung Koo$^{3}$ &  Jeongyeol Choe$^{3}$ & Taegyun Jeon$^{3}$
    \end{tabular}
  \\[12pt]
    \begin{tabular}{c@{\extracolsep{2.0em}}c@{\extracolsep{2.0em}}c}
      $^1$Department of Mathematical Sciences, KAIST   &  $^2$Satrec Initiative & $^3$SI Analytics  \\
      \multicolumn{3}{c}{ \texttt{\small 3141kbs@kaist.ac.kr} \hspace{1.5em} \texttt{\small \{sjh,jsh\}@satreci.com} \hspace{1.5em} \texttt{\small \{jmkoo,jychoe,tgjeon\}@si-analytics.ai}} 
    \end{tabular}
}

\maketitle

\begin{abstract}
Saliency Map, the gradient of the score function with respect to the input, is the most basic technique for interpreting deep neural network decisions. However, saliency maps are often visually noisy. Although several hypotheses were proposed to account for this phenomenon, there are few works that provide rigorous analyses of noisy saliency maps. In this paper, we firstly propose a new hypothesis that noise may occur in saliency maps when irrelevant features pass through ReLU activation functions. Then, we propose \textit{Rectified Gradient}, a method that alleviates this problem through layer-wise thresholding during backpropagation. Experiments with neural networks trained on CIFAR-10 and ImageNet showed effectiveness of our method and its superiority to other attribution methods.
\end{abstract}

\section{Introduction}

Saliency Map \cite{Erhan2009,Baehrens2009,Simonyan2013}, the gradient of the score function with respect to the input, is the most basic technique for interpreting deep neural networks (DNNs). It is also a baseline for advanced attribution methods. However, previous studies such as \cite{Springenberg2014} and \cite{Selvaraju2016} have noted that saliency maps are visually noisy.
To explain this phenomenon, \cite{Sundararajan2016} and \cite{Smilkov2017} suggested saturation and discontinuous gradients as the causes. There were several attribution methods attempting to improve saliency maps by tackling these hypothesized causes \cite{Bach2015,Montavon2015,Sundararajan2016,Shrikumar2017,Smilkov2017,Sundararajan2017}.

Even though such attribution methods generally produce better visualizations, we find troubling that the hypotheses regarding noisy saliency maps have not been rigorously verified. In other words, numerous attribution methods were built upon unproven claims that gradient discontinuity or saturation truly causes saliency maps to be noisy. This situation gives rise to two major problems. First, if the hypotheses regarding noisy saliency maps are incorrect, current and future works based on those hypotheses will also be erroneous. Second, as we do not know precisely why saliency maps are noisy, we have to rely on heuristics and guesswork to develop better attribution methods.

In this paper, we propose a new hypothesis on noisy saliency maps to address these problems. We claim saliency maps are noisy because deep neural networks do not filter out irrelevant features during forward propagation. Then, we introduce \textit{Rectified Gradient}, or RectGrad in short, a technique that significantly improves the quality of saliency maps by alleviating the problem through layer-wise thresholding during backpropagation. Finally, we demonstrate that RectGrad produces attributions qualitatively and quantitatively superior to those of other attribution methods. Specifically, we have the following key contributions:

\begin{figure*}
	\centering
    \includegraphics[width=0.90\linewidth]{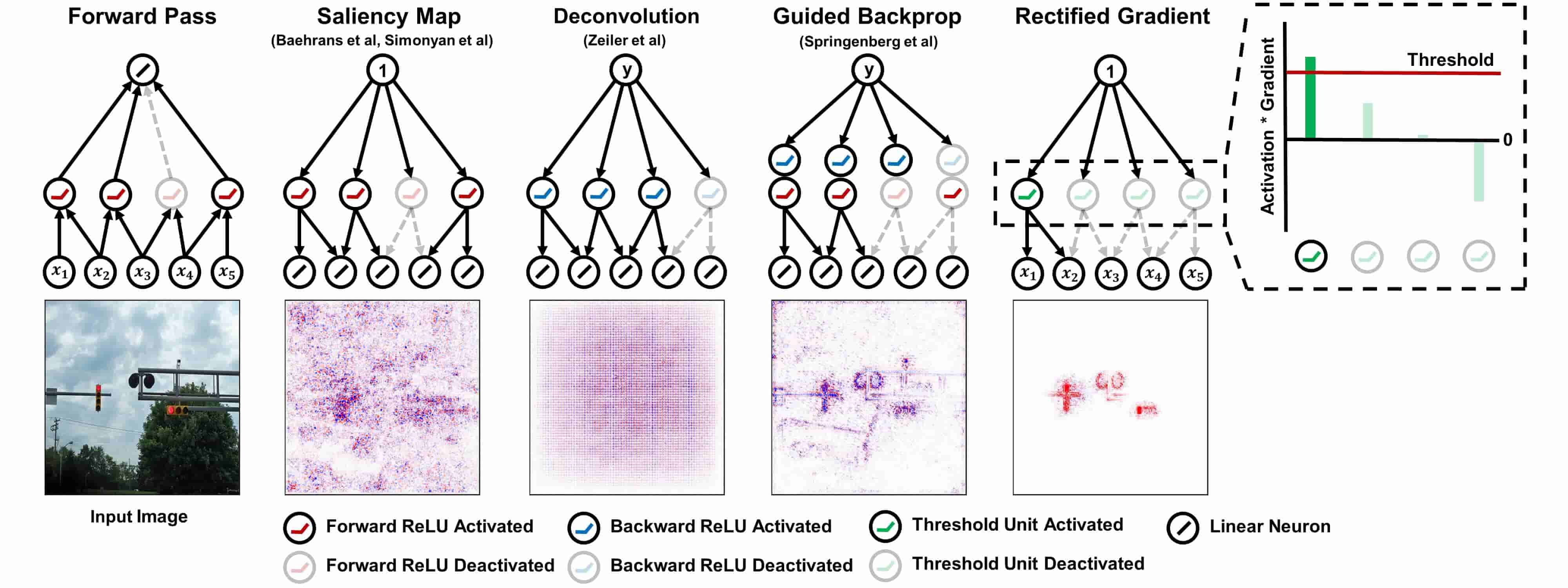}
    \caption{Comparison of attribution methods. See Appendix \ref{section:visualization} for details on the visualization.}
    \label{fig:comparison}
\end{figure*}

\begin{itemize}
    \item \textbf{Novel Hypothesis on Noisiness.} We propose a new perspective on the cause of noisy saliency maps. In our view, noise occurs in saliency maps when irrelevant features have positive pre-activation values and consequently pass through ReLU activation functions. This causes gradients to be nonzero at unimportant regions. We perform experiments with networks trained on CIFAR-10 to justify our claims (Section \ref{section:explanation}).
    \item \textbf{Proposal of a Solution.} We introduce RectGrad, an attribution method that removes noise from saliency maps by thresholding irrelevant units at ReLU binary gates during backpropagation (Section \ref{section:rectgrad formulation}). We prove that RectGrad generalizes Deconvolution and Guided Backpropagation (Section \ref{section:generalization}).
    \item \textbf{Solution Analysis.} We first investigate the effect of threshold level on attribution maps produced by RectGrad (Section \ref{section:threshold}). Then, we apply RectGrad to networks trained on  CIFAR-10 and ImageNet to demonstrate that it produces qualitatively and quantitatively superior attribution maps (Sections \ref{section:qualitative} and \ref{section:quantitative}).
    \item \textbf{Reproducibility.} We provide codes for all experiments as well as 1.5k samples for randomly chosen ImageNet images comparing RectGrad with other attribution methods at here\footnote{\texttt{https://github.com/1202kbs/Rectified-Gradient}}.
\end{itemize}

\section{Related Works}

Let $S_c$ be the score function of an image classification network for a class $c$.  Since functions comprising $S_c$ are differentiable or piecewise linear, the score function is also piecewise differentiable. Using this fact, \cite{Erhan2009}, \cite{Baehrens2009} and \cite{Simonyan2013} proposed the Saliency Map, or the gradient of $S_c$ with respect to input image $x$, to highlight features within $x$ that the network associates with the given class. In an ideal case, saliency maps highlight objects of interest. However, previous studies such as \cite{Springenberg2014} and \cite{Selvaraju2016} have pointed out that saliency maps tend to be visually noisy, as verified by Figure \ref{fig:comparison}. Three hypotheses were proposed to account for this phenomenon. We describe them in the next section.

\subsection{Previous Hypotheses} \label{section:hypotheses}

\textbf{Truthful Saliency Maps.} \cite{Smilkov2017} suggested the hypothesis that noisy saliency maps are faithful descriptions of what the network is doing. That is, pixels scattered seemingly at random are crucial to how the network makes a decision. In short, this hypothesis claims that noise is actually informative.

\textbf{Discontinuous Gradients.} \cite{Smilkov2017} and \cite{Shrikumar2017} proposed that saliency maps are noisy due to the piece-wise linearity of the score function. Specifically, since typical DNNs use ReLU activation functions and max pooling, the derivative of the score function with respect to the input will not be continuously differentiable. Under this hypothesis, noise is caused by meaningless local variations in the gradient.

\textbf{Saturating Score Function.} \cite{Shrikumar2017} and \cite{Sundararajan2017} suggested that important features may have small gradient due to saturation. In other words, the score function can flatten in the proximity of the input and have a small derivative. This hypothesis explains why informative features may not be highlighted in saliency maps even though they contributed significantly to the decision of the DNN.

\subsection{Previous Works on Improving Saliency Maps} \label{section:attribution methods}

DNN interpretation methods that assign a signed \emph{attribution} value to each input feature are collectively called \emph{attribution methods}. Attributions are usually visualized as a heatmap by arranging them to have the same shape as the input sample. Such heatmaps are called \emph{attribution maps}. We now describe attribution methods that have been proposed to improve saliency maps.

\textbf{Attribution Methods for Discontinuity.} SmoothGrad \cite{Smilkov2017} attempts to smooth discontinuous gradient with a Gaussian kernel. Since calculating the local average in a high dimensional space is intractable, the authors proposed a stochastic approximation which takes random samples in a neighborhood of the input $x$ and then averages their gradients.

\begin{figure*}[t]
	\centering
    \includegraphics[width=0.30\linewidth]{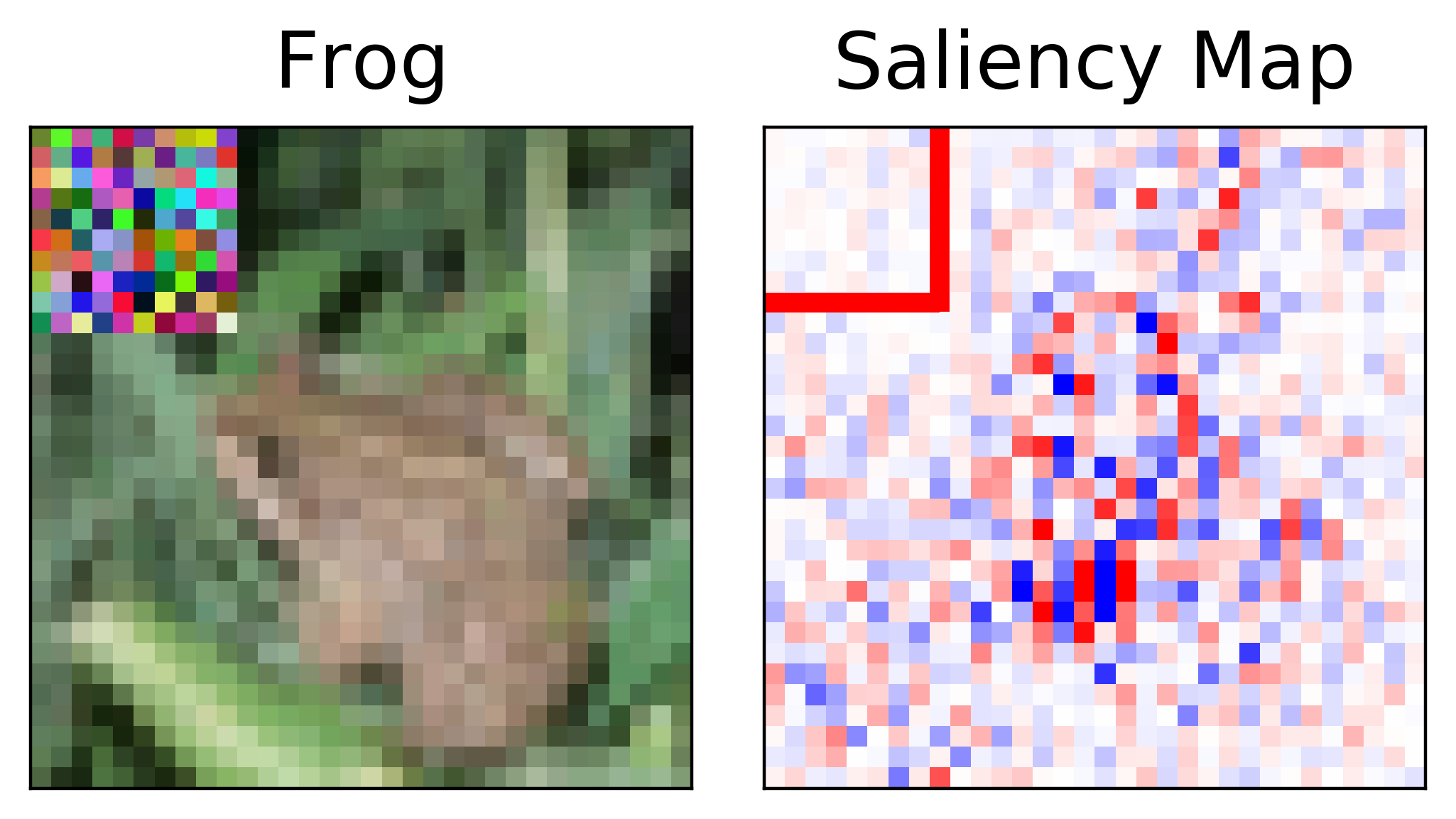}\hfill
    \includegraphics[width=0.30\linewidth]{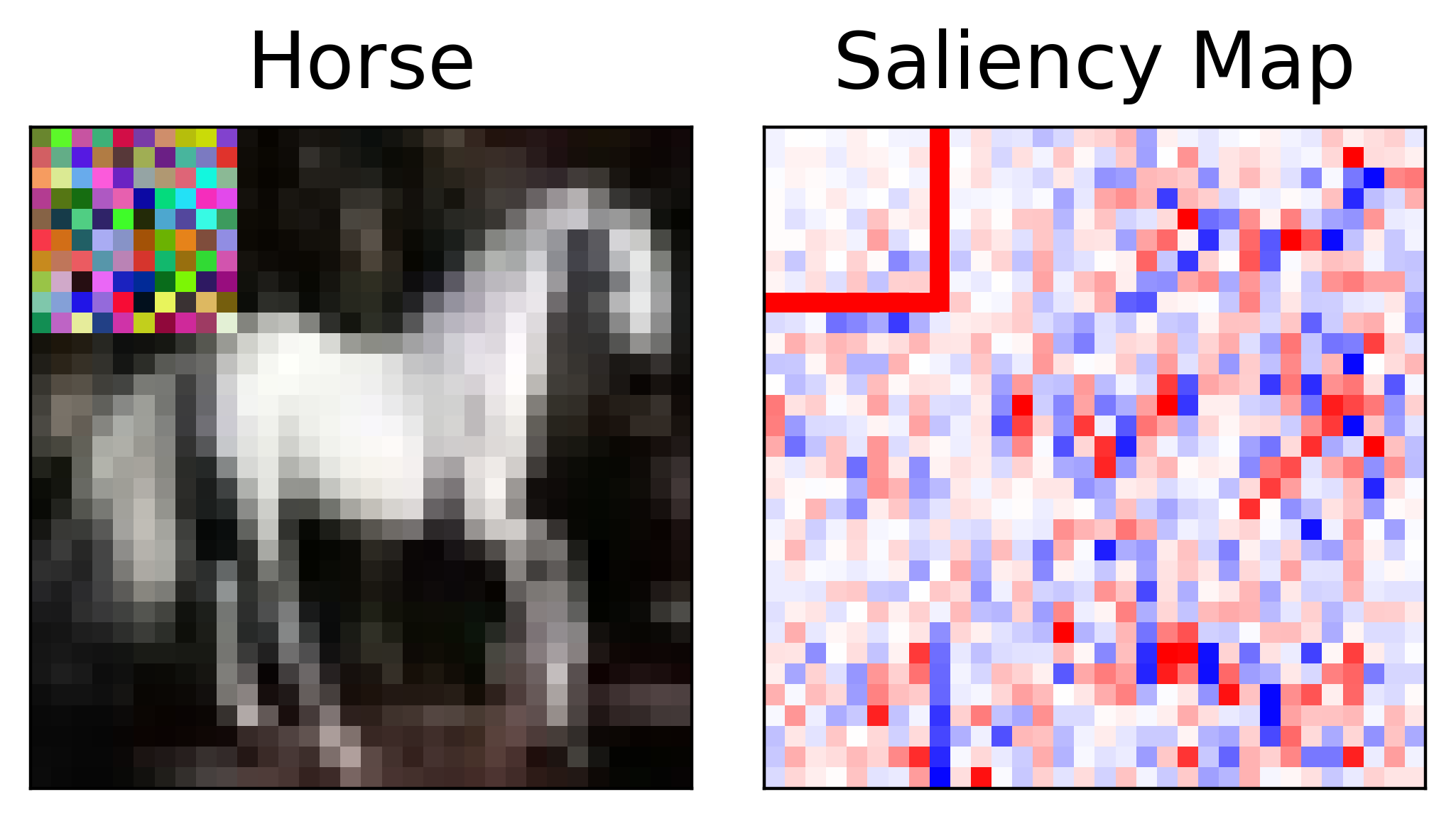}\hfill
    \includegraphics[width=0.30\linewidth]{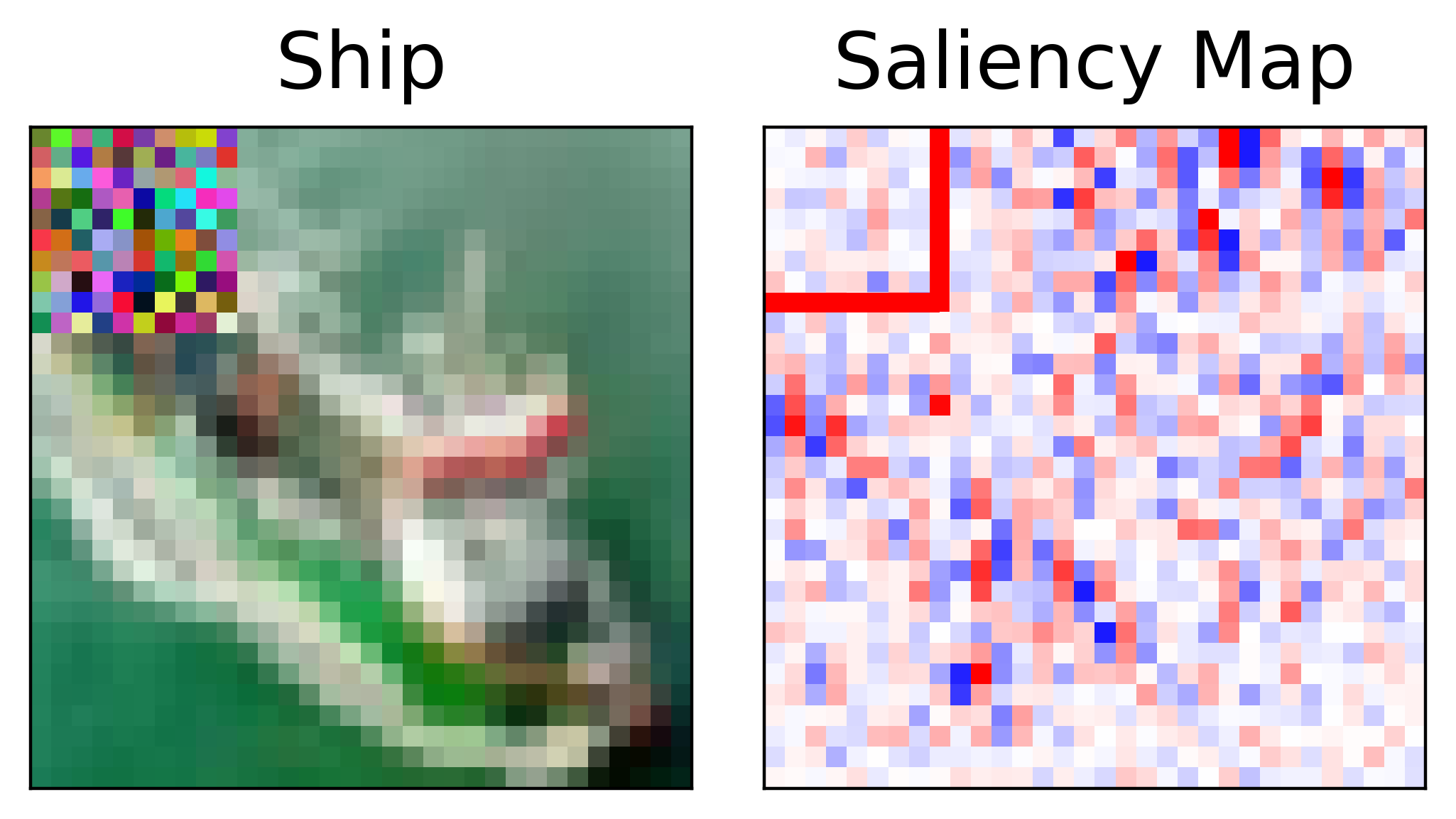}
    \caption{Saliency maps produced from a CNN trained on occluded images. The upper left corner of all the images in the training dataset is replaced with a 10 $\times$ 10 random patch, as shown above. Readers should examine the 8 $\times$ 8 patch enclosed by the red square instead of the entire 10 $\times$ 10 patch due to the receptive field of filters in the first convolution layer (3 $\times$ 3).}
    \label{fig:dataset occlusion}
\end{figure*}

\textbf{Attribution Methods for Saturation.} Since saliency maps estimate the local importance of each input feature, they are vulnerable to saturation. Therefore, attribution methods such as Gradient $*$ Input (Grad $*$ Input) \cite{Shrikumar2017}, Layer-wise Relevance Propagation (LRP) \cite{Bach2015}, DeepLIFT \cite{Shrikumar2017} and Integrated Gradient (IntegGrad) \cite{Sundararajan2017} attempt to alleviate saturation by estimating the global importance of each pixel \cite{Ancona2017}.

\textbf{Other Attribution Methods.} The rest of attribution methods take a different approach to improving saliency maps. Deconvolution (Deconv) \cite{Zeiler2014} and Guided Backpropagation (Guided BP) \cite{Springenberg2014} remove negative gradient during backpropagation. Due to this imputation procedure, Deconv and Guided BP yield attribution maps sharper than those of other methods. However, \cite{Nie2018} has recently proven that these methods are actually doing partial image recovery which is unrelated to DNN decisions.

\section{Explaining Noisy Saliency Maps} \label{section:explanation}

\subsection{Examining Untruthfulness of Saliency Maps}
If noisy saliency maps are truthful as \cite{Smilkov2017} mentioned, features corresponding to noise are crucial in the model's decision and there is little need to improve saliency maps. We claim that this is not the case. In this section, we give an evidence that saliency maps can be untruthful.

Our idea is quite simple. To show that saliency maps can highlight completely uninformative features, we can train DNNs on data including explicitly uninformative features and examine their saliency maps. Data with explicitly uninformative features can be constructed from normal data by replacing some features with uninformative values. For example, in the image domain, some specific spatial part of each image can be replaced by uniform noise. If saliency maps highlight the irrelevant features, it implies that saliency maps  are untruthful. We call this data synthesis based approach \emph{Training Dataset Occlusion}.

We performed training dataset occlusion with the CIFAR10 dataset. Specifically, we occluded the upper left corner of all images in the training dataset with a $10 \times 10$ random patch. Each pixel value in the random patch was randomly and uniformly sampled from the interval $[0,1]$. Then, we trained a convolutional neural network (CNN) on the modified dataset. For details of the experiment, please refer to Appendix \ref{section:setup}. Note that the final test accuracy did not change significantly.

Figure \ref{fig:dataset occlusion} illustrates some samples of our training dataset occlusion experiments. It is obvious that \textbf{saliency maps are nonzero on occluded patch although the patch is completely irrelevant to the classification task}. We observed this tendency was consistent across test data. The results clearly imply that saliency maps of the common CNN based classification models can be untruthful. We collected some statistics of this examination, and they are reported on Section \ref{quant: training}. However, note that this result does \textit{not} imply that all noise in saliency maps correspond to irrelevant features.

\paragraph{Where does Untruthfulness Arise From?}

Recently, several works \cite{morcos2018importance, tsipras2019there, zhang2019theoretically} have pointed out that normally trained classifiers tend to learn weakly correlated features.
In connection to this, there are some works \cite{ross2018improving, tsipras2019there, kim2019safeml} which have reported that adversarially trained DNNs have much cleaner saliency maps than normally trained DNNs.
These works imply that standard DNNs tend to retain features entirely, rather than drop features according to relevance.

Considering these discussions, we propose a new hypothesis on the noisiness of saliency maps.
For ReLU DNNs, nonzero gradient for some feature indicates the presence of at least one positive pre-activation in each layer spatially corresponding to that feature.
Thus, saliency maps are noisy and less relevant to decision \textbf{when deep neural networks do not filter out irrelevant features during forward propagation}.
In the next subsection, we will demonstrate that DNNs trained under practical classification settings indeed propagate irrelevant features.

\subsection{Examining Filtering Ability of Classifier}

\paragraph{Exploring Feature Importance in Feature Space}
Given that noise in saliency maps often occurs in the background, not on the object, we first define two types of feature depending on its position in the image: \emph{background feature} and \emph{foreground feature}.
As their names suggest, background features are pixels on the background and object features are pixels on the object.
We now show that background features are much less relevant to the model decision than object features.

To achieve this, we occluded background / object feature activations at intermediate layers and analyzed the effect of this occlusion on the final decision.
Note that this process differs from the Sensitivity metric \cite{Bach2015,Samek2017} which is also based on occlusion.
Sensitivity measures the impact of occlusion in the data space (e.g. pixel occlusion) while we measured the impact of occlusion in each feature space.
We first created segmentation masks for 10 correctly classified images of each class (total 100 images).
We then plotted the average of (class logit) $-$ (largest logit among the other 9 classes) as we incrementally occluded background feature activations in a random order (average is taken over 50 random trials) across all 100 images.
We call this experiment \emph{Feature Map Occlusion} in order to avoid confusion with training dataset occlusion.

\begin{figure}[t]
	\centering
    \includegraphics[width=0.7\linewidth]{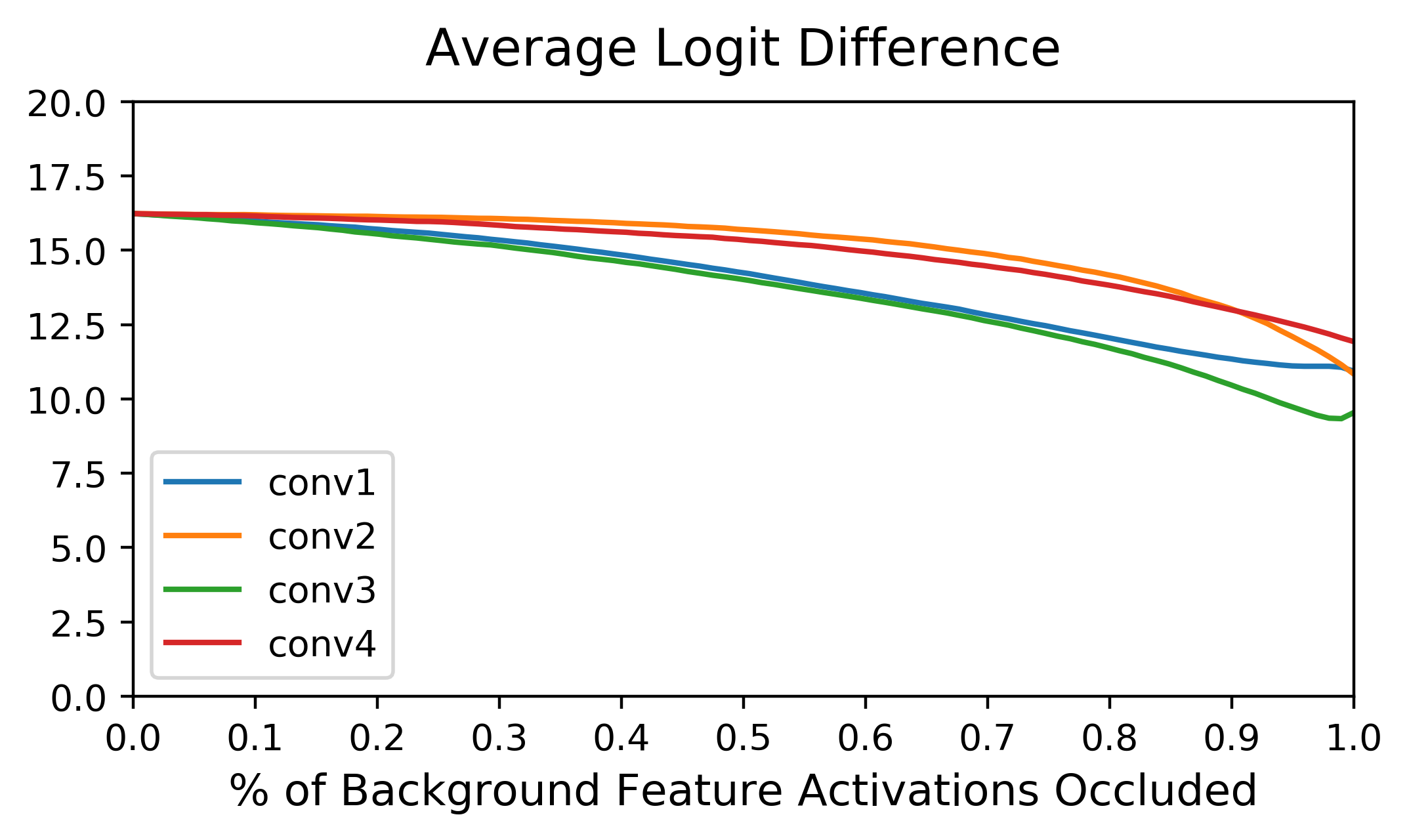}
    \includegraphics[width=0.7\linewidth]{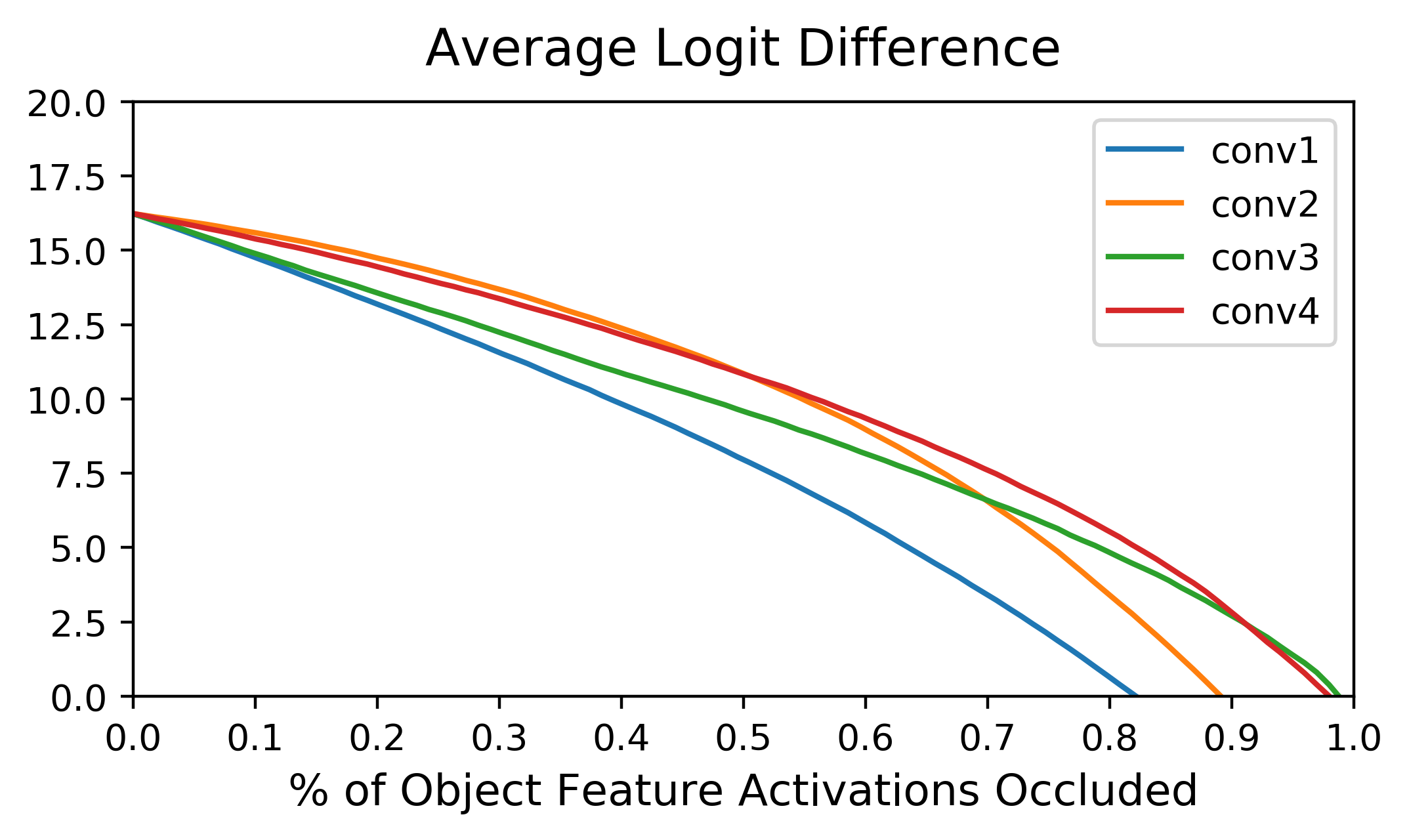}
    \caption{Average of (class logit) $-$ (largest logit among the other 9 classes) as background or object feature activations are incrementally occluded in a random order (average is taken over 50 random trials) across 100 images.}
    \label{fig:occlusion}
\end{figure}

Figure \ref{fig:occlusion} shows that occluding object features has a larger impact on the final decision than occluding background features.
From this, we can infer that object features are much more relevant to the model decision than background features.\\

\paragraph{Observation on Feature Appearance Frequency}

Figure \ref{fig:feature map} in Appendix \ref{section:feature visualization} shows noisy saliency maps and convolutional layer feature maps for the corresponding images.
The tendency of the CNN to retain or drop the two types of feature is not significantly different, although object features are far more relevant to the model decision than background features.
These qualitative observations constitute the evidence that DNN filters tend to retain or drop features regardless of their relevance.
Quantitative measurement of feature activation frequency is left as a future study.

\section{Proposed Solution to Noisy Saliency Maps} \label{section:rectgrad}

As we have shown in Section \ref{section:explanation}, the standard DNNs do not retain or discard features in terms of their relevance.
In our view, this is why saliency maps describing DNN  decisions are noisy.
This fact suggests that we need a new attribution method which drops irrelevant features while retaining relevant features in layer-wise fashion.
To this end, we propose \textit{Rectified Gradient}, or RectGrad in short, where the gradient propagates only through units whose importance scores exceed some threshold.
By construction, RectGrad drops irrelevant features while retaining relevant features in a layer-wise fashion.
Importance score for an unit is calculated by multiplying its activation with gradient propagated up to the unit.

\subsection{Formulation of Rectified Gradient} \label{section:rectgrad formulation}

Suppose we have a $L$-layer ReLU DNN. Denote input feature $i$ as $x_i$, pre-activation of unit $i$ in layer $l$ as $z^{(l)}_i$, its activation as $a^{(l)}_i$ and gradient propagated up to $a^{(l)}_i$ as $R^{(l + 1)}_i$. Let $\mathbb{I}(\cdot)$ be the indicator function. Then, the relation between $a^{(l)}_i$ and $z^{(l)}_i$  is given by $a^{(l)}_i = ReLU(z^{(l)}_i) = \max (z^{(l)}_i, 0)$ when $l < L$ and $a^{(L)}_i = softmax(z^{(L)}_i)$. By the chain rule, backward pass through the ReLU nonlinearity for vanilla gradient is achieved by $R^{(l)}_i = \mathbb{I} (a^{(l)}_i > 0) \cdot R^{(l + 1)}_i$.

We modify this rule such that $R^{(l)}_i = \mathbb{I}(a^{(l)}_i \cdot R^{(l + 1)}_i > \tau) \cdot R^{(l + 1)}_i$ for some threshold $\tau$. Backward pass through affine transformations and pooling operations is carried out in the same manner as backpropagation. Finally, importance scores for input features are calculated by multiplying gradient propagated up to input layer ($l = 0$) with input features and thresholding at zero: $x_i \cdot R^{(1)}_i \cdot \mathbb{I}(x_i \cdot R^{(1)}_i > 0)$. Instead of setting $\tau$ to a constant value, we use the $q$\textsuperscript{th} percentile of importance scores at each layer. This prevents the gradient from entirely dying out during the backward pass. Note that this notion of thresholding units by importance score holds regardless of the DNN architecture and activation used. We explain the rationale behind this propagation rule in Appendix \ref{section:rationale}.

Due to the simplicity of the propagation rule, RectGrad can easily be applied to DNNs in graph computation frameworks such as TensorFlow \cite{tensorflow2015-whitepaper} or PyTorch \cite{paszke2017automatic}. Listing \ref{lst:rectified relu} in Appendix \ref{section:code1} shows how to implement RectGrad in TensorFlow. In Appendix \ref{section:techniques} we also introduce two techniques, namely the padding trick and the proportional redistribution rule (PRR) that enhance the visual quality of RectGrad attribution maps.

\begin{figure*}
	\centering
    \includegraphics[width=0.92\linewidth]{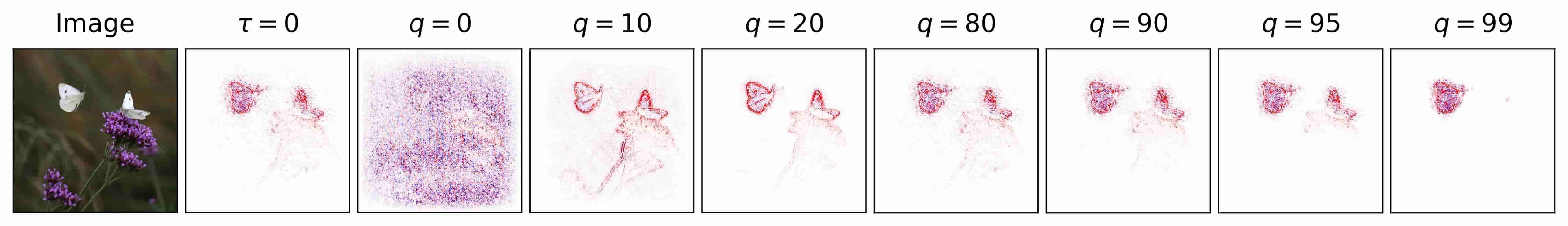}
    \includegraphics[width=0.92\linewidth]{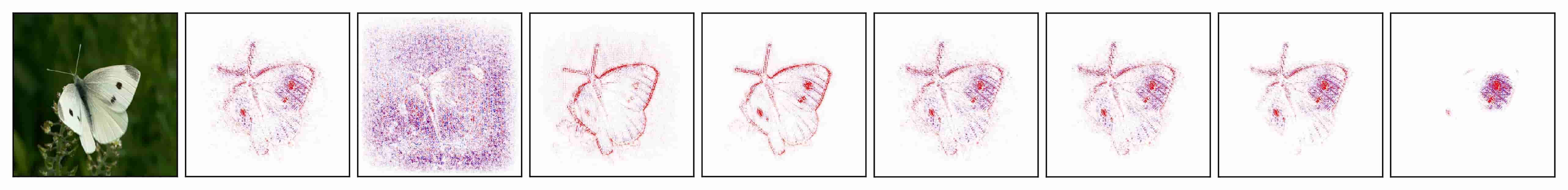}
    \includegraphics[width=0.92\linewidth]{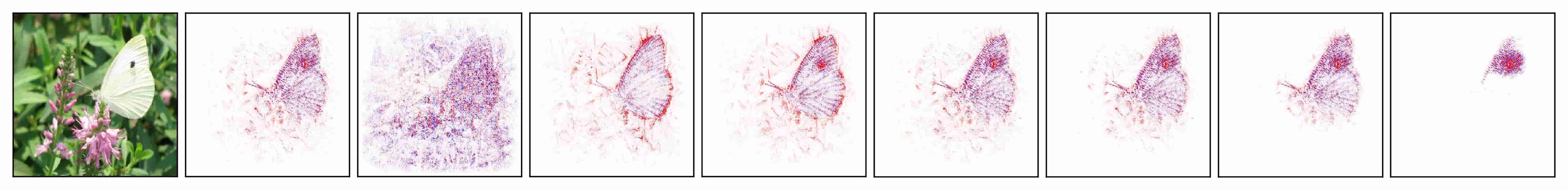}
    \caption{Effect of threshold $\tau$ (columns) on RectGrad for 3 images of the cabbage butterfly class in ImageNet (rows). The second column shows attribution maps with $\tau = 0$, which is equivalent to Guided Backpropagation $*$ Input. For the following columns, $\tau$ is set to $q$\textsuperscript{th} percentile of importance scores. The padding trick was used for all attribution maps above.}
    \label{fig:hyperparameter}
\end{figure*}

\subsection{Relation to Deconvolution and Guided Backpropagation} \label{section:generalization}

\begin{proposition} \label{claim:deconv}
Deconvolution $*$ Input with final zero thresholding is equivalent to RectGrad with the propagation rule
\begin{equation*}
R^{(l)}_i = \mathbb{I} \left[ \left( a^{(l)}_i + \epsilon \right) \cdot R^{(l + 1)}_i > 0 \right] \cdot R^{(l + 1)}_i
\end{equation*}
for small $\epsilon > 0$.
\end{proposition}

\begin{proposition} \label{claim:gbp}
Guided Backpropagation $*$ Input with final zero thresholding is equivalent to RectGrad when $\tau = 0$:
\begin{equation*}
R^{(l)}_i = \mathbb{I} \left( a^{(l)}_i \cdot R^{(l + 1)}_i > 0 \right) \cdot R^{(l + 1)}_i.
\end{equation*}
\end{proposition}

We provide the proofs for Claims \ref{claim:deconv} and \ref{claim:gbp} in Appendix \ref{section:proofs}. These results indicate that RectGrad generalizes Deconv and Guided BP. Figure \ref{fig:comparison} illustrates the relation between the Saliency Map, Deconv, Guided BP and RectGrad.

However, \cite{Nie2018} has recently proven that Deconv and Guided BP are actually doing partial image recovery which is unrelated to DNN decisions. RectGrad does \emph{not} suffer from this problem as it does not satisfy the assumptions of the analyses of \cite{Nie2018} for two reasons. First, the threshold criterion is based on the product of activation and gradient which is not Gaussian distributed.\footnote{Product of a half normal random variable and a normal random variable is not Gaussian distributed.} Second, we set $\tau$ as the $q$\textsuperscript{th} percentile of importance scores and therefore $\tau$ will vary layer by layer. We also show in Section \ref{section:qualitative} with adversarial attacks that attributions produced by RectGrad are class sensitive. Therefore, RectGrad inherits the sharp visualizations of Deconv and Guided BP while amending their disadvantages with layer-wise importance score thresholding.

\section{Experiments} \label{section:experiments}

To evaluate RectGrad, we performed a series of experiments using Inception V4 network \cite{Szegedy2016} trained on ImageNet \cite{Imagenet2015} and CNNs trained on CIFAR-10 \cite{cifar2009}. In order to prove that the superior visual quality of RectGrad attributions is not simply due to final zero thresholding, we visualize RectGrad attributions without final zero thresholding. See Appendix \ref{section:setup} for details on experiment settings and attribution map visualization method.

\subsection{Effect of Threshold Percentile} \label{section:threshold}

RectGrad has one hyper-parameter $\tau$, which is set to $q$\textsuperscript{th} percentile of importance scores for each layer. Figure \ref{fig:hyperparameter} shows the effect of threshold percentile for several images from ImageNet. While the attribution maps were incomprehensible for $q = 0$, the visual quality dramatically improved as we incremented $q$ up to $20$. There was no significant change up to $q = 80$. Then the attribution maps began to sparse out again as we incremented $q$ further. We also observed that regions of high attributions did not change from $q > 20$.

We speculate that the attributions stay constant between $q = 20$ and $80$ because of zero activations. That is, since we use ReLU activation functions, the majority of activations and consequently importance scores will be zero. Hence, $\tau \approx 0$ for $20 \leq q \leq 80$. This causes RectGrad attribution maps to resemble those produced by Guided Backpropagation $*$ Input. It indicates that we have to increment $q > 80$ in order to produce sparser attribution maps that highlight important regions instead of reconstruct input images.

\begin{figure*}[!htb]
	\centering
	\begin{subfigure}{0.495\linewidth}
	\centering
	\includegraphics[width=0.49\linewidth]{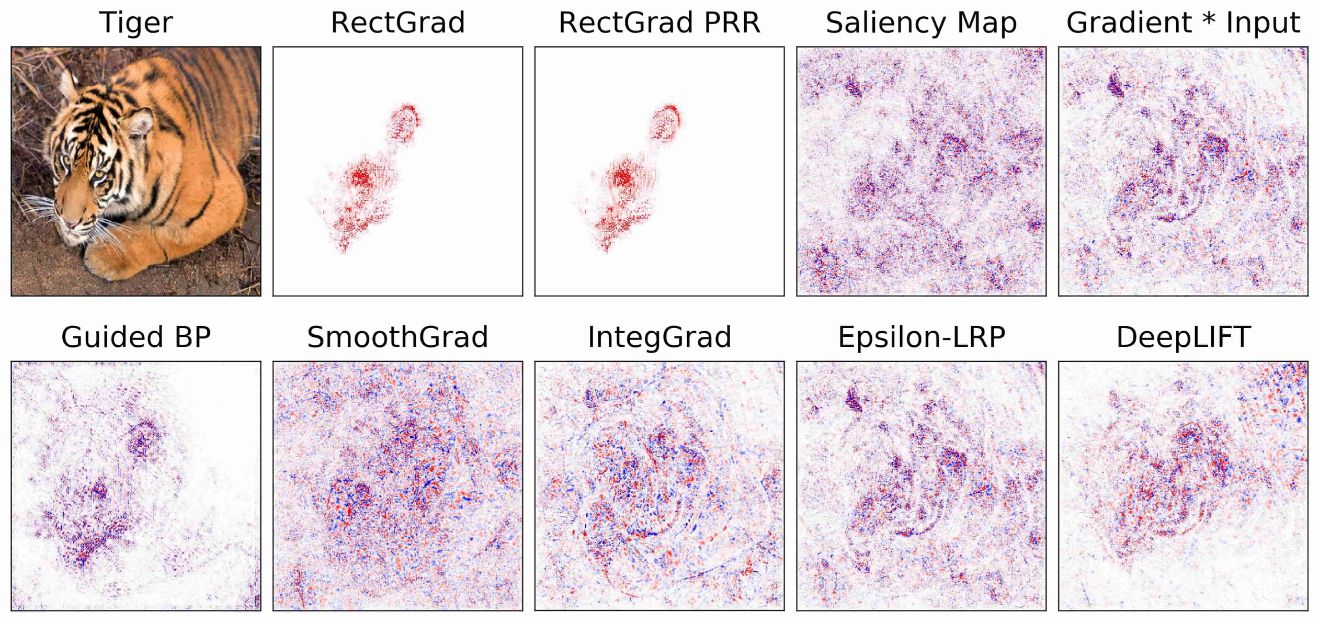}
    \includegraphics[width=0.49\linewidth]{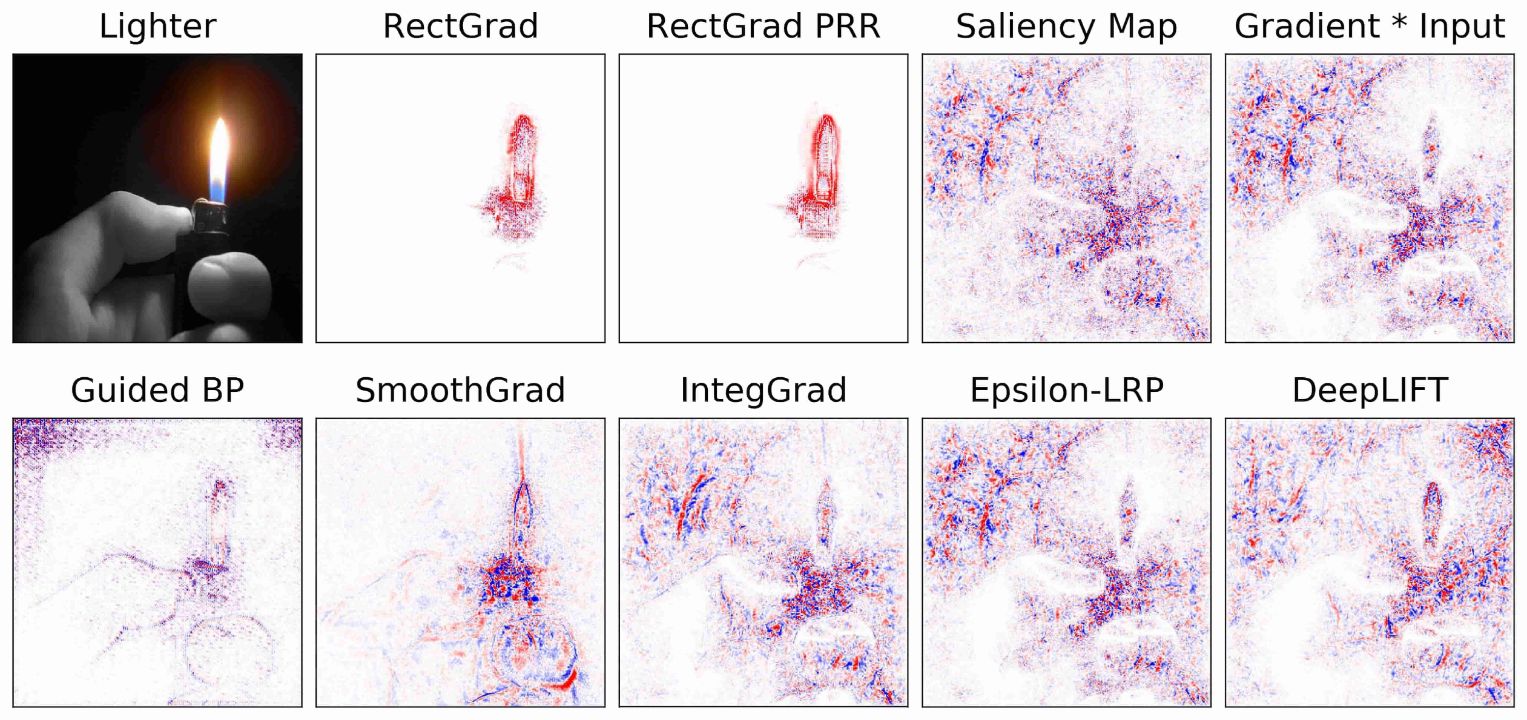}
    \par\smallskip
    \includegraphics[width=0.49\linewidth]{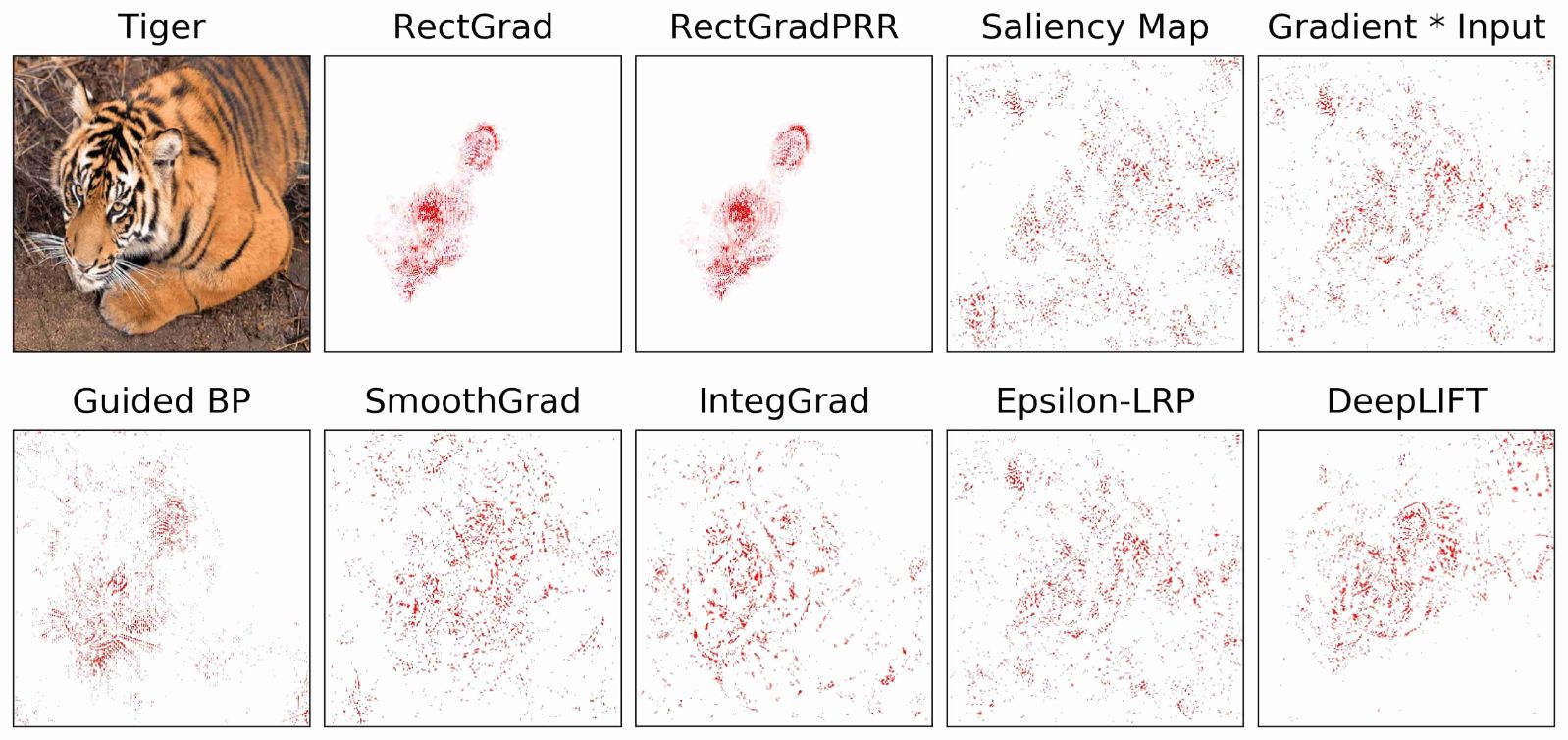}
    \includegraphics[width=0.49\linewidth]{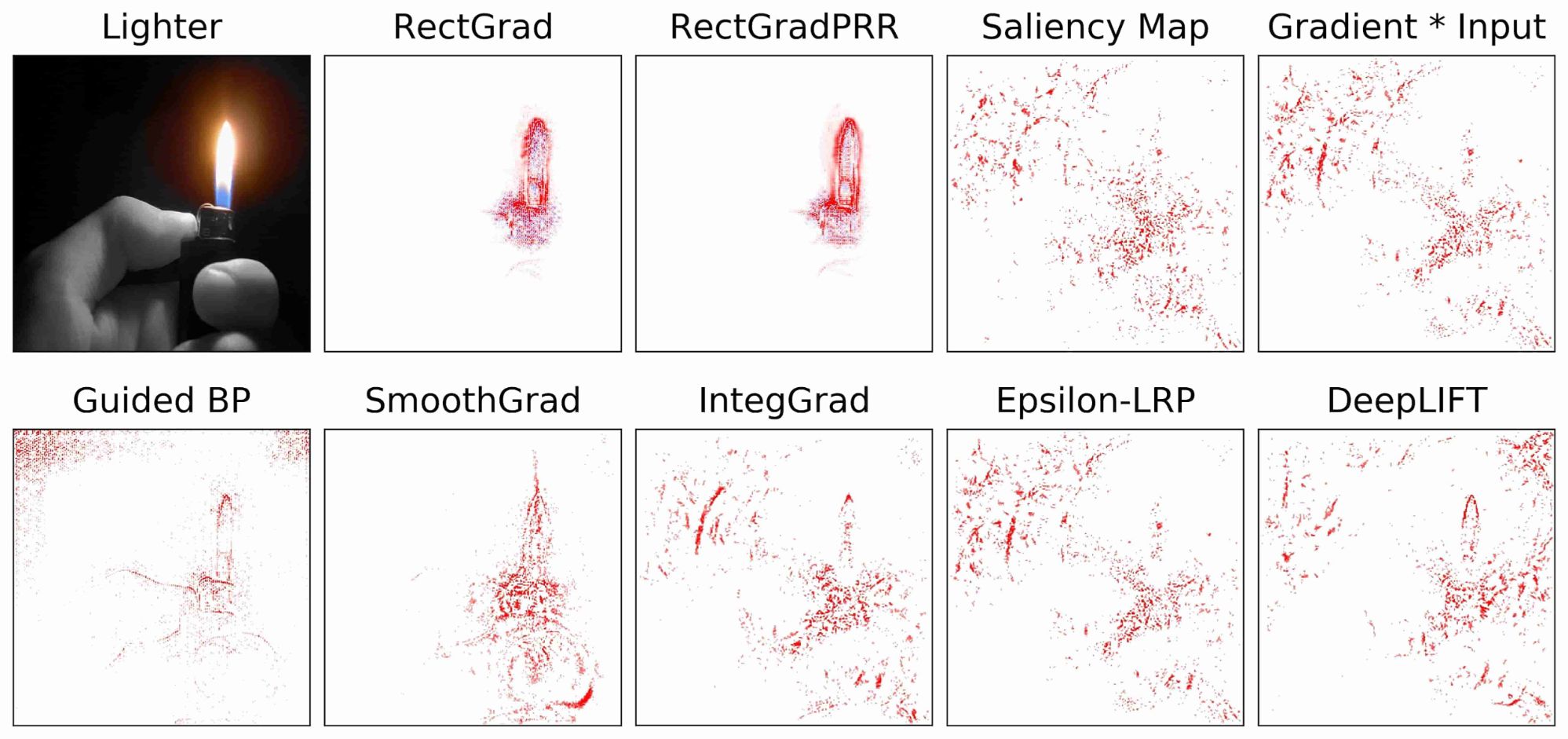}
    \caption{Evaluation of coherence across different classes without and with final thresholding.}
    \label{fig:coherence1}
    \end{subfigure} \hfill
    \begin{subfigure}{0.495\linewidth}
	\centering
    \includegraphics[width=0.49\linewidth]{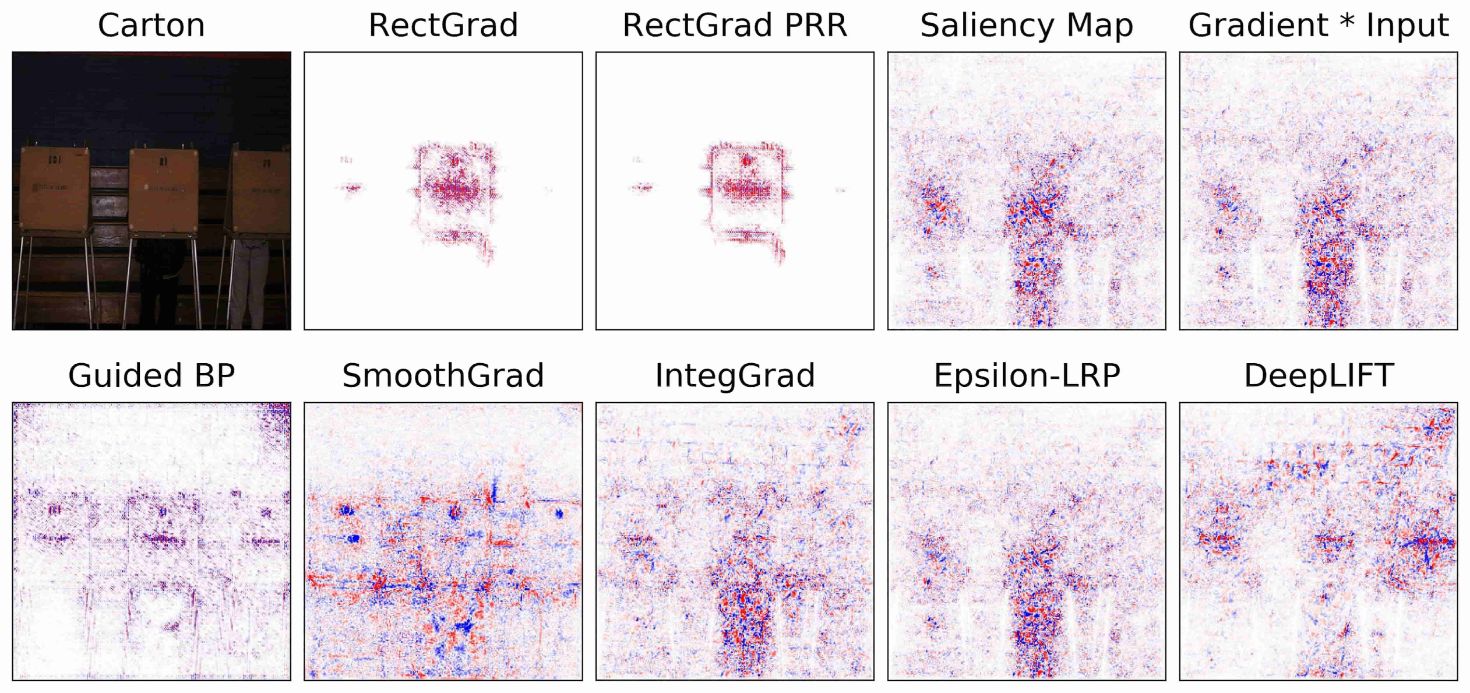}
    \includegraphics[width=0.49\linewidth]{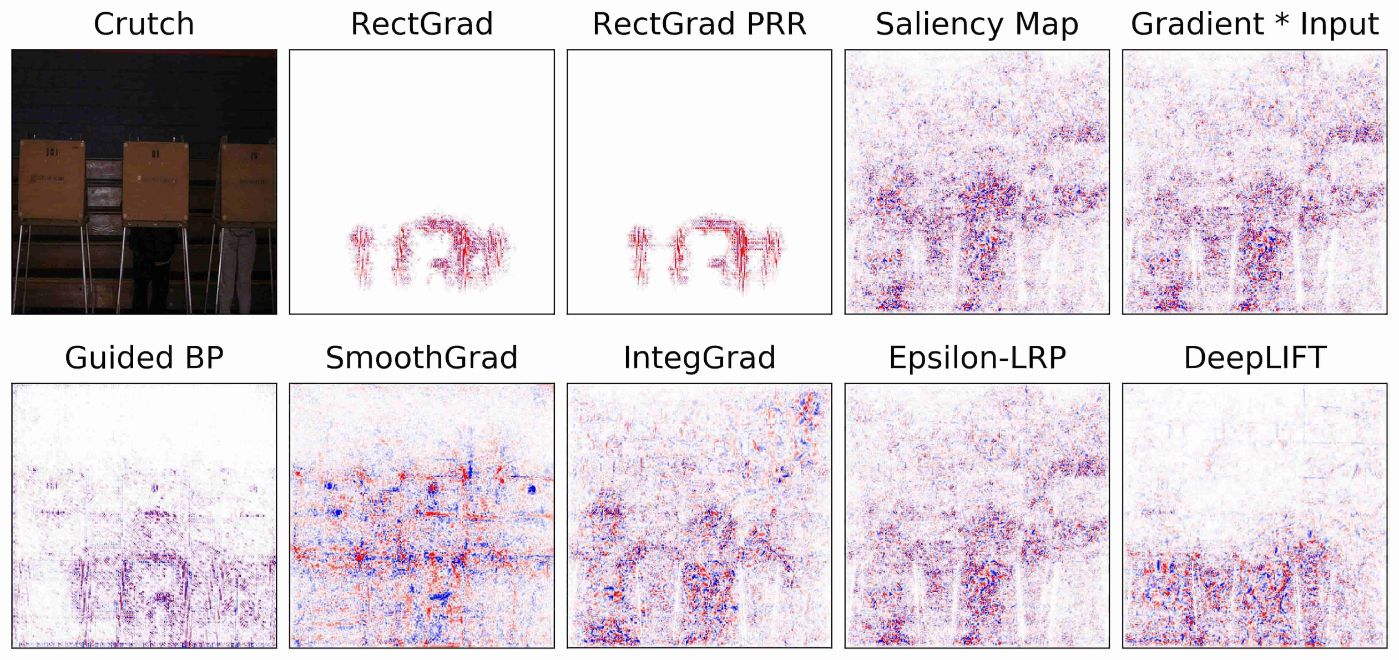}
    \par\smallskip    
    \includegraphics[width=0.49\linewidth]{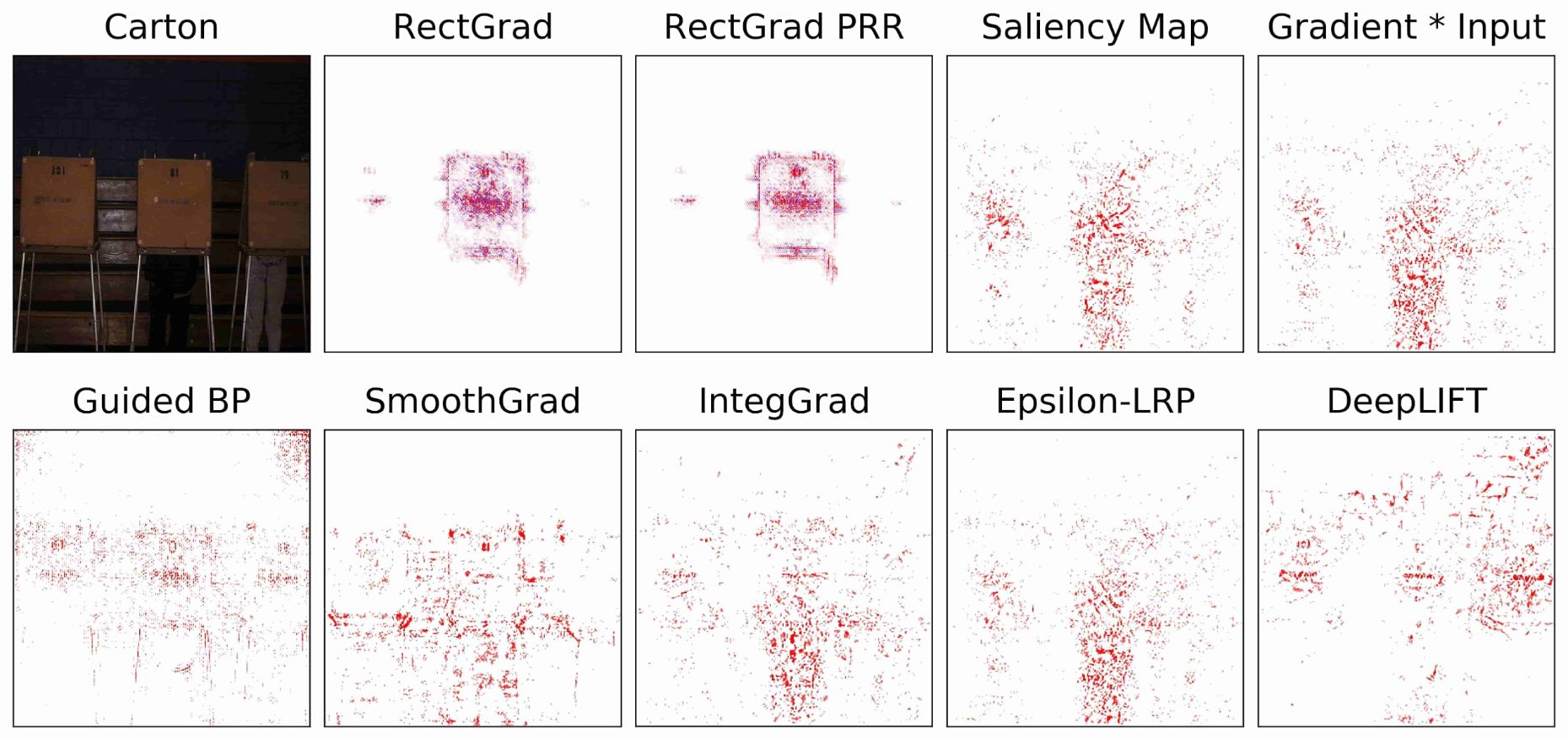}
    \includegraphics[width=0.49\linewidth]{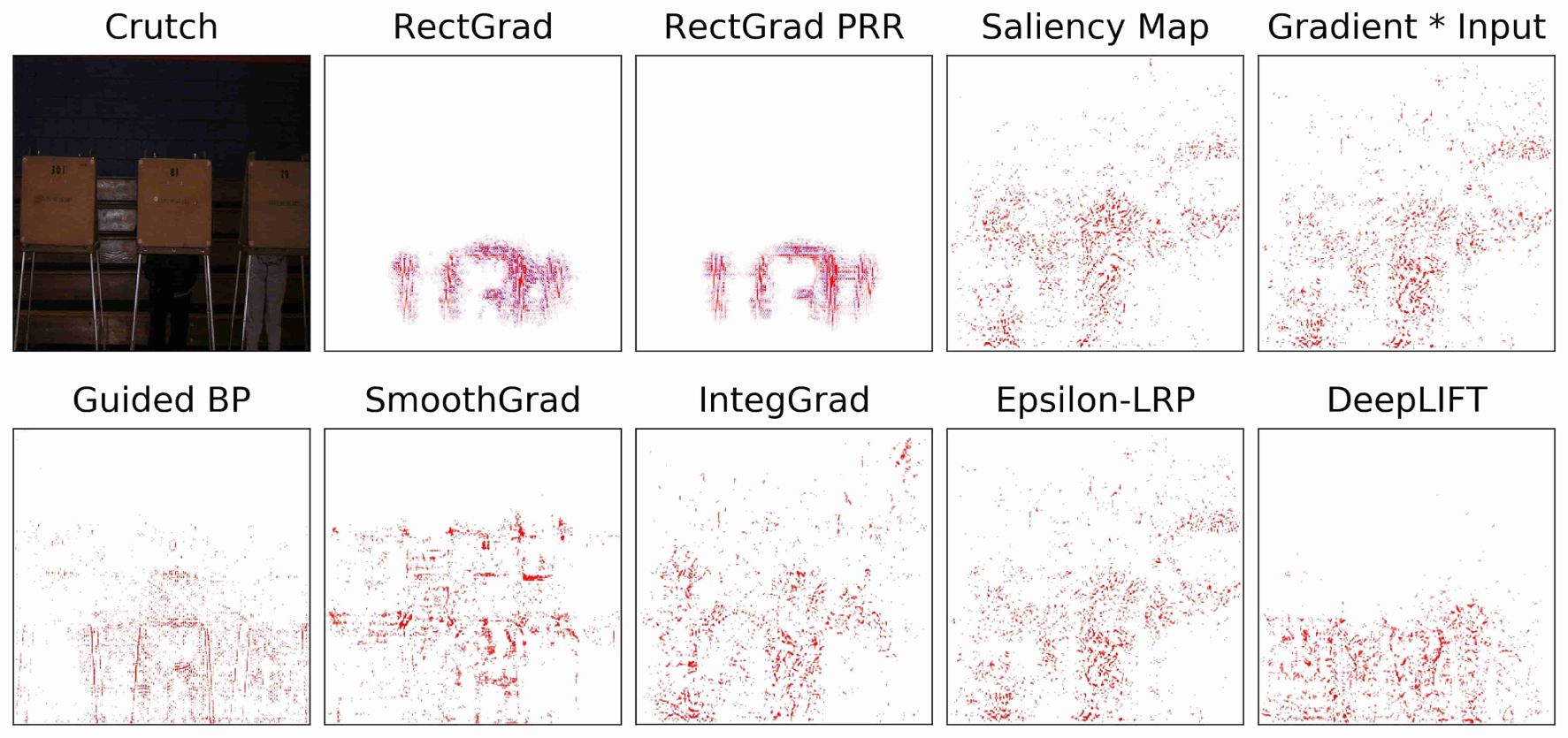}
    \caption{Attribution maps for images (left column) and their adversarial examples (right column) without and with final thresholding.}
    \label{fig:adv1}
    \end{subfigure}
    \caption{Qualitative comparison of coherence and class sensitivity.}
\end{figure*}

\begin{figure*}[!htb]
	\raggedright
    \includegraphics[width=0.3\linewidth]{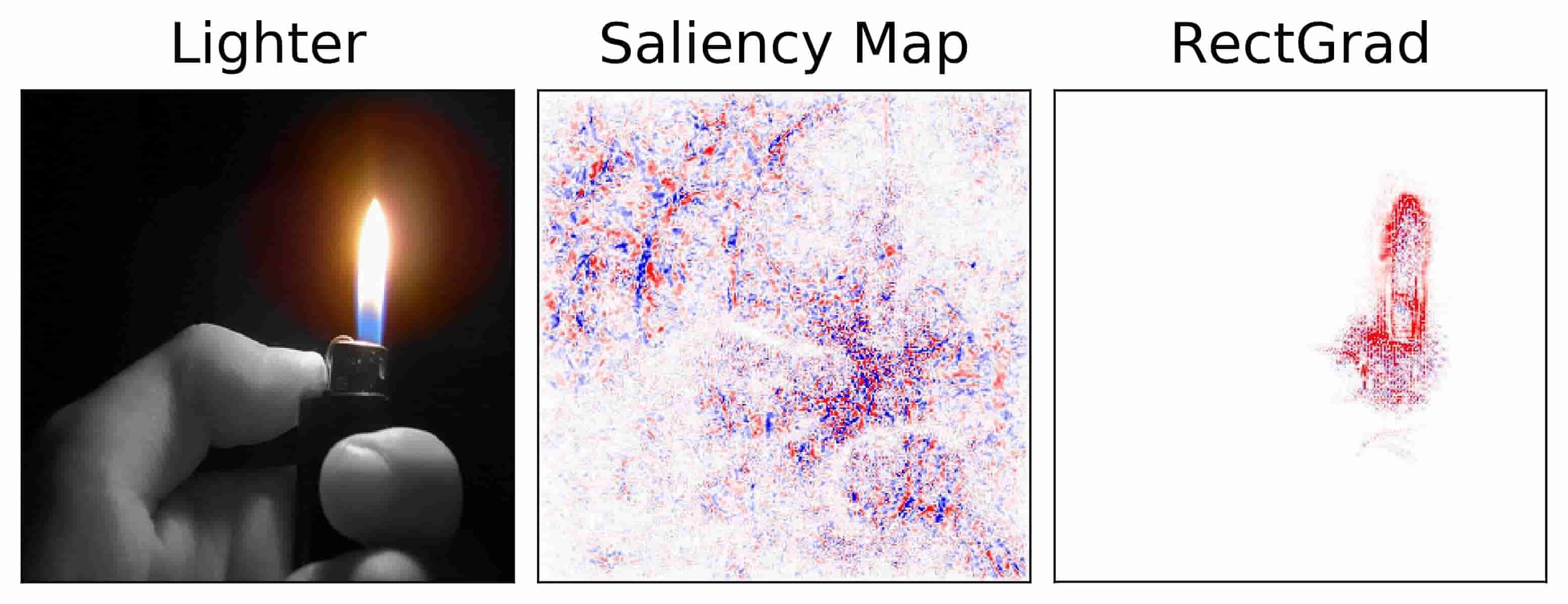}
    \includegraphics[width=\linewidth]{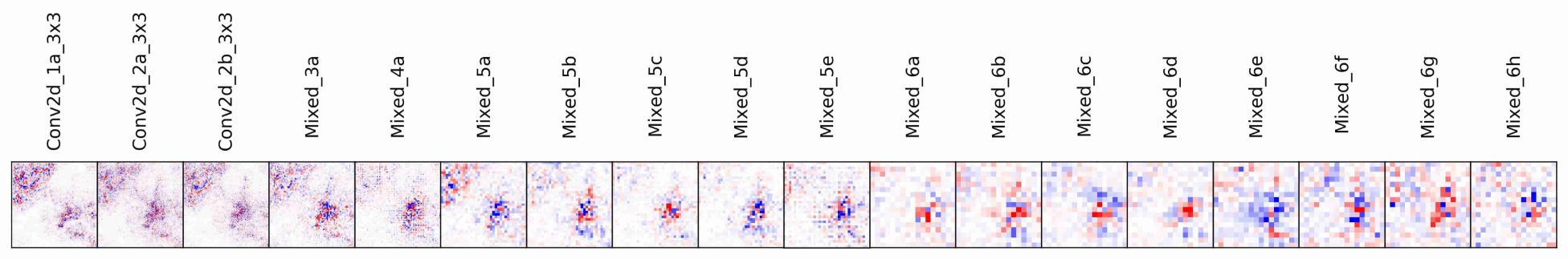}
    \includegraphics[width=\linewidth]{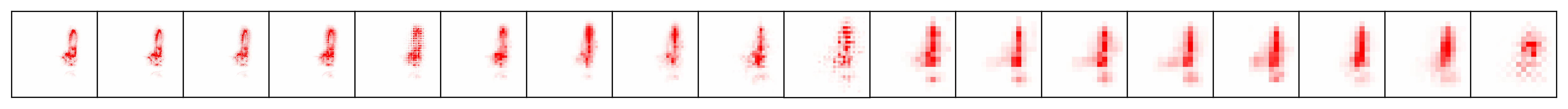}
    \caption{Saliency Map (middle) and RectGrad attributions (bottom) at Inception v4 intermediate layers as they are propagated toward the input layer. We show channel-wise average attributions for hidden layer inputs with respect to the output layer. An attribution map is closer to the output layer if it is closer to the right.}
    \label{fig:accum1}
\end{figure*}

This shows we can control the hyper-parameter threshold percentile $q$ to vary the degree to which RectGrad emphasizes important features. Attribution maps with $80 < q < 90$ highlight the object of interest (the cabbage butterfly) along with auxiliary objects such as flowers or grass that may be helpful to the DNN in identifying the object. On the other hand, attribution maps with $q > 95$ highlight features that may have been most influential to the final decision, namely the spots on the butterfly's wing.

\subsection{Qualitative Comparison with Baseline Methods} \label{section:qualitative}

We used Saliency Map, Grad $*$ Input, Guided BP, SmoothGrad, IntegGrad, Epsilon-LRP and DeepLIFT as baseline methods. For RectGrad, we used the padding trick with $q = 98$. We show attributions both with and without application of the proportional redistribution rule (PRR). In this subsection, we compare RectGrad with other attribution methods through two experiments that each focus on different aspect of qualitative evaluation.

To show that applying simple final thresholding to baseline methods is not enough to replicate the benefits of RectGrad, we applied 95 percentile final threshold to baseline attribution methods such that RectGrad and baseline attribution maps have similar levels of sparsity. \footnote{Note that we did not apply $q = 98$ threshold, which was used in our RectGrad results. Baseline attribution maps with $q = 98$ are slightly more sparse than RectGrad attribution maps under the same setting. This is because for RectGrad, $q = 98$ threshold is applied up to the first hidden layer, not the input layer.}

\textbf{Coherence.} Following prior work \cite{Simonyan2013,Zeiler2014}, we inspected two types of visual coherence. First, the attributions should fall on discriminative features (e.g. the object of interest), not the background. Second, the attributions should highlight similar features for images of the same class.

For the first type of visual coherence, Figure \ref{fig:coherence1} shows a side-by-side comparison between our method and baseline methods. It can clearly be seen that RectGrad produced attribution maps more visually coherent and focused than other methods---background noise was nearly nonexistent.

To further investigate why Saliency Map assigns large attributions to irrelevant regions (e.g. uniform background in the \enquote{lighter} example) while RectGrad does not, we compared their attributions as they are propagated towards the input layer. The results are shown in Figure \ref{fig:accum1}. We observed that the background noise in saliency maps is due to noise accumulation. Specifically, irrelevant features may have relatively small gradient at high intermediate layers. However, since gradient is calculated by successive multiplication, the noise grows exponentially as gradient is propagated towards the input layer. This results in confusing attribution maps which assign high attribution to irrelevant regions. We also observed that RectGrad does not suffer from this problem since it thresholds irrelevant features at every layer and hence stops noise accumulation. In this situation, final thresholding cannot replicate RectGrad's ability to remove noise. We show additional examples in Appendix \ref{section:accum}.

For the second type of visual coherence, Figure \ref{fig:coherence2} in Appendix \ref{section:qualitative figures} shows attribution maps for a pair of images belonging to the same class. Attribution maps generated by RectGrad consistently emphasized similar parts of the object of interest. On the contrary, Saliency Map, Gradient $*$ Input and Epsilon-LRP emphasized different regions for each image instance. Attributions for SmoothGrad, Guided Backpropagation, Integrated Gradient and DeepLIFT were generally coherent across images of the same class. Nevertheless, they also highlighted background features and hence failed to satisfy the first type of visual coherence. This observation also holds for attribution maps with final thresholding.\footnote{\label{footnote:note1} We have also surveyed attribution maps for 1.5k randomly chosen ImageNet images and found them to be generally consistent with our claims. The links to Google drives containing the random samples can be found at \texttt{https://github.com/1202kbs/Rectified-Gradient}.}

\textbf{Adversarial Attack.} We evaluated class sensitivity following prior work by \cite{Nie2018}. Specifically, we compared the attributions for an image and its adversarial example. If the attribution method is class sensitive, attribution maps should change significantly since ReLU activations and consequently the predicted class have changed. On the other hand, if the attribution method merely does image reconstruction, attribution maps will not change much since we add an indistinguishable adversarial perturbation to the image. In this experiment, we used the fast gradient sign method \cite{Goodfellow2014} with $\epsilon = 0.01$ to generate adversarial examples.

Figure \ref{fig:adv1} shows large changes in attribution maps produced by RectGrad. We observed that only RectGrad attributions were coherent with the class labels. Figure \ref{fig:adv2} in Appendix \ref{section:qualitative figures} shows some instances where there was no significant change in attribution maps produced by RectGrad. In those cases, attribution maps for other methods also showed little change. Hence, we can conclude that RectGrad is equally or more class sensitive than baseline attribution methods. We observed that this conclusion also holds with final thresholding. It is also possible that adversarial attacks only trivially modified  ReLU activations (i.e. the images were near the decision boundary), causing little change in attribution maps.\footnote{See footnote \ref{footnote:note1}.}

\subsection{Quantitative Comparison with Baseline Methods} \label{section:quantitative}

In this section, we quantitatively compare RectGrad with baseline methods using DNNs trained on CIFAR-10. We did not include Epsilon-LRP since it is equivalent to Gradient $*$ Input for ReLU DNNs \cite{Ancona2017}. We conducted the experiments with final thresholding to the baselines for comparison in similar sparsity setting. \\

\begin{figure}[t]
	\centering
    \includegraphics[width=0.7\linewidth]{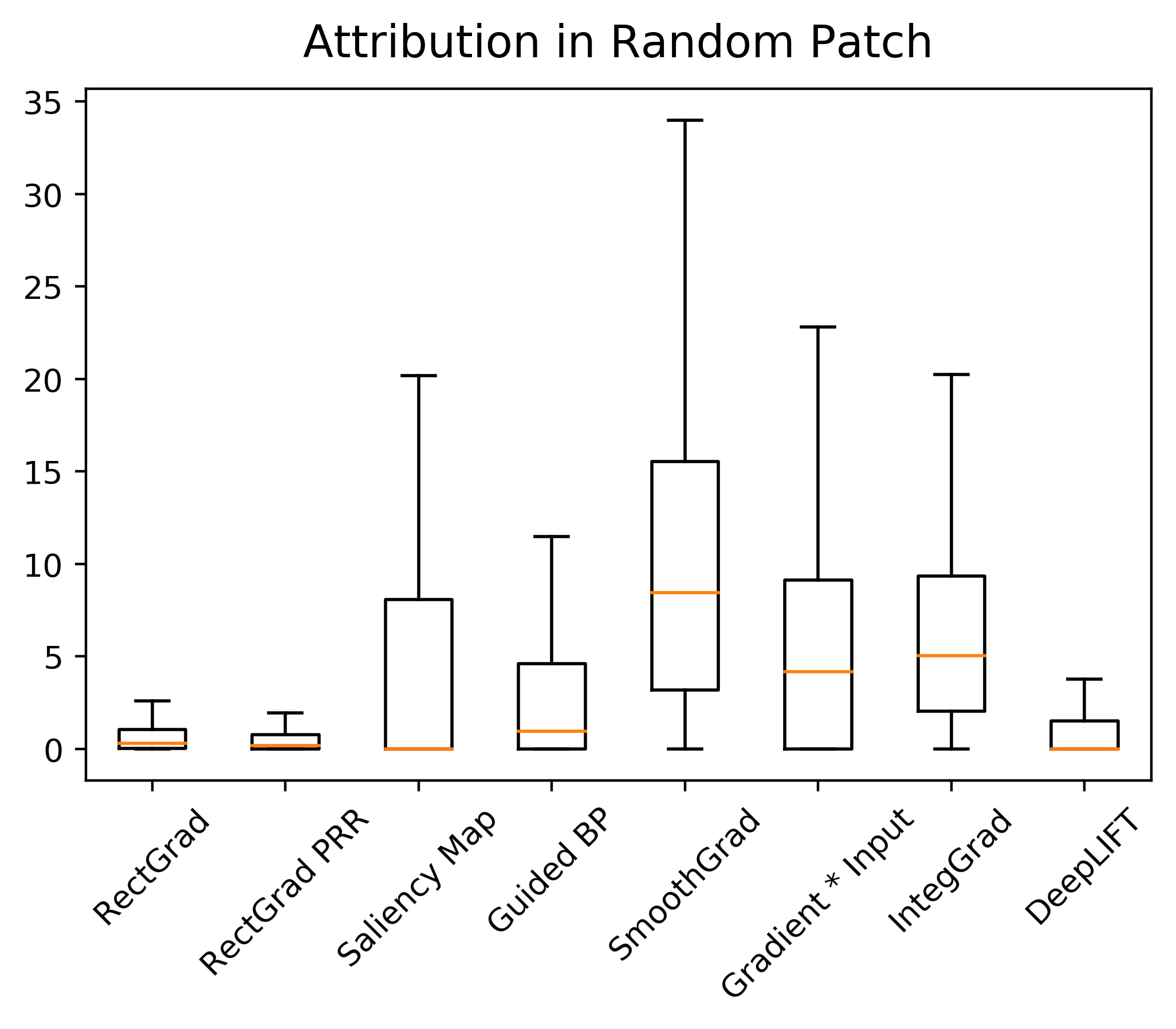}
    \caption{Boxplots of amount of attribution in random patch for attribution maps of images in the test dataset.}
    \label{fig:patch}
\end{figure}

\paragraph{Training Dataset Occlusion} \label{quant: training} Just like the training dataset occlusion experiment in Section \ref{section:explanation}, we occluded the upper left corner of all images in CIFAR-10 training dataset with a 10 $\times$ 10 random patch and trained a randomly initialized CNN on the modified dataset. We used the same patch for all images. We then summed all absolute attribution within the patch. A reasonable attribution method should assign nearly zero attribution to the patch as it is completely irrelevant to the classification task. Figure \ref{fig:patch} compares the amount attribution in the patch between attribution methods. We observed that RectGrad PRR assigned the least attribution to the random patch. RectGrad and DeepLIFT showed similarly good performance while all other methods assigned non-trivial amounts of attribution to the patch. This indicates that baseline methods, with the exception of DeepLIFT, may not be working in a reasonable manner.

\begin{figure}[t]
	\centering
    \includegraphics[width=0.7\linewidth]{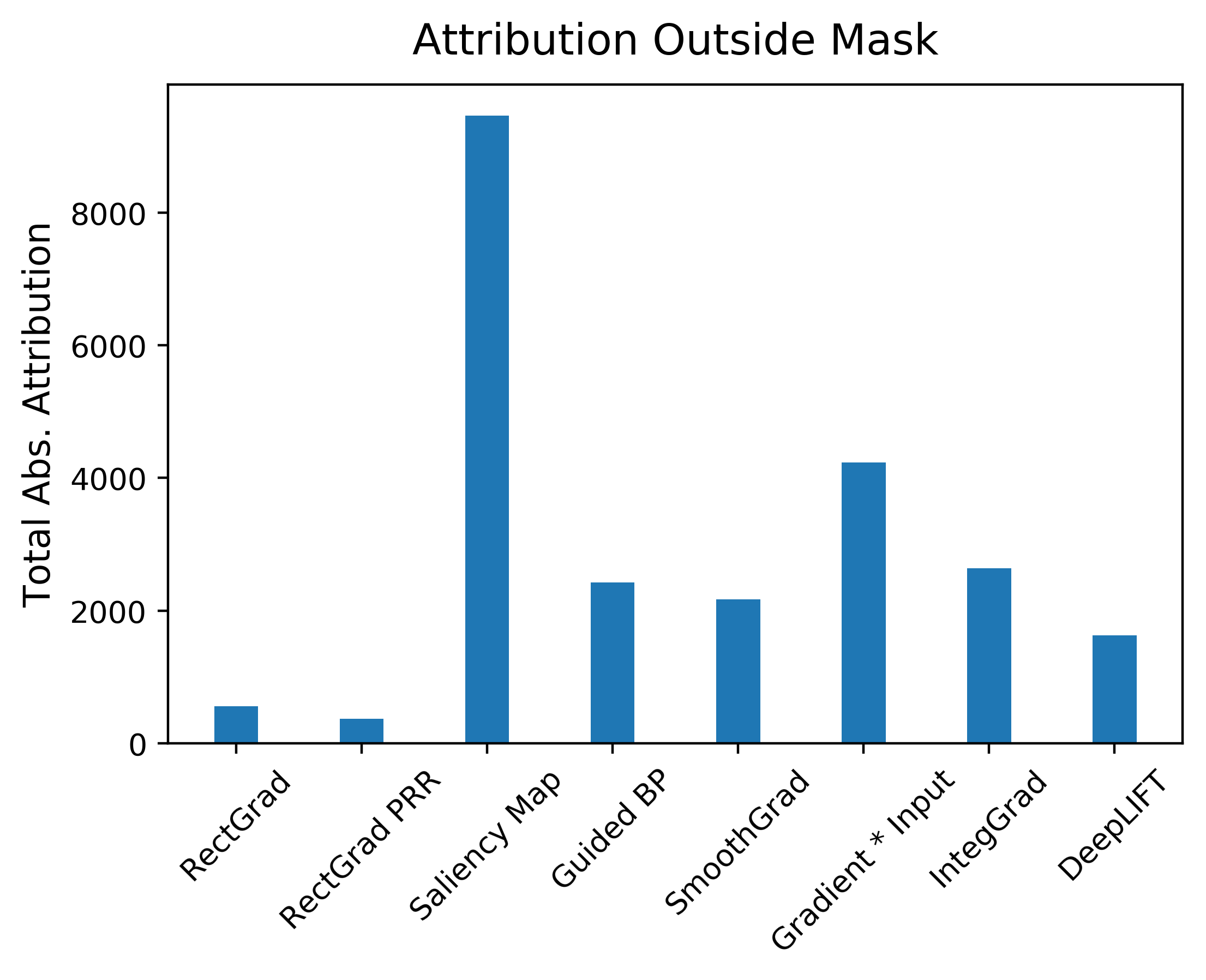}
    \par\medskip
    \includegraphics[width=0.7\linewidth]{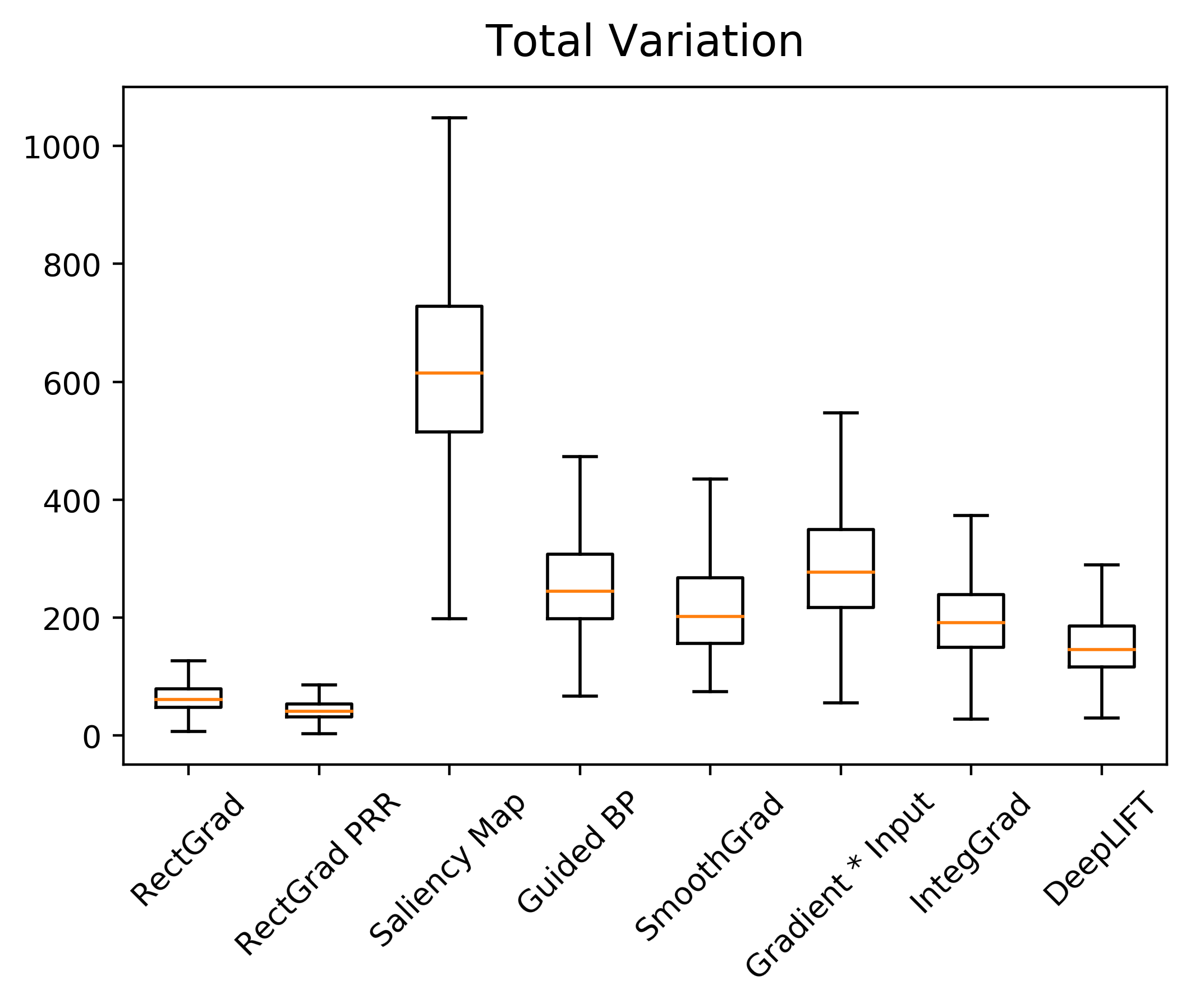}
    \caption{Top: comparison of amount of attribution outside mask (on background). Bottom: boxplots of total variation for attribution maps of images in the test dataset.}
    \label{fig:noise level}
\end{figure}

\paragraph{Noise Level} We evaluated whether RectGrad really reduces noise through two experiments with a CNN trained on CIFAR-10. For the first test, we created segmentation masks for 10 correctly classified images of each class (total 100 images) and measured how much attribution falls on the background. Specifically, we compared the sum of absolute value of attribution on the background. For the second test, we measured the total variation of attribution maps for each attribution method. Figure \ref{fig:noise level} shows the results. We observed that RectGrad outperformed baseline methods in both cases. The results imply that baseline methods cannot replicate RectGrad's ability to reduce noise.

\paragraph{ROAR and KAR} We evaluated RectGrad using Remove and Retrain (ROAR) and Keep and Retrain (KAR) proposed by \cite{Hooker2018}.\footnote{We did not use Sensitivity \cite{Bach2015,Samek2017}. \cite{Hooker2018} has pointed out that without retraining, we do not know whether the degradation in model performance after feature occlusion is due to the replacement value being outside of the training data manifold or due to the accuracy of the attribution. Hence we have used ROAR and KAR which do not suffer from this problem.} Specifically, we measured how the performance of the classifier changed as features were occluded based on the ordering assigned by the attribution method. For ROAR, we replaced a fraction of all CIFAR-10 pixels estimated to be \emph{most} important with a constant value. For KAR, we replaced pixels estimated to be \emph{least} important. We then retrained a CNN on the modified dataset and measured the change in test accuracy. We trained 3 CNNs per estimator for each fraction $\{0.1, 0.3, 0.5, 0.7, 0.9\}$ and measured test accuracy as the average of theses 3 CNNs. An attribution method is better if it has a lower ROAR area under curve (AUC) and KAR area over curve (AOC). We show the results in Figure \ref{fig:roar kar} (in Appendix \ref{section:roar_kar_curves}) and Table \ref{tab:roar kar}. RectGrad outperformed all baseline methods in both ROAR and KAR. The results indicate that RectGrad attributions not only have superior visual quality but also identify important features.
It is especially remarkable that RectGrad largely outperformed the baseline methods in KAR.
This indicates that RectGrad is an appropriate method for discarding attributions which correspond to irrelevant features.
This property corresponds very well with the motivation of RectGrad.

\begin{table}[t]
\vskip 0.15in
\begin{center}
\begin{small}
\begin{sc}
\begin{tabular}{lcccr}
\toprule
Method & ROAR (AUC) & KAR (AOC) \\
\midrule
Random         & 0.5562          & 0.4438 \\
RectGrad       & \textbf{0.5385} & \textbf{0.3382} \\
Saliency Map   & 0.5568          & 0.4455 \\
Guided BP      & 0.5504          & 0.4270 \\
SmoothGrad     & 0.5536          & 0.4444 \\
Grad $*$ Input & 0.5408          & 0.4176 \\
IntegGrad      & 0.5428          & 0.4185 \\
DeepLIFT       & 0.5519          & 0.3928 \\
\bottomrule
\end{tabular}
\end{sc}
\end{small}
\end{center}
\vskip -0.1in
\caption{Comparison of ROAR AUCs and KAR AOCs. An attribution method is better if it has a lower score. The best scores are written in bold.}
\label{tab:roar kar}
\end{table}

\section{Conclusions}

Saliency Map is the most basic technique for interpreting deep neural network decisions. However, it is often visually noisy. Although several hypotheses were proposed to account for this phenomenon, there is few work that provides a thorough analysis of noisy saliency maps. Therefore, we identified saliency maps are noisy because DNNs do not filter out irrelevant features during forward propagation. We then proposed \textit{Rectified Gradient} which solves this problem through layer-wise thresholding during backpropagation. We showed that Rectified Gradient generalizes Deconvolution and Guided Backpropagation and moreover, overcomes the class-insensitivity problem. We also demonstrated through qualitative experiments that Rectified Gradient, unlike other attribution methods, produces visually sharp and coherent attribution maps. Finally, we verified with quantitative experiments that Rectified Gradient not only removes noise from attribution maps, but also outperforms other methods at highlighting important features.

{\small
\bibliographystyle{ieee}
\bibliography{egbib}

\begin{thebibliography}{10}\itemsep=-1pt

\bibitem{tensorflow2015-whitepaper}
M.~Abadi, P.~Barham, J.~Chen, Z.~Chen, A.~Davis, J.~Dean, M.~Devin,
  S.~Ghemawat, G.~Irving, M.~Isard, M.~Kudlur, J.~Levenberg, R.~Monga,
  S.~Moore, D.~G. Murray, B.~Steiner, P.~Tucker, V.~Vasudevan, P.~Warden,
  M.~Wicke, Y.~Yu, and X.~Zheng.
\newblock Tensorflow: A system for large-scale machine learning.
\newblock In {\em Proceedings of the 12th USENIX Conference on Operating
  Systems Design and Implementation}, OSDI'16, pages 265--283, Berkeley, CA,
  USA, 2016. USENIX Association.

\bibitem{Ancona2017}
M.~Ancona, E.~Ceolini, C.~{\"O}ztireli, and M.~Gross.
\newblock Towards better understanding of gradient-based attribution methods
  for deep neural networks.
\newblock In {\em International Conference on Learning Representations}, 2018.

\bibitem{Bach2015}
S.~Bach, A.~Binder, G.~Montavon, F.~Klauschen, K.~R. M{\"{u}}ller, and
  W.~Samek.
\newblock {On pixel-wise explanations for non-linear classifier decisions by
  layer-wise relevance propagation}.
\newblock {\em PLoS ONE}, 10(7):1--46, 2015.

\bibitem{Baehrens2009}
D.~Baehrens, T.~Schroeter, S.~Harmeling, M.~Kawanabe, K.~Hansen, and K.-R.
  M{\"u}ller.
\newblock How to explain individual classification decisions.
\newblock {\em Journal of Machine Learning Research}, 11(Jun):1803--1831, 2010.

\bibitem{Erhan2009}
D.~Erhan, Y.~Bengio, A.~Courville, and P.~Vincent.
\newblock Visualizing higher-layer features of a deep network.
\newblock {\em University of Montreal 1341.3}, 2009.

\bibitem{Goodfellow2014}
I.~J. Goodfellow, J.~Shlens, and C.~Szegedy.
\newblock Striving for simplicity: The all convolutional net.
\newblock In {\em International Conference on Learning Representations}, 2015.

\bibitem{Hooker2018}
S.~Hooker, D.~Erhan, P.-J. Kindermans, and B.~Kim.
\newblock Evaluating feature importance estimates.
\newblock In {\em ICML Workshop on Human Interpretability in Machine Learning},
  2018.

\bibitem{kim2019safeml}
B.~Kim, J.~Seo, and T.~Jeon.
\newblock Bridging adversarial robustness and gradient interpretability.
\newblock {\em Safe Machine Learning workshop at {ICLR}}, 2019.

\bibitem{cifar2009}
A.~Krizhevsky and G.~Hinton.
\newblock Learning multiple layers of features from tiny images.
\newblock 2009.

\bibitem{Montavon2015}
G.~Montavon, S.~Lapuschkin, A.~Binder, W.~Samek, and K.-R. M{\"u}ller.
\newblock Explaining nonlinear classification decisions with deep taylor
  decomposition.
\newblock {\em Pattern Recognition}, 65:211--222, 2017.

\bibitem{morcos2018importance}
A.~S. Morcos, D.~G. Barrett, N.~C. Rabinowitz, and M.~Botvinick.
\newblock On the importance of single directions for generalization.
\newblock In {\em International Conference on Learning Representations}, 2018.

\bibitem{Nie2018}
W.~Nie, Y.~Zhang, and A.~Patel.
\newblock A theoretical explanation for perplexing behaviors of
  backpropagation-based visualizations.
\newblock In {\em International Conference on Machine Learning}, 2018.

\bibitem{paszke2017automatic}
A.~Paszke, S.~Gross, S.~Chintala, G.~Chanan, E.~Yang, Z.~DeVito, Z.~Lin,
  A.~Desmaison, L.~Antiga, and A.~Lerer.
\newblock Automatic differentiation in pytorch.
\newblock In {\em NIPS Workshop on Autodiff}, 2017.

\bibitem{ross2018improving}
A.~S. Ross and F.~Doshi-Velez.
\newblock Improving the adversarial robustness and interpretability of deep
  neural networks by regularizing their input gradients.
\newblock In {\em AAAI Conference on Artificial Intelligence}, 2018.

\bibitem{Imagenet2015}
O.~Russakovsky, J.~Deng, H.~Su, J.~Krause, S.~Satheesh, S.~Ma, Z.~Huang,
  A.~Karpathy, A.~Khosla, M.~Bernstein, et~al.
\newblock Imagenet large scale visual recognition challenge.
\newblock {\em International Journal of Computer Vision}, 115(3):211--252,
  2015.

\bibitem{Samek2017}
W.~Samek, A.~Binder, G.~Montavon, S.~Lapuschkin, and K.-R. M{\"u}ller.
\newblock Evaluating the visualization of what a deep neural network has
  learned.
\newblock {\em IEEE transactions on neural networks and learning systems},
  28(11):2660--2673, 2017.

\bibitem{Selvaraju2016}
R.~R. Selvaraju, M.~Cogswell, A.~Das, R.~Vedantam, D.~Parikh, and D.~Batra.
\newblock Grad-cam: Visual explanations from deep networks via gradient-based
  localization.
\newblock In {\em The IEEE International Conference on Computer Vision}, Oct
  2017.

\bibitem{Shrikumar2017}
A.~Shrikumar, P.~Greenside, and A.~Kundaje.
\newblock Learning important features through propagating activation
  differences.
\newblock In {\em International Conference on Machine Learning}, 2017.

\bibitem{Simonyan2013}
K.~Simonyan, A.~Vedaldi, and A.~Zisserman.
\newblock Deep inside convolutional networks: Visualising image classification
  models and saliency maps.
\newblock In {\em International Conference on Learning Representations
  Workshop}, 2014.

\bibitem{Smilkov2017}
D.~Smilkov, N.~Thorat, B.~Kim, F.~Vi{\'e}gas, and M.~Wattenberg.
\newblock Smoothgrad: removing noise by adding noise.
\newblock In {\em ICML Workshop on Visualization for Deep Learning}, 2017.

\bibitem{Springenberg2014}
J.~T. Springenberg, A.~Dosovitskiy, T.~Brox, and M.~Riedmiller.
\newblock Striving for simplicity: The all convolutional net.
\newblock {\em International Conference on Learning Representations Workshop},
  2015.

\bibitem{Sundararajan2016}
M.~Sundararajan, A.~Taly, and Q.~Yan.
\newblock Gradients of counterfactuals.
\newblock {\em arXiv preprint arXiv:1611.02639}, 2016.

\bibitem{Sundararajan2017}
M.~Sundararajan, A.~Taly, and Q.~Yan.
\newblock Axiomatic attribution for deep networks.
\newblock In {\em International Conference on Machine Learning}, 2017.

\bibitem{Szegedy2016}
C.~Szegedy, S.~Ioffe, V.~Vanhoucke, and A.~A. Alemi.
\newblock Inception-v4, inception-resnet and the impact of residual connections
  on learning.
\newblock In {\em AAAI Conference on Artificial Intelligence}, 2017.

\bibitem{tsipras2019there}
D.~Tsipras, S.~Santurkar, L.~Engstrom, A.~Turner, and A.~Madry.
\newblock Robustness may be at odds with accuracy.
\newblock In {\em International Conference on Learning Representations}, 2019.

\bibitem{Zeiler2014}
M.~D. Zeiler and R.~Fergus.
\newblock Visualizing and understanding convolutional networks.
\newblock In {\em European Conference on Computer Vision}, pages 818--833.
  Springer, 2014.

\bibitem{zhang2019theoretically}
H.~Zhang, Y.~Yu, J.~Jiao, E.~P. Xing, L.~E. Ghaoui, and M.~I. Jordan.
\newblock Theoretically principled trade-off between robustness and accuracy.
\newblock In {\em International Conference on Machine Learning}, 2019.

\end{thebibliography}
}

\onecolumn
\appendix

\section{Experiment Results} \label{section:figures}

\subsection{Feature Map Visualization} \label{section:feature visualization}

\begin{figure}[H]
	\centering
    \begin{subfigure}[b]{\linewidth}
    \centering
    \includegraphics[width=0.4\linewidth]{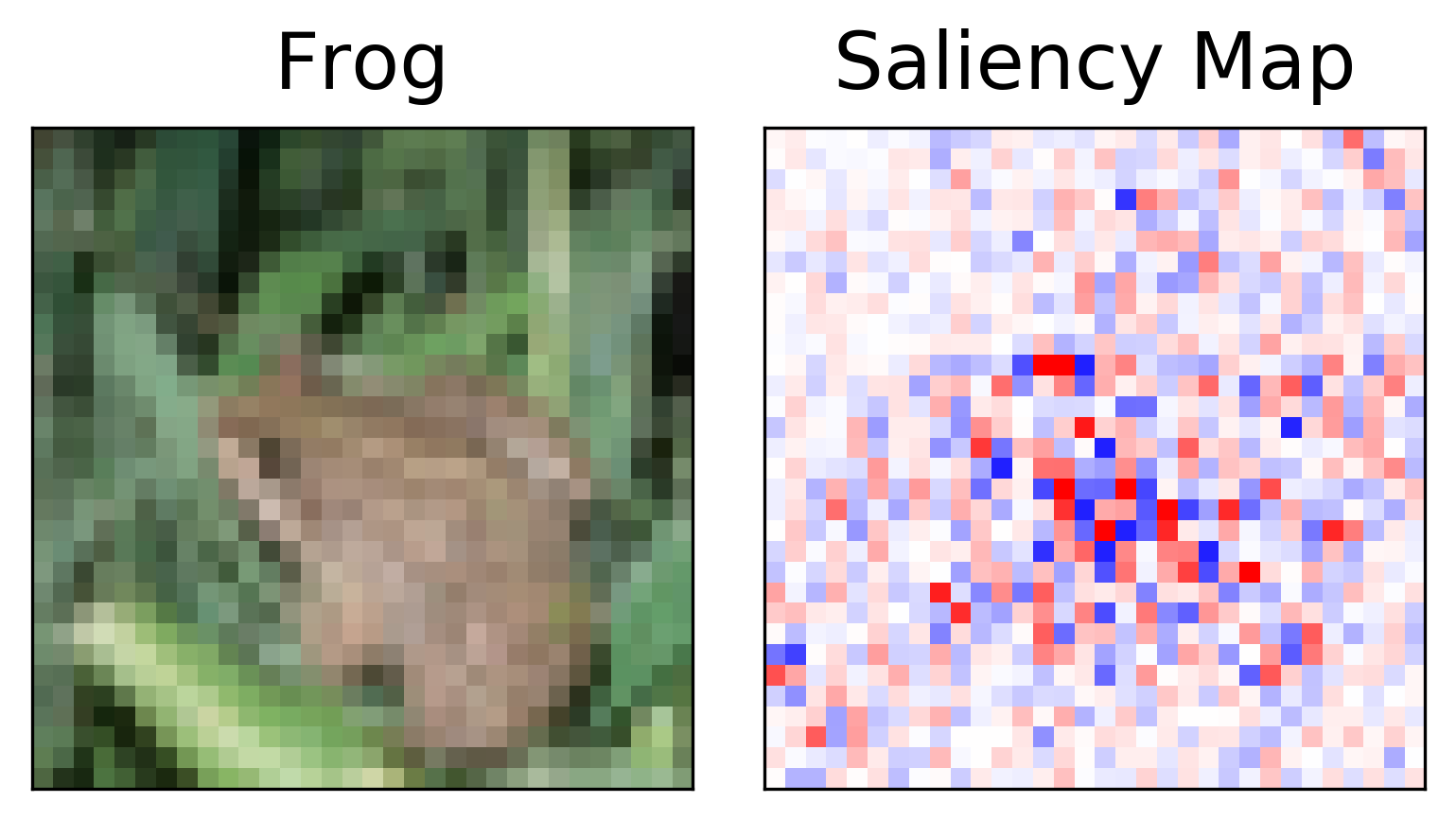}
    \caption{Sample image and its saliency map.}
    \label{fig:feature map (a)}
    \end{subfigure} \\ \vspace{1em}
	\begin{subfigure}[b]{0.8\linewidth}
    \includegraphics[width=0.48\linewidth]{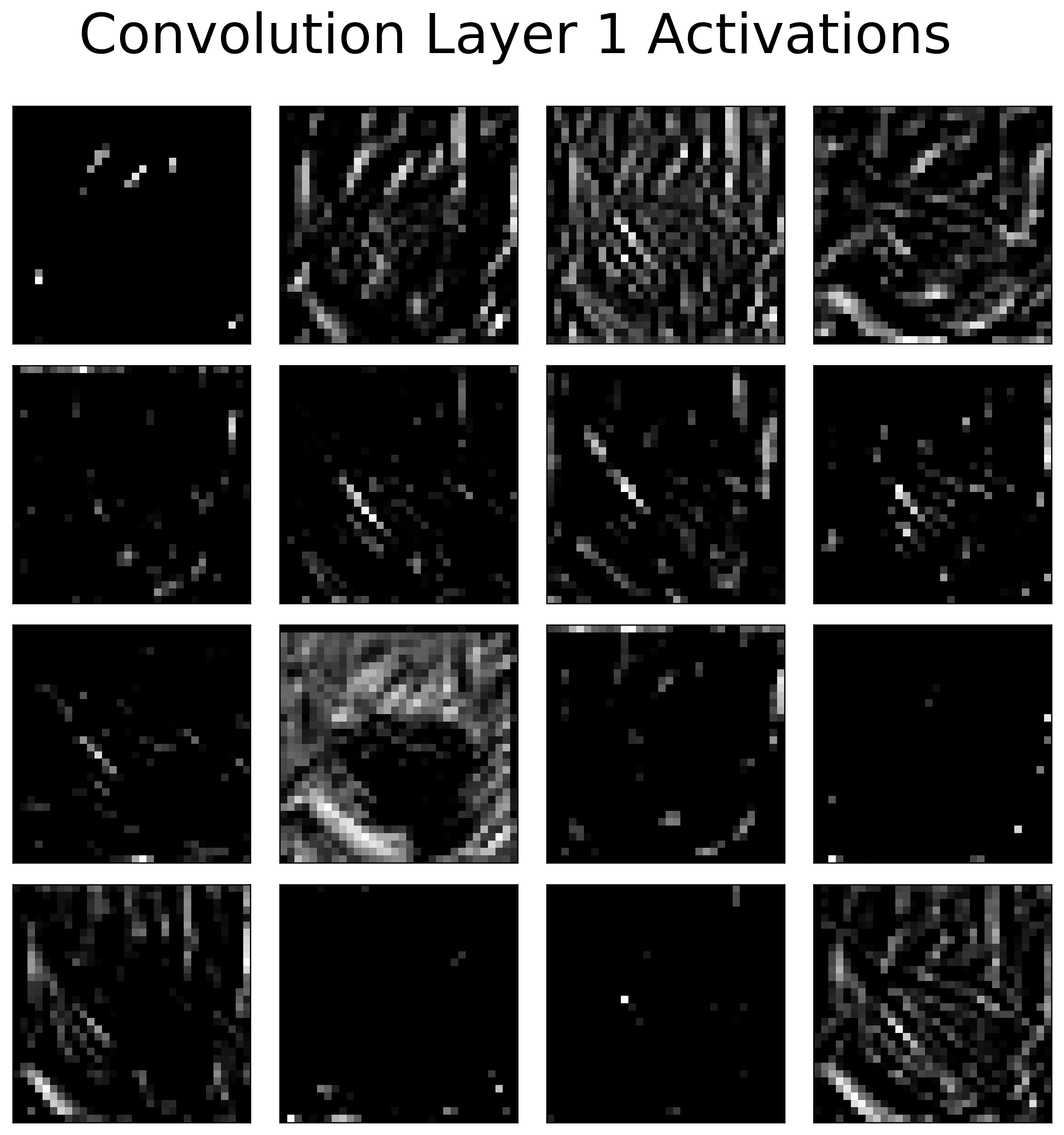}\hfill
    \includegraphics[width=0.48\linewidth]{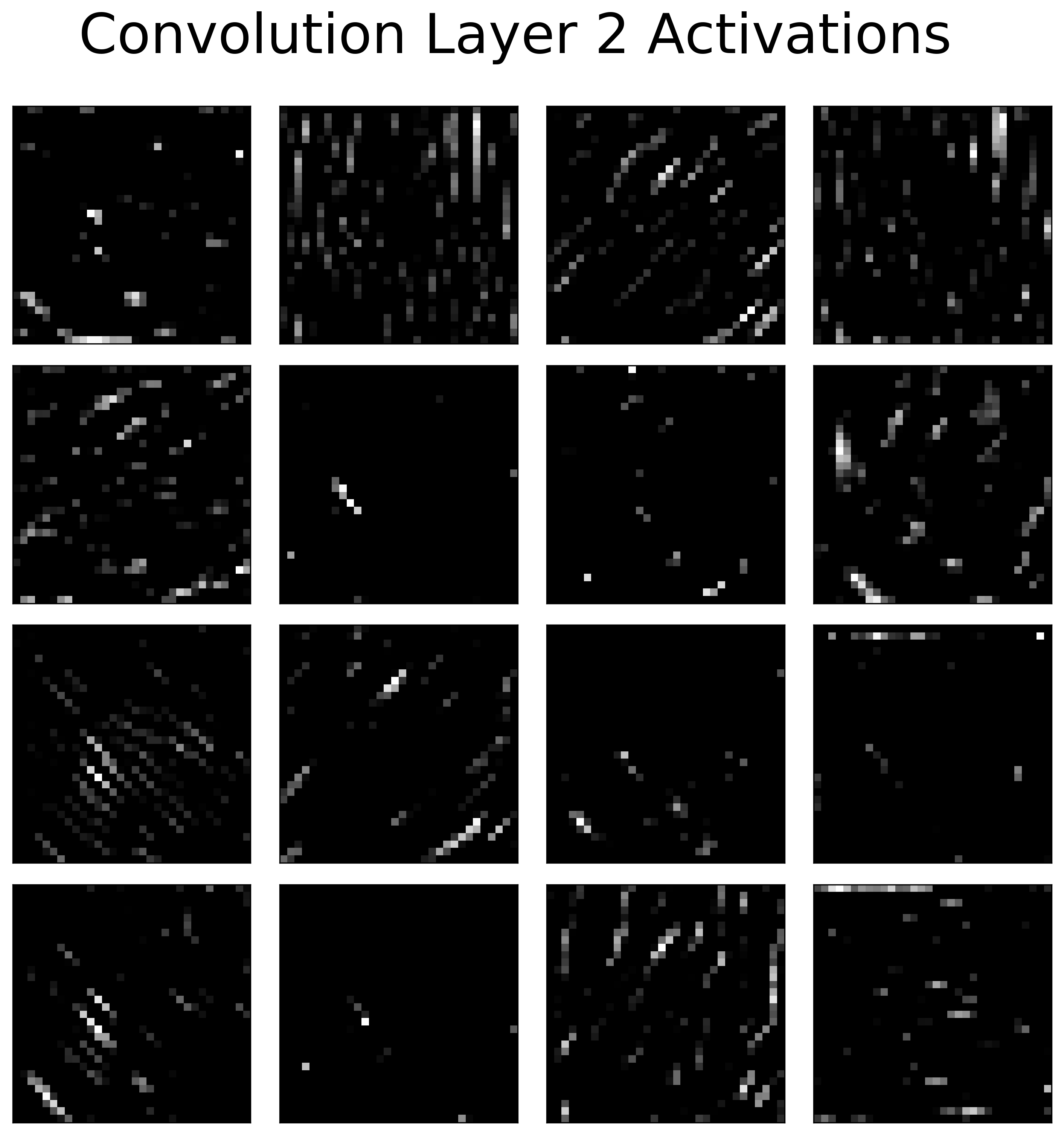}
    \par\medskip
    \includegraphics[width=0.48\linewidth]{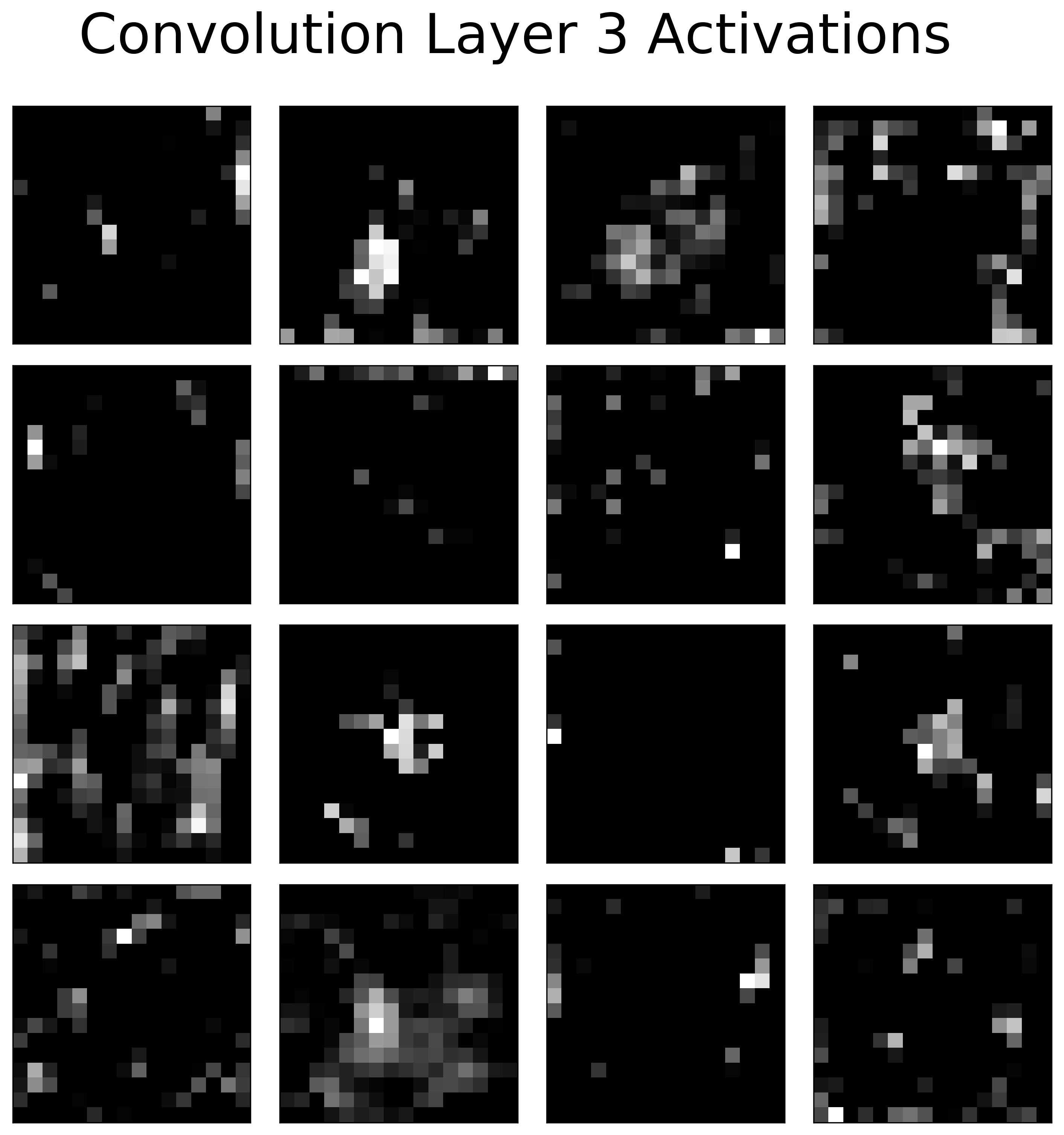}\hfill
    \includegraphics[width=0.48\linewidth]{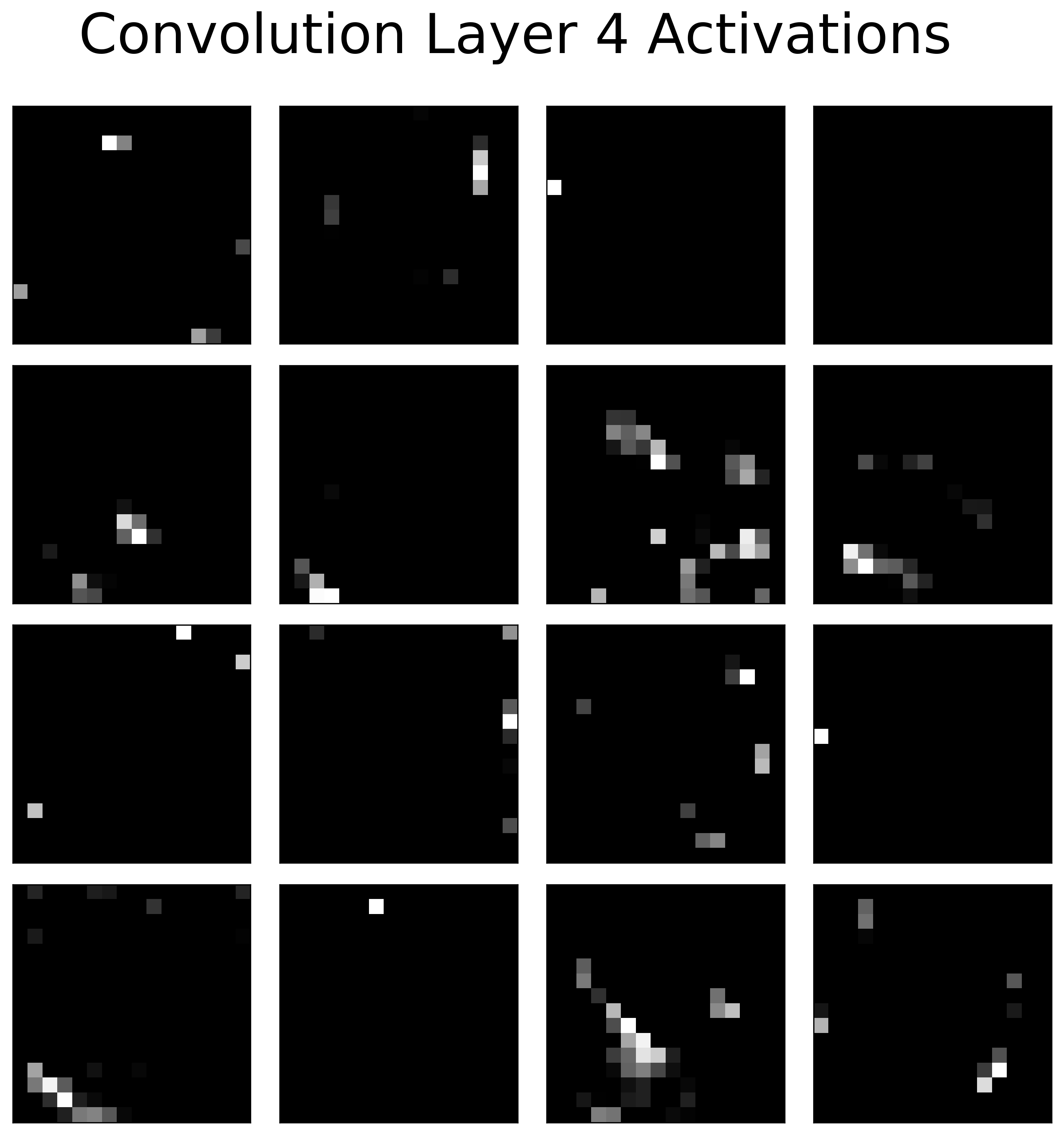}
    \caption{Intermediate layer activations.}
    \label{fig:feature map (b)}
    \end{subfigure}
\end{figure}

\newpage

\begin{figure}[H]
	\centering
    \begin{subfigure}[b]{\linewidth}
    \centering
    \includegraphics[width=0.4\linewidth]{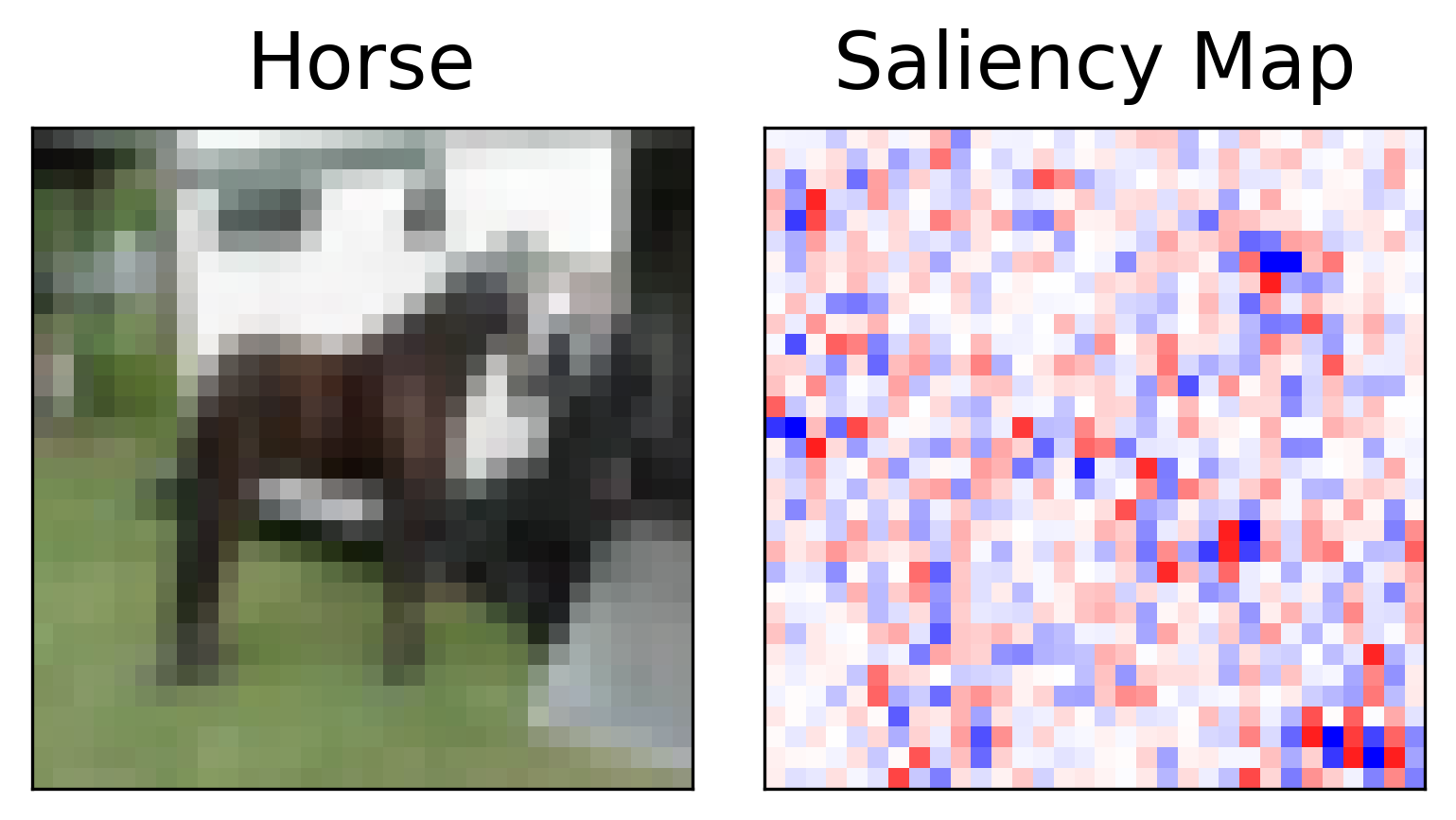}
    \caption{Sample image and its saliency map.}
    \label{fig:feature map (a)}
    \end{subfigure} \\ \vspace{1em}
	\begin{subfigure}[b]{0.8\linewidth}
    \includegraphics[width=0.48\linewidth]{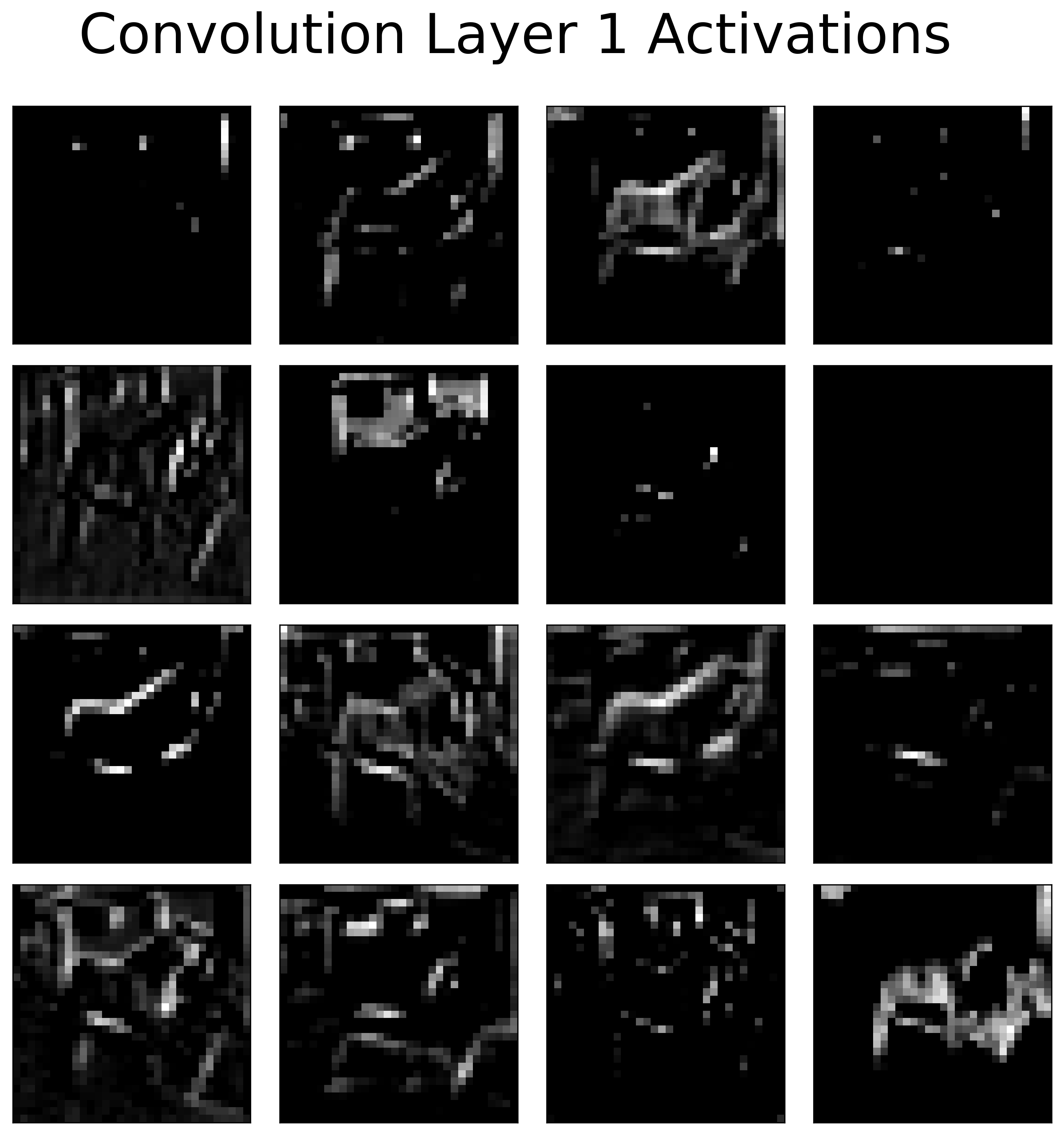}\hfill
    \includegraphics[width=0.48\linewidth]{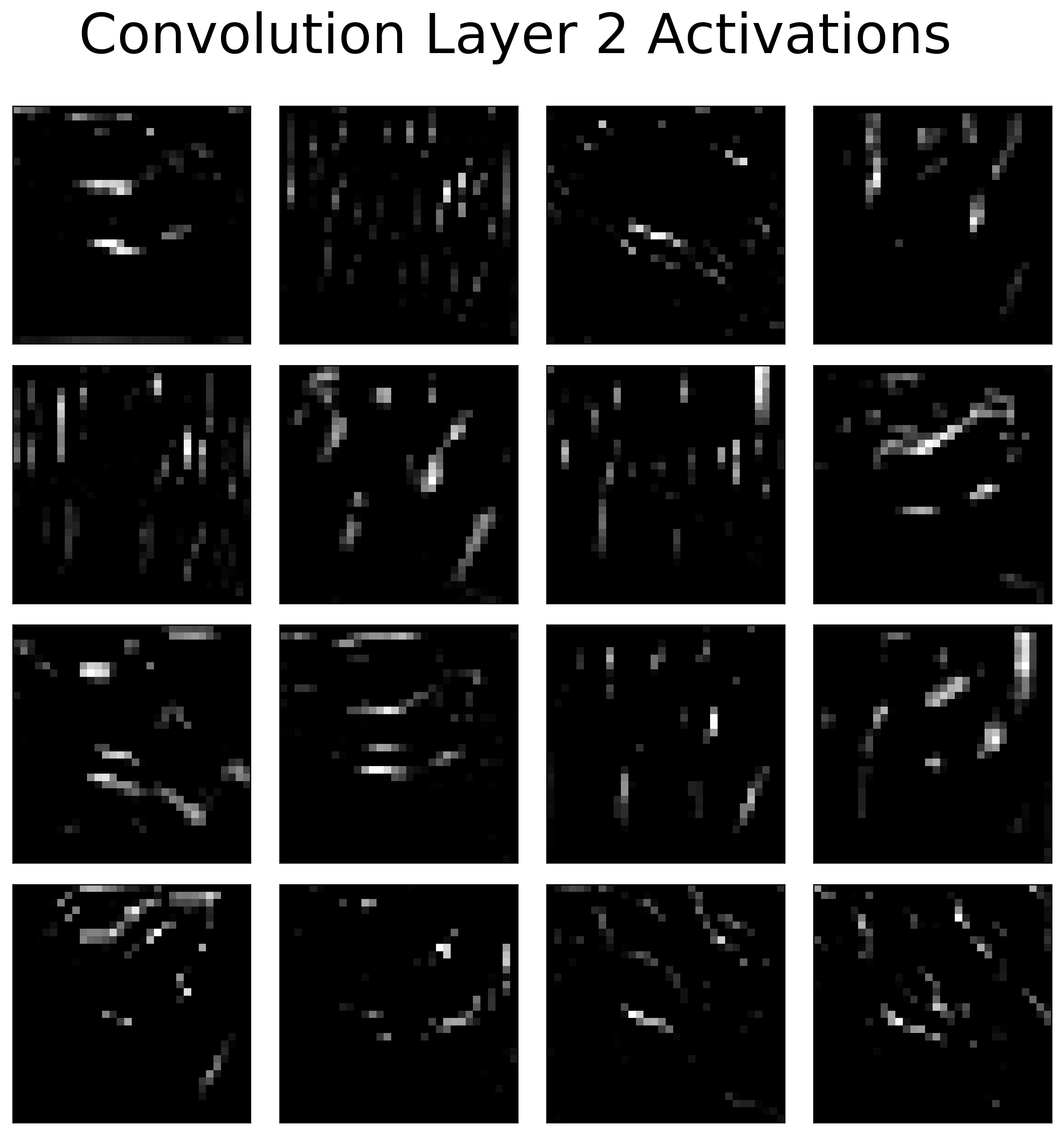}
    \par\medskip
    \includegraphics[width=0.48\linewidth]{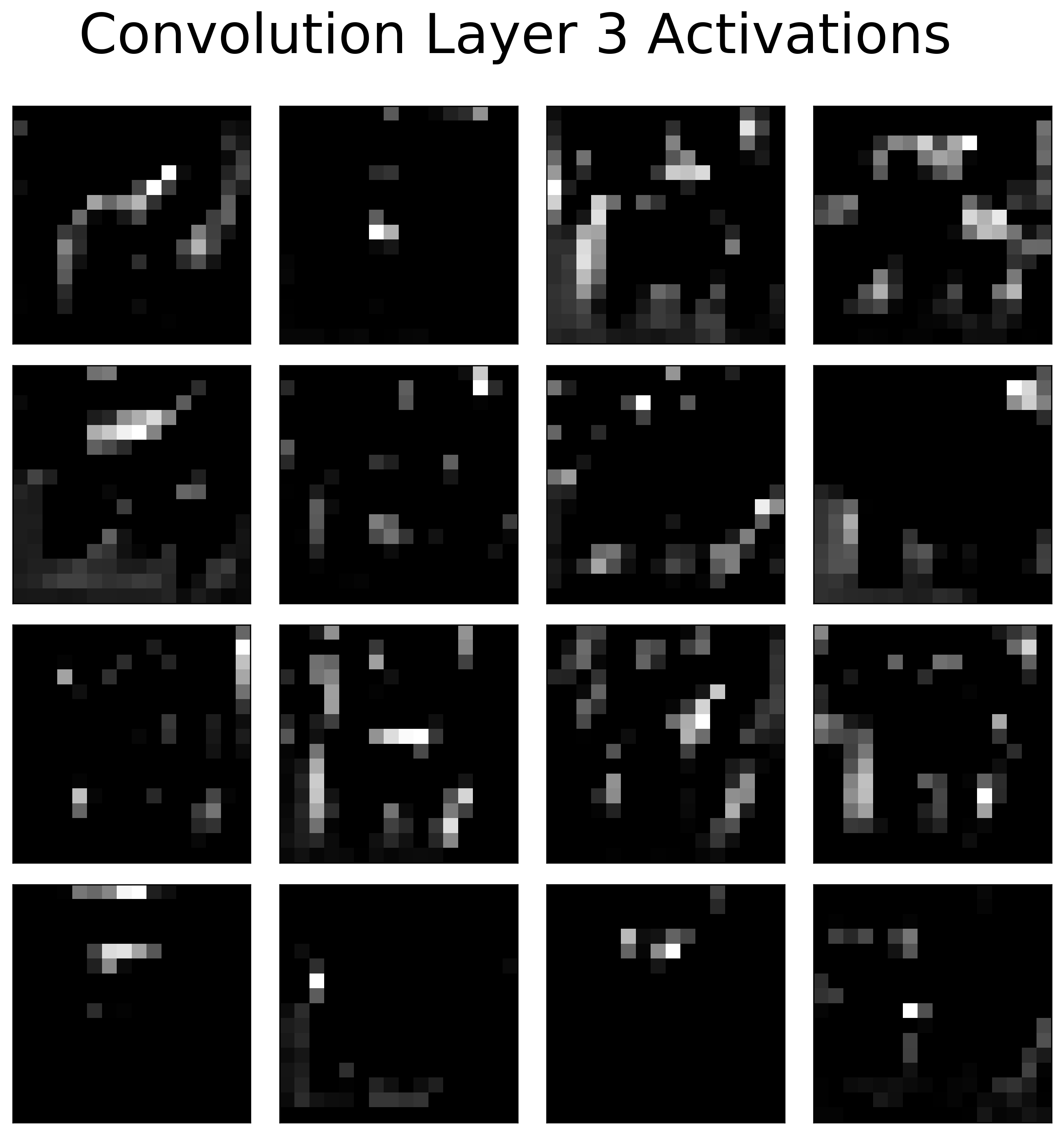}\hfill
    \includegraphics[width=0.48\linewidth]{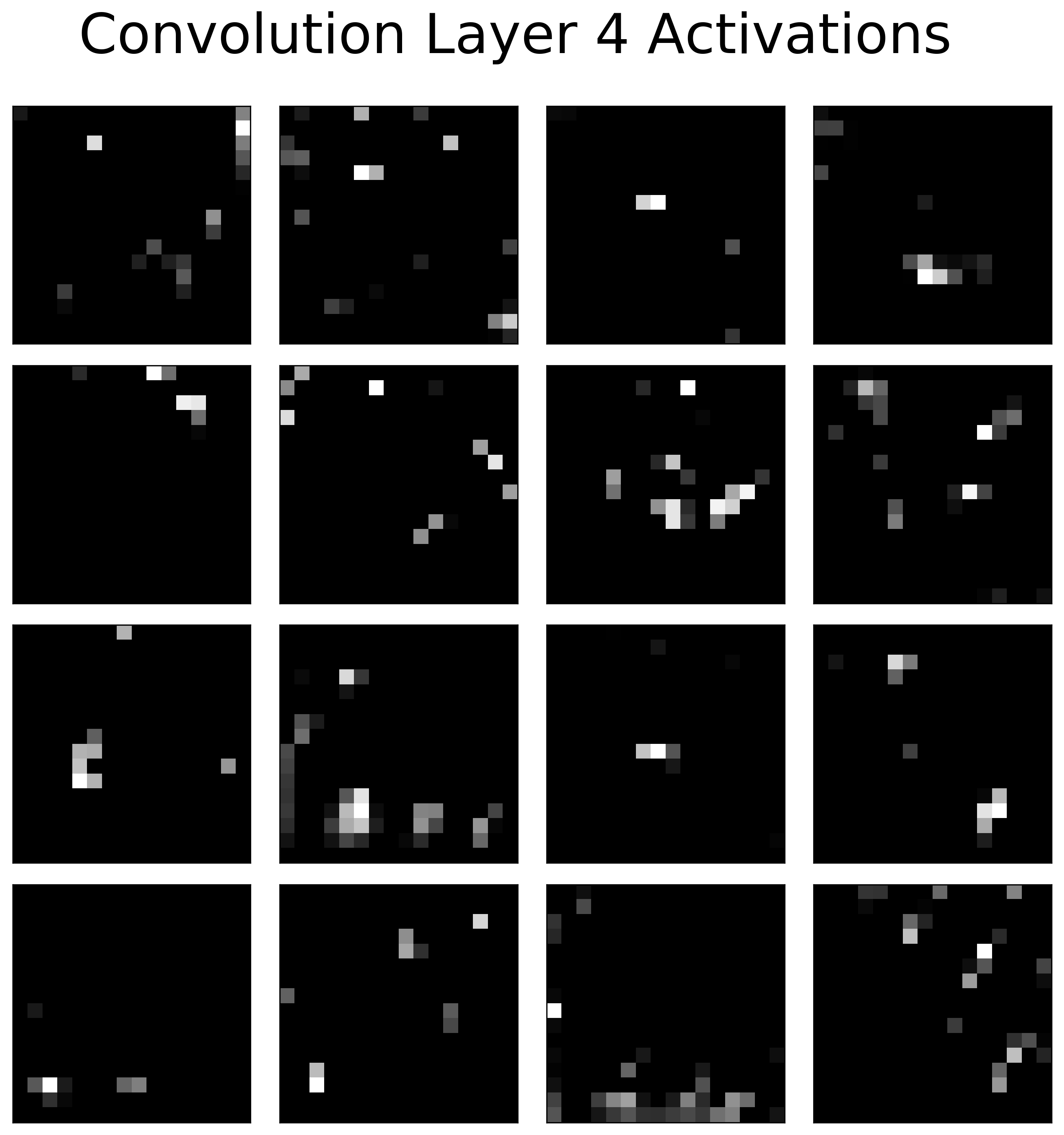}
    \caption{Intermediate layer activations.}
    \label{fig:feature map (b)}
    \end{subfigure}
\end{figure}

\newpage

\begin{figure}[H]
	\centering
    \begin{subfigure}[b]{\linewidth}
    \centering
    \includegraphics[width=0.4\linewidth]{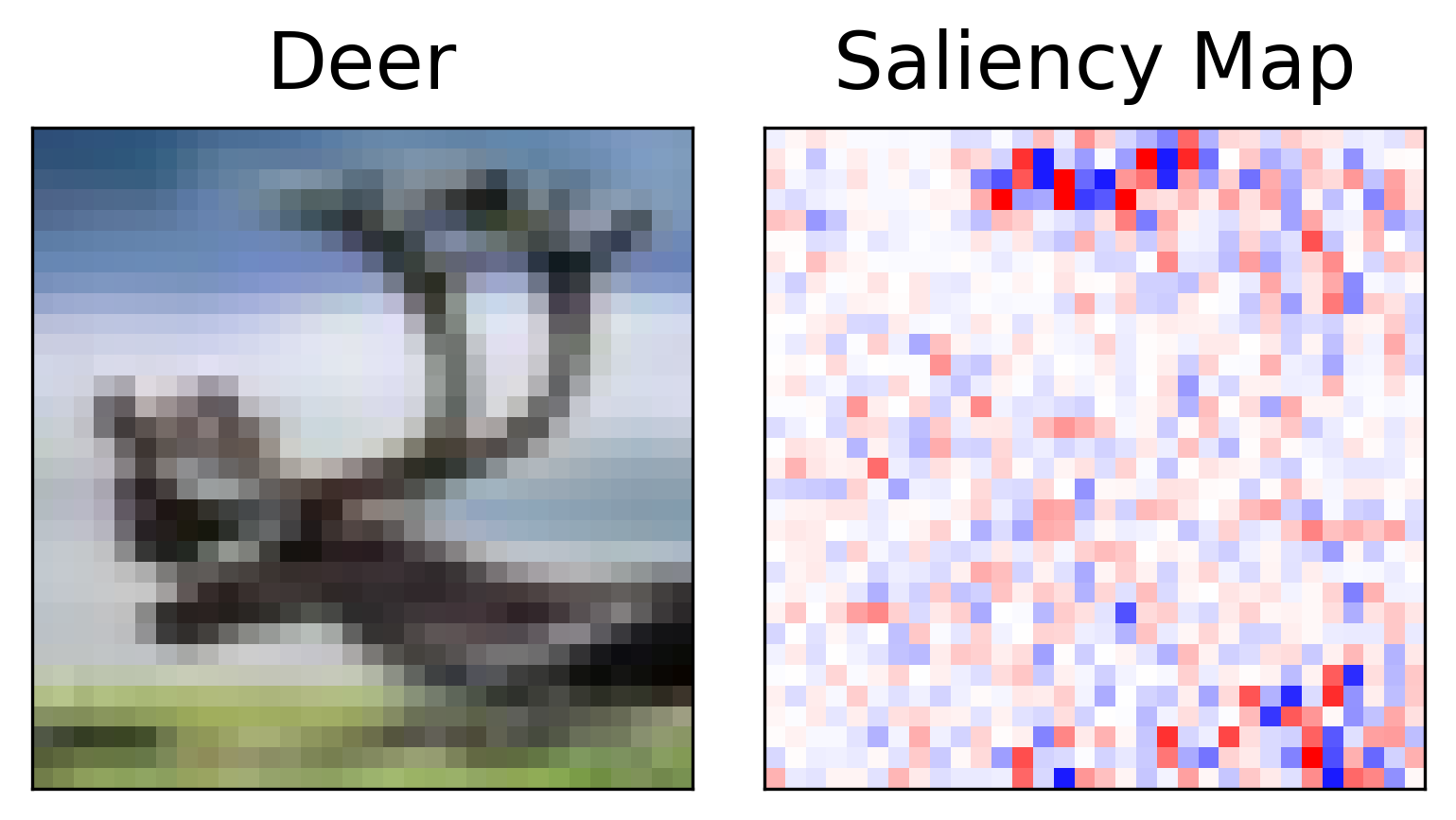}
    \caption{Sample image and its saliency map.}
    \label{fig:feature map (a)}
    \end{subfigure} \\ \vspace{1em}
	\begin{subfigure}[b]{0.8\linewidth}
    \includegraphics[width=0.48\linewidth]{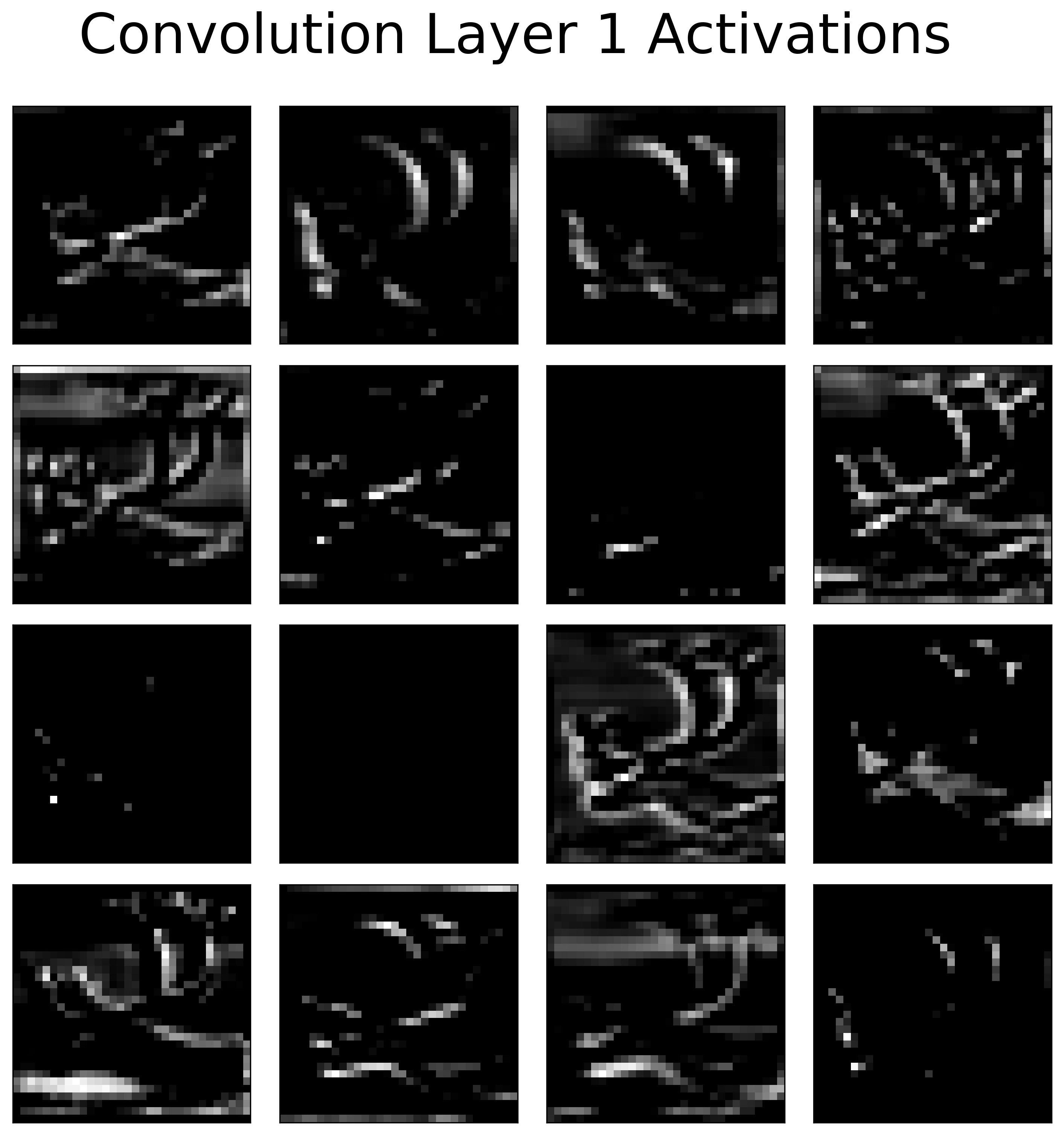}\hfill
    \includegraphics[width=0.48\linewidth]{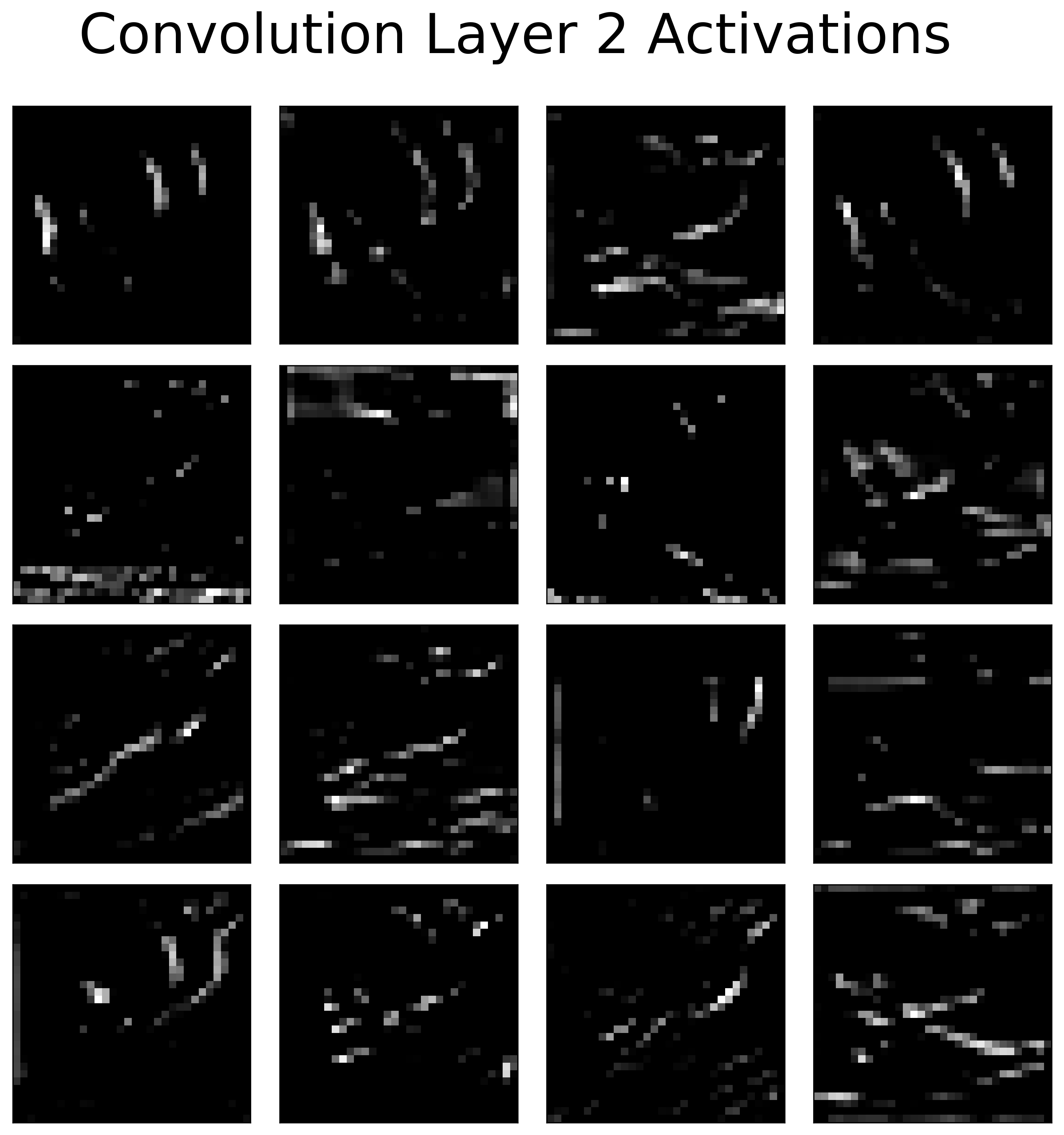}
    \par\medskip
    \includegraphics[width=0.48\linewidth]{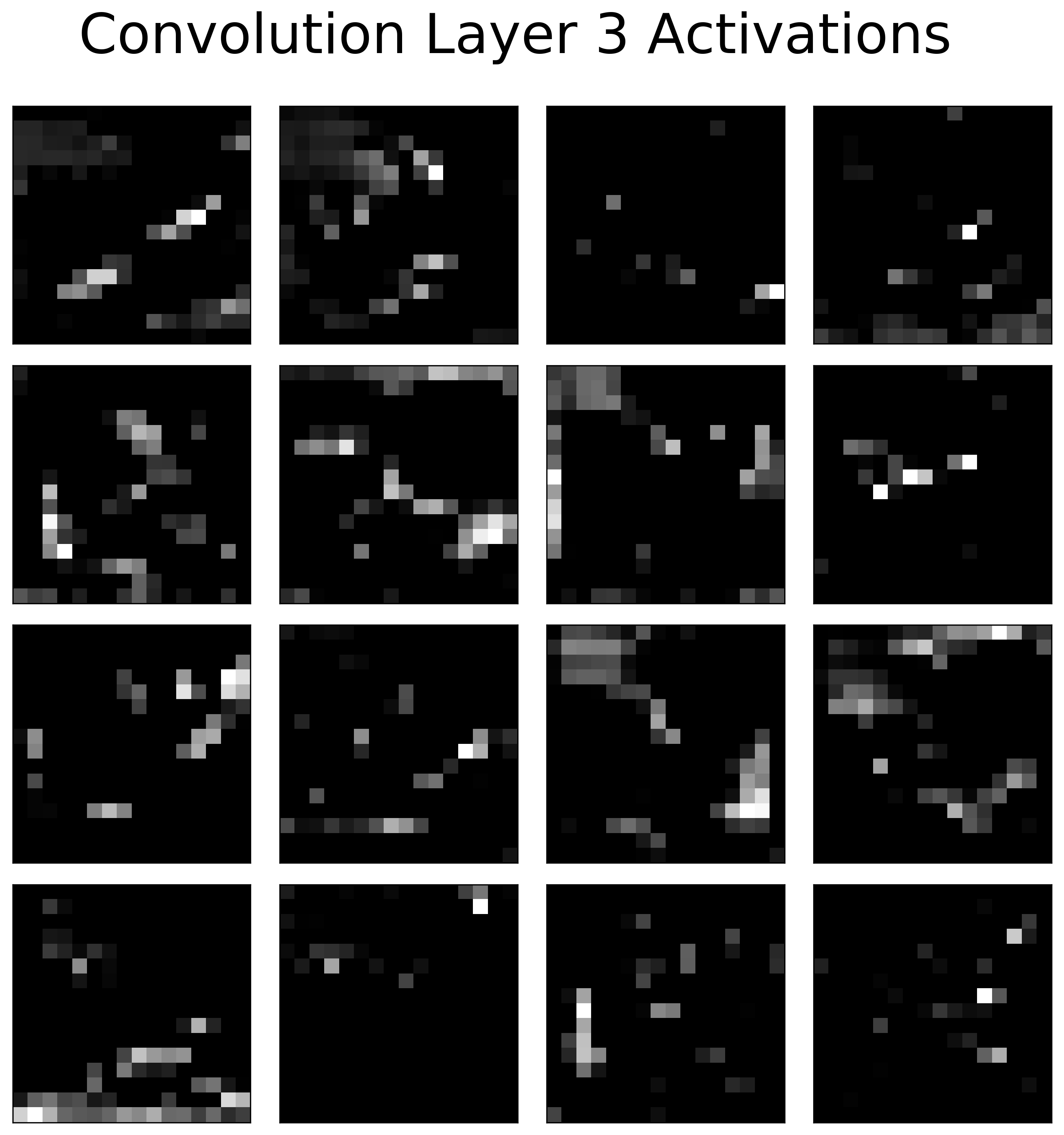}\hfill
    \includegraphics[width=0.48\linewidth]{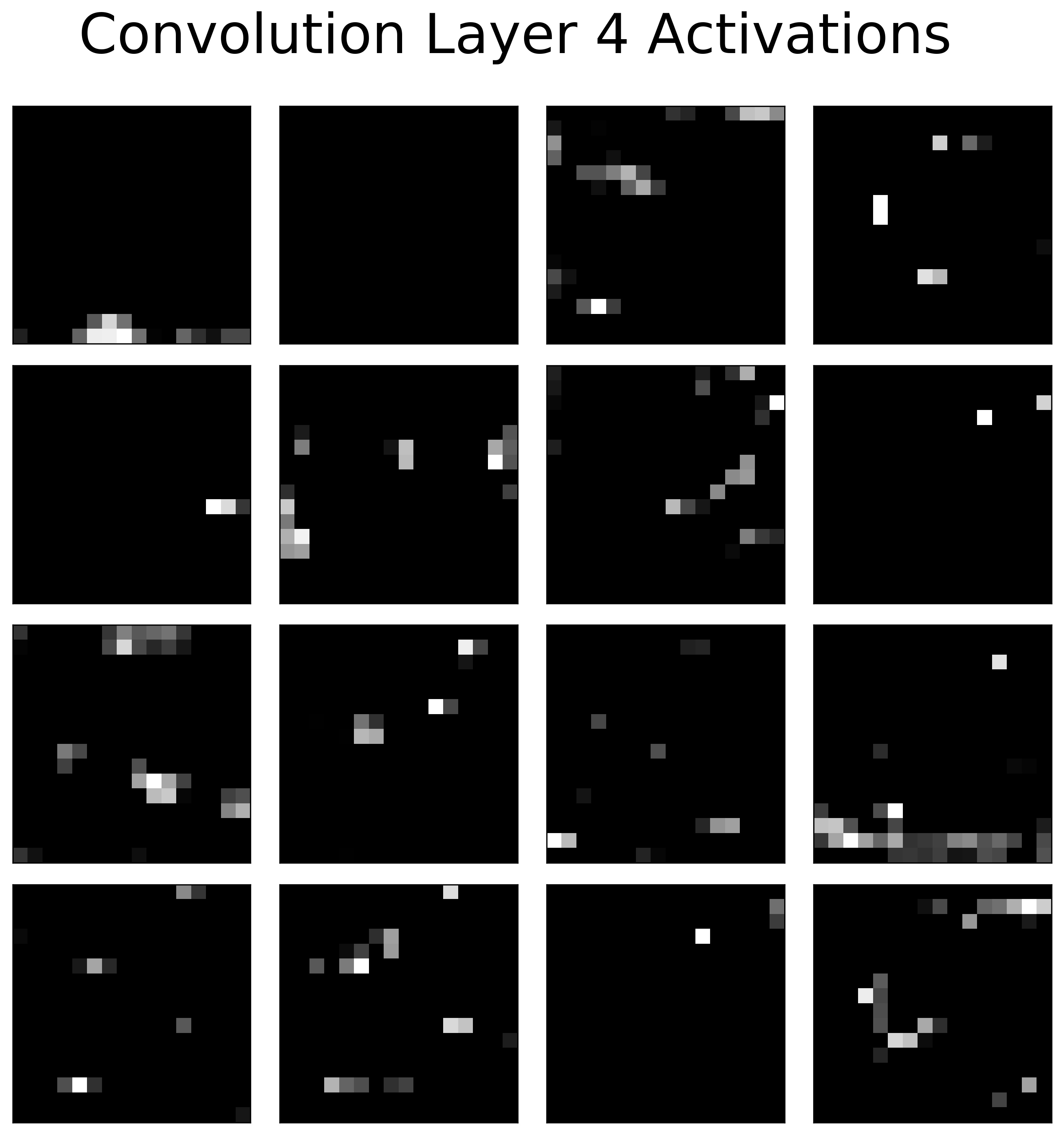}
    \caption{Intermediate layer activations.}
    \label{fig:feature map (b)}
    \end{subfigure}
\end{figure}

\newpage

\begin{figure}[H]
	\centering
    \begin{subfigure}[b]{\linewidth}
    \centering
    \includegraphics[width=0.4\linewidth]{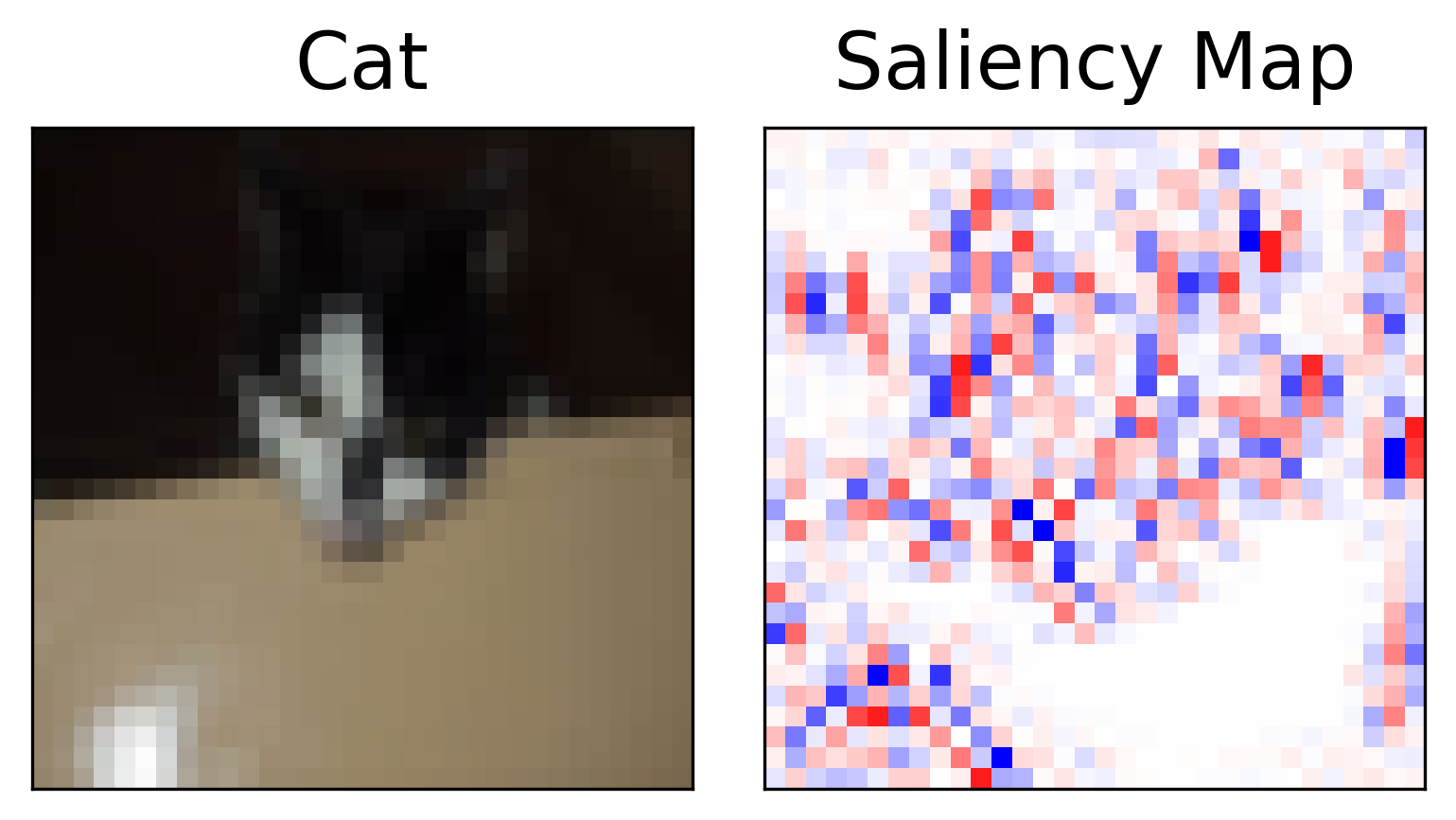}
    \caption{Sample image and its saliency map.}
    \label{fig:feature map (a)}
    \end{subfigure} \\ \vspace{1em}
	\begin{subfigure}[b]{0.8\linewidth}
    \includegraphics[width=0.48\linewidth]{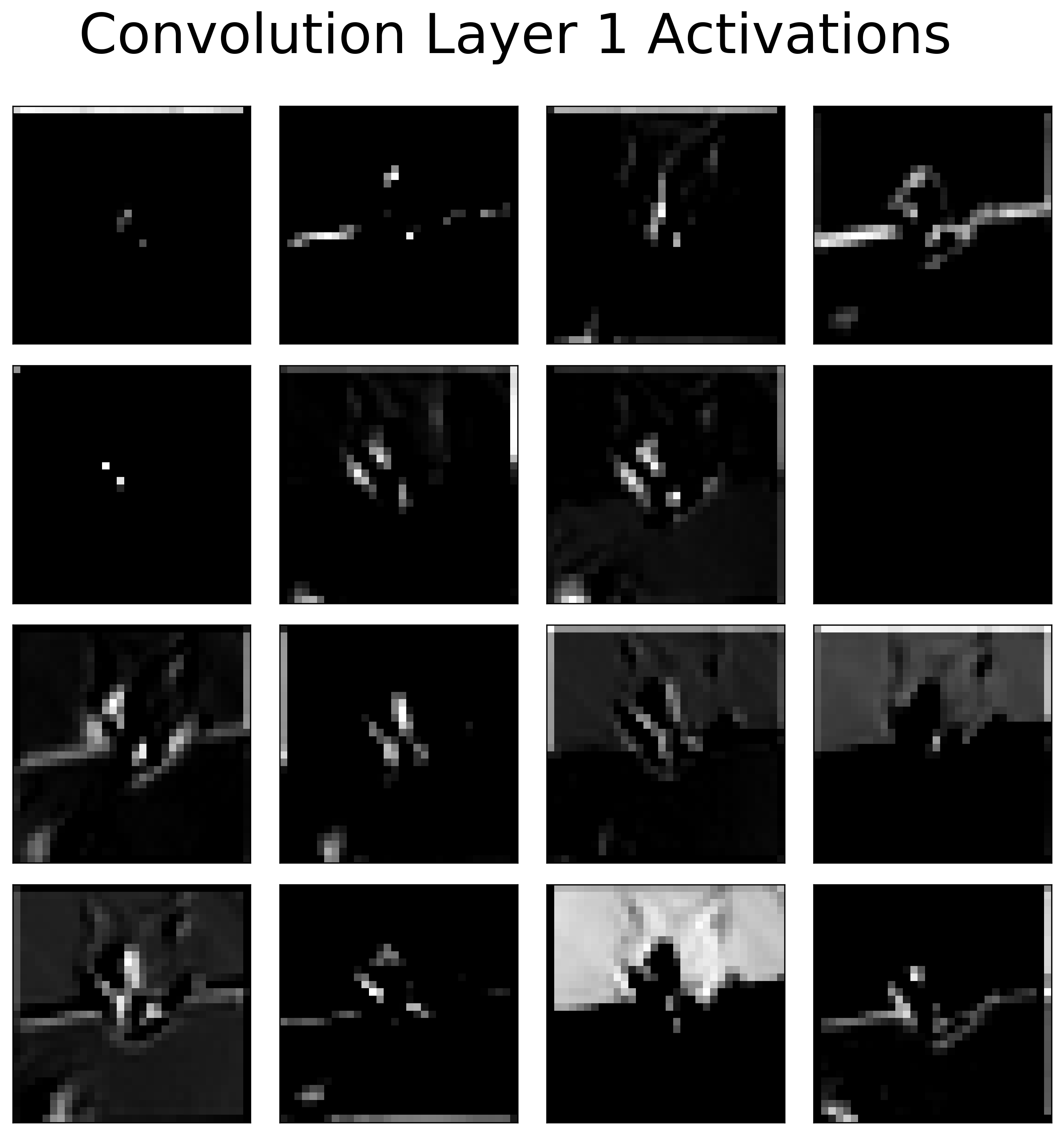}\hfill
    \includegraphics[width=0.48\linewidth]{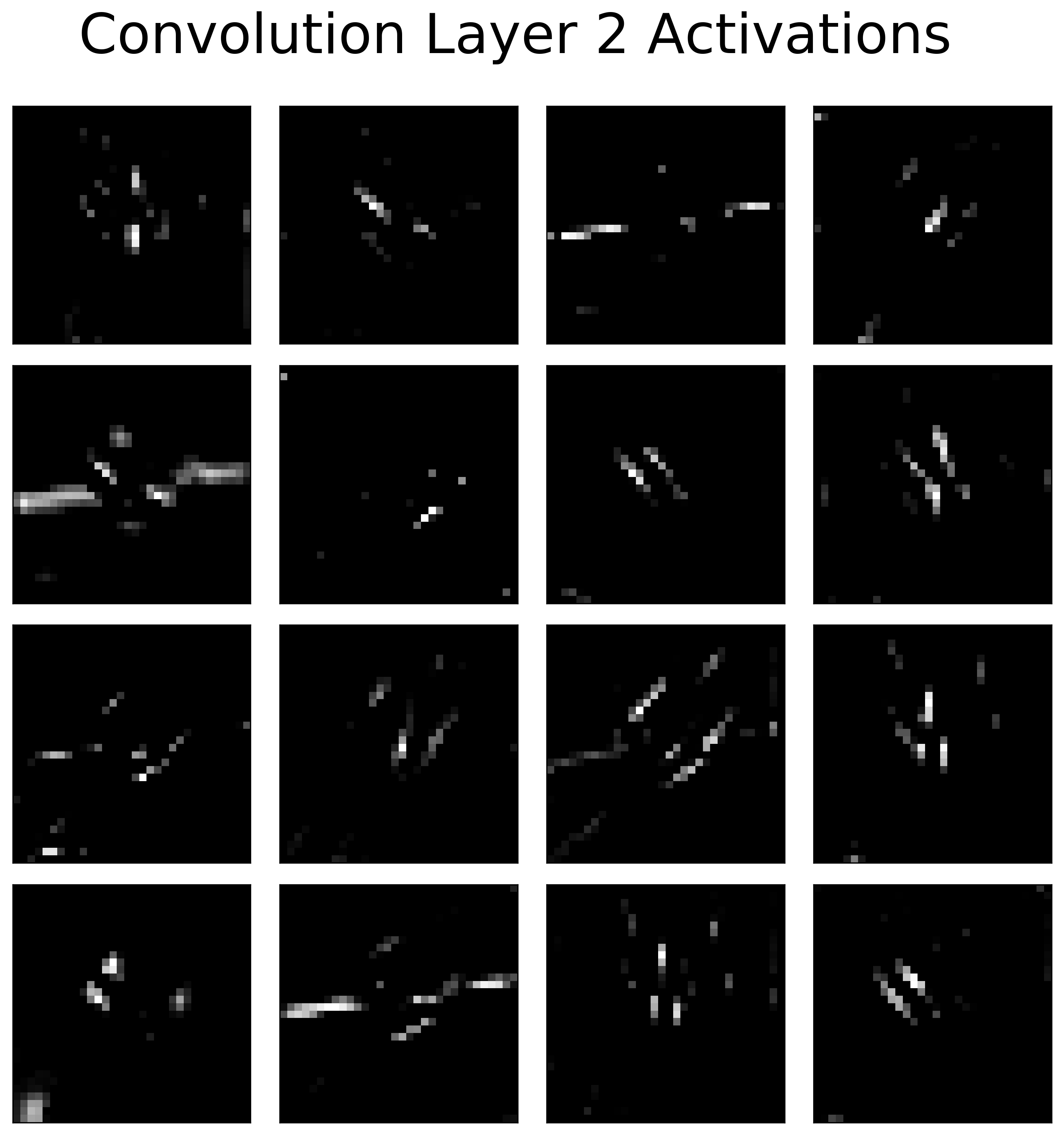}
    \par\medskip
    \includegraphics[width=0.48\linewidth]{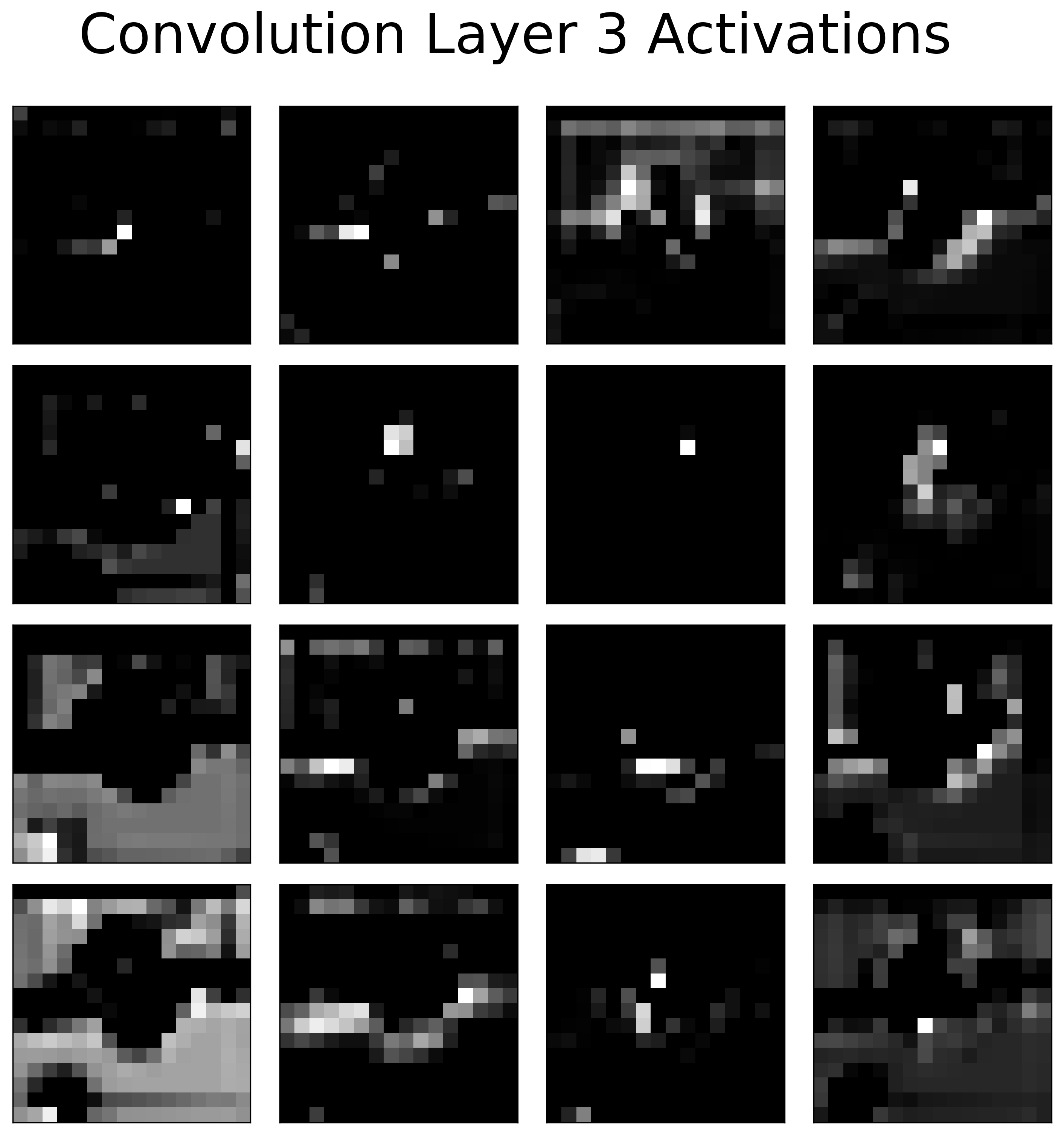}\hfill
    \includegraphics[width=0.48\linewidth]{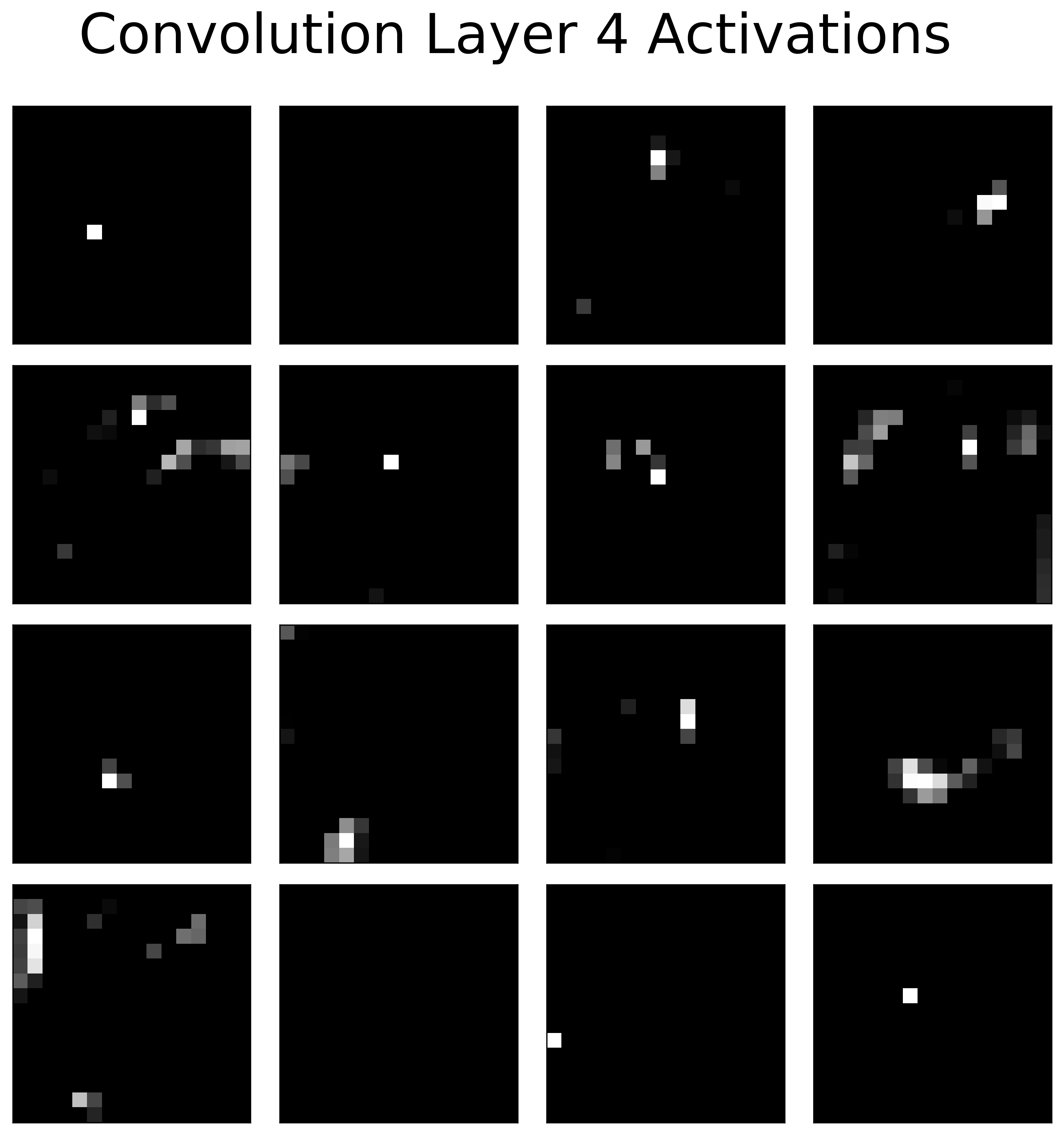}
    \caption{Intermediate layer activations.}
    \label{fig:feature map (b)}
    \end{subfigure}
\end{figure}

\newpage

\begin{figure}[H]
	\centering
    \begin{subfigure}[b]{\linewidth}
    \centering
    \includegraphics[width=0.4\linewidth]{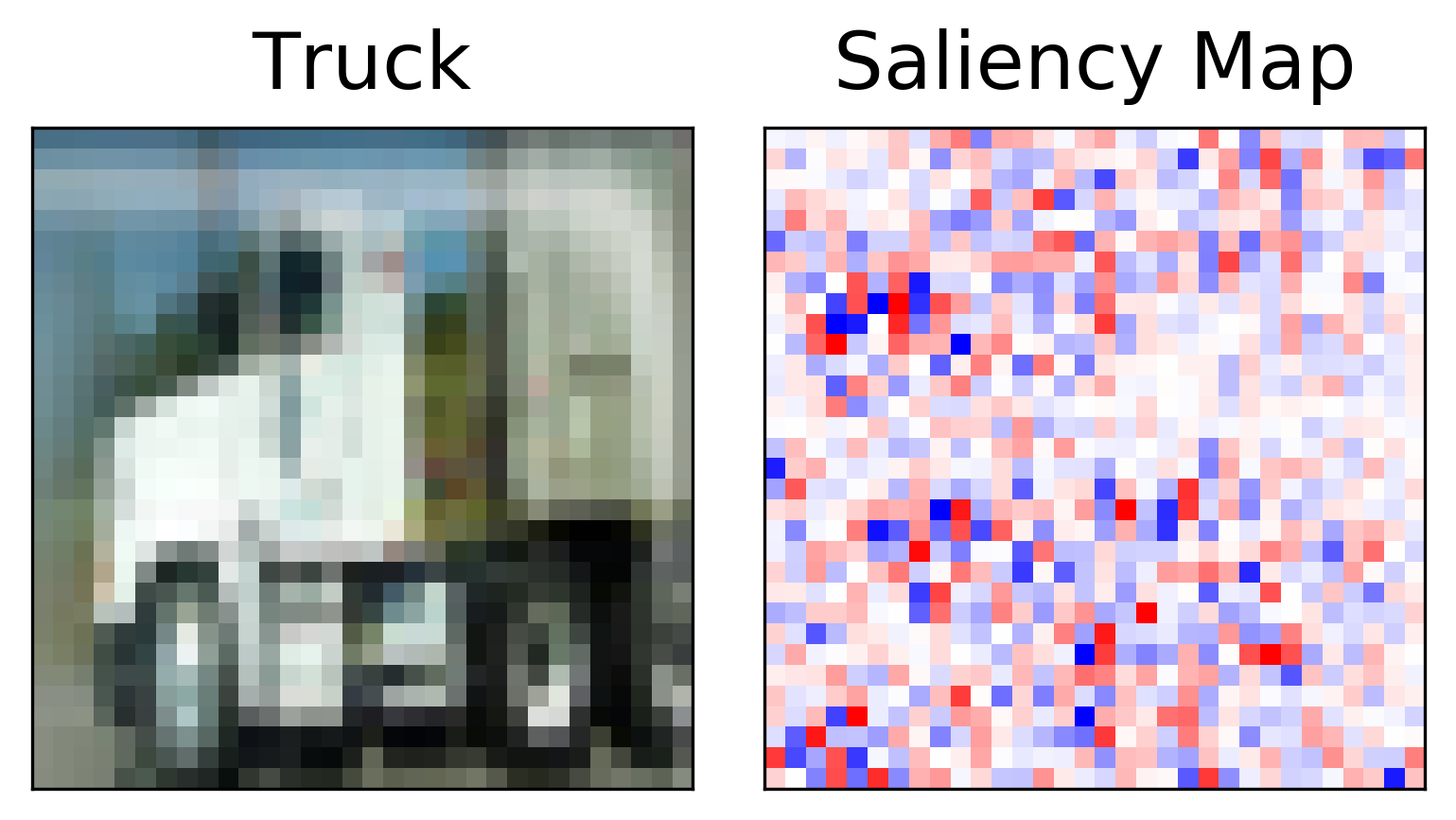}
    \caption{Sample image and its saliency map.}
    \label{fig:feature map (a)}
    \end{subfigure} \\ \vspace{1em}
	\begin{subfigure}[b]{0.8\linewidth}
    \includegraphics[width=0.48\linewidth]{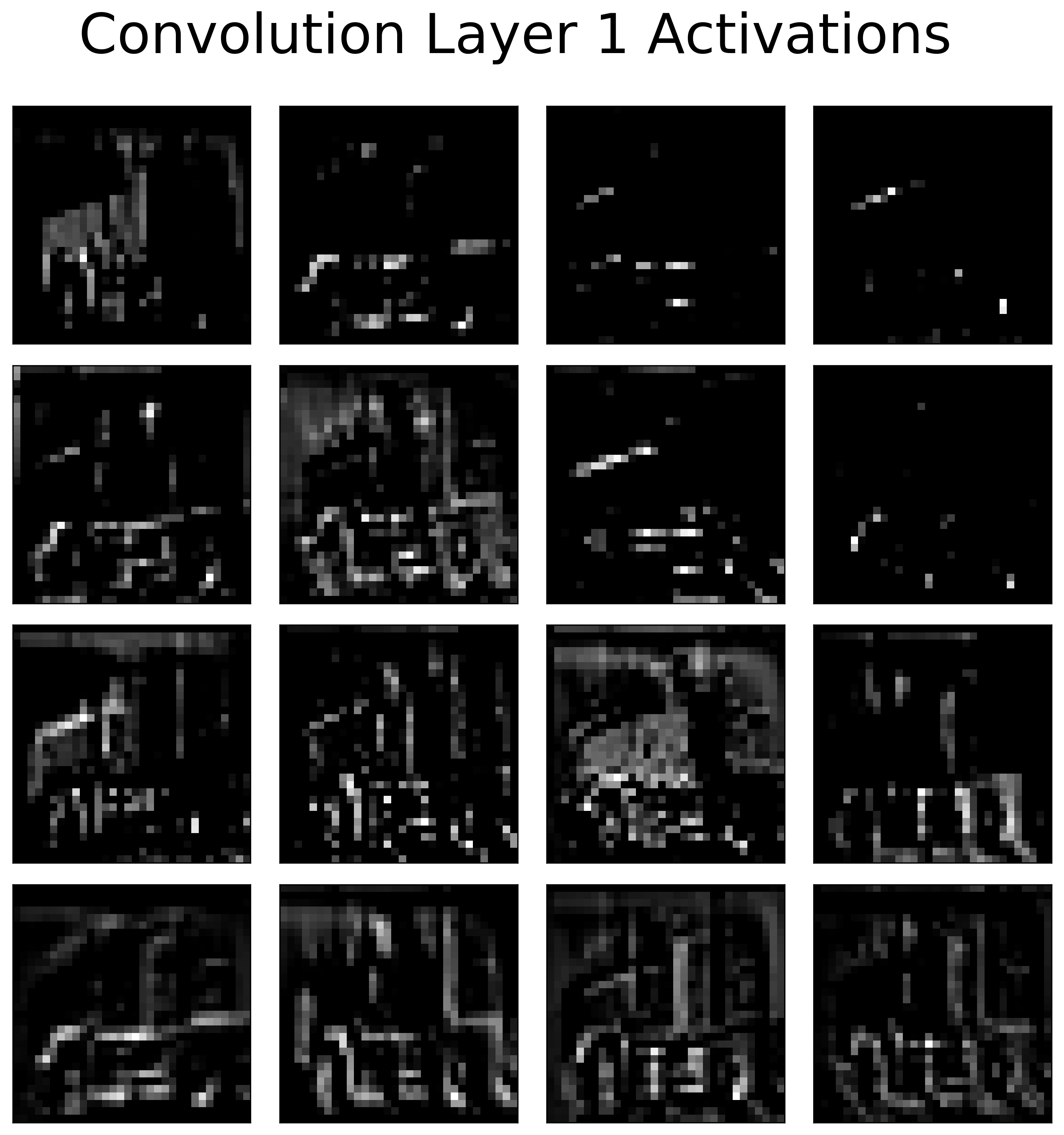}\hfill
    \includegraphics[width=0.48\linewidth]{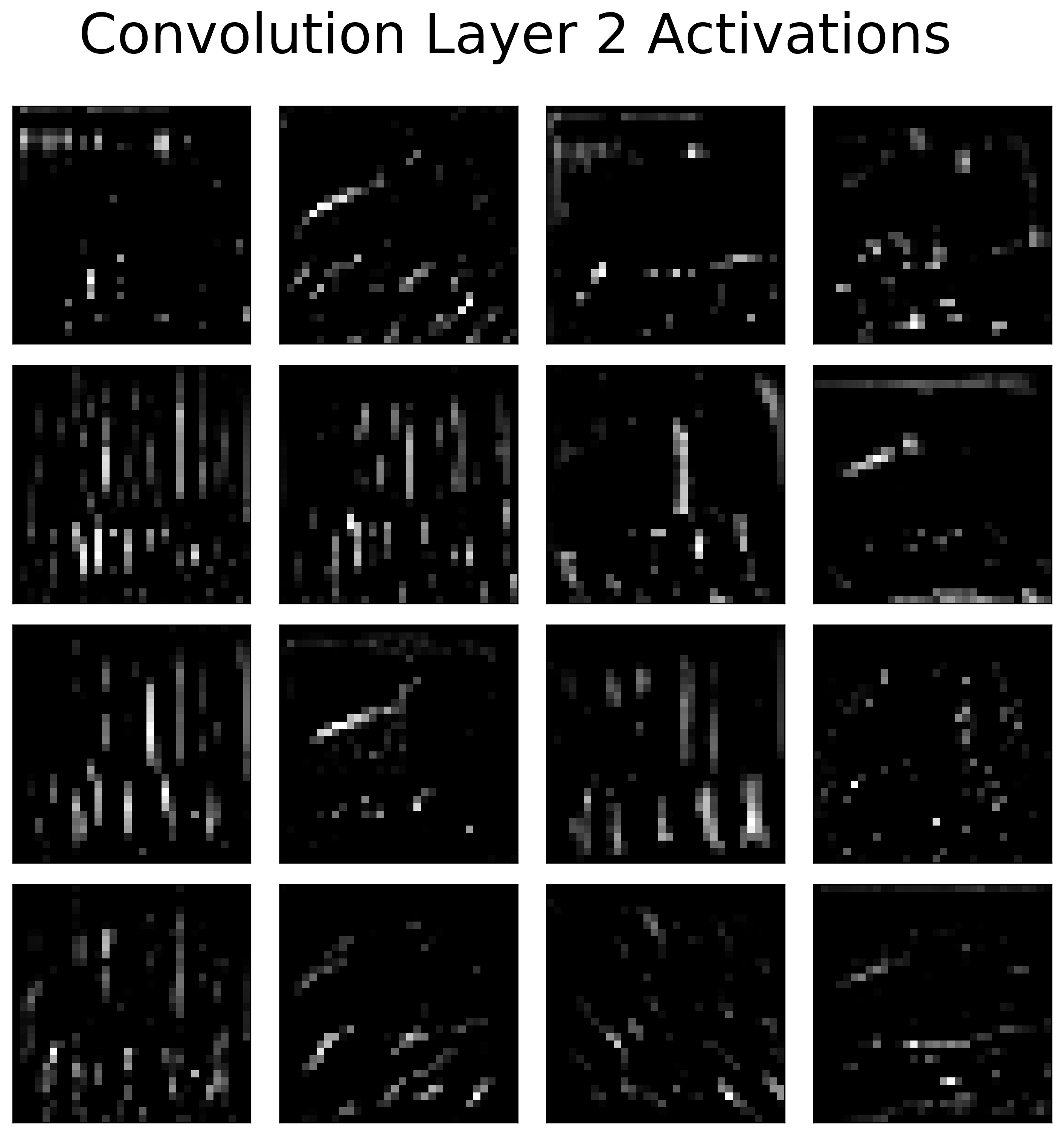}
    \par\medskip
    \includegraphics[width=0.48\linewidth]{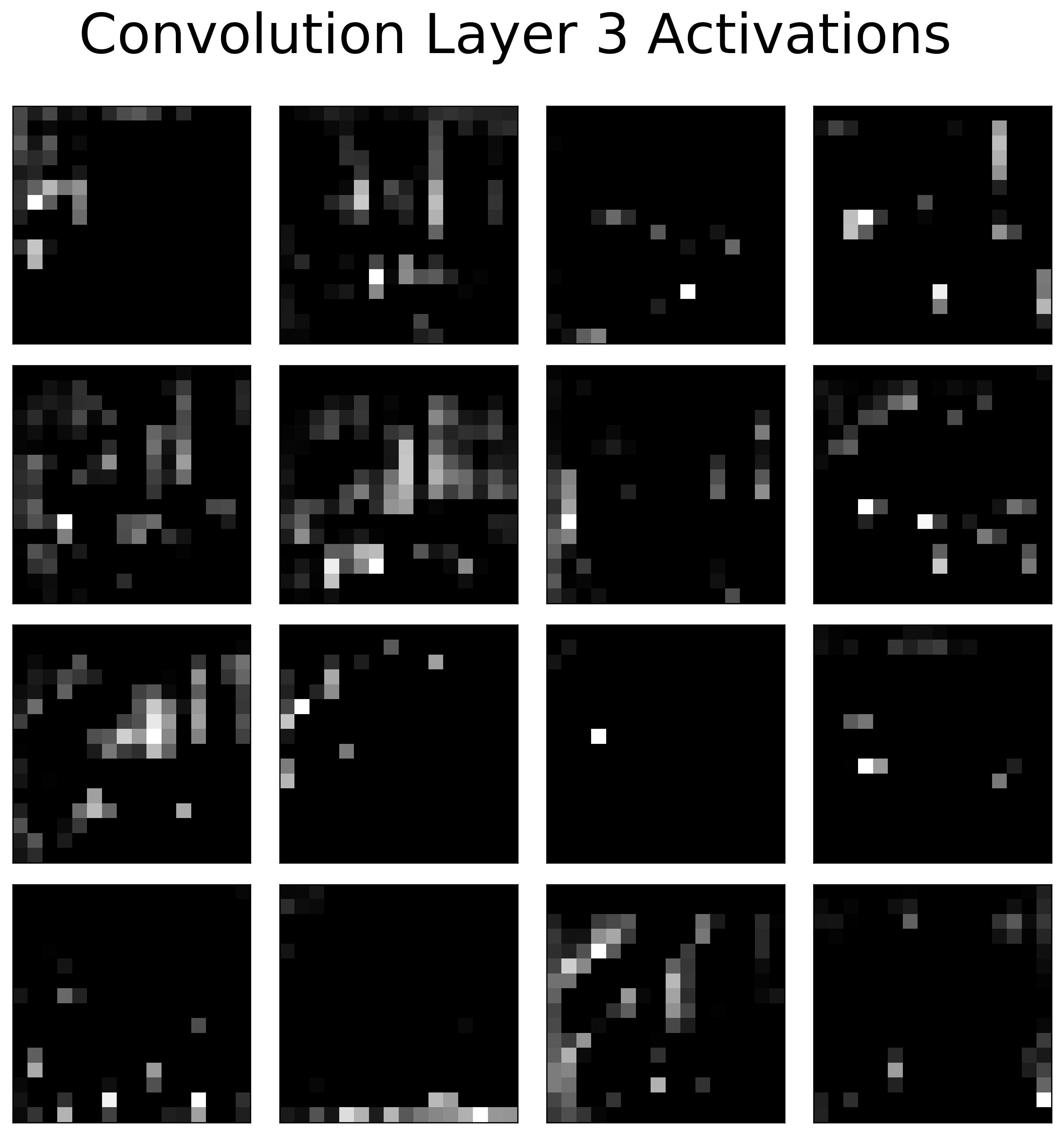}\hfill
    \includegraphics[width=0.48\linewidth]{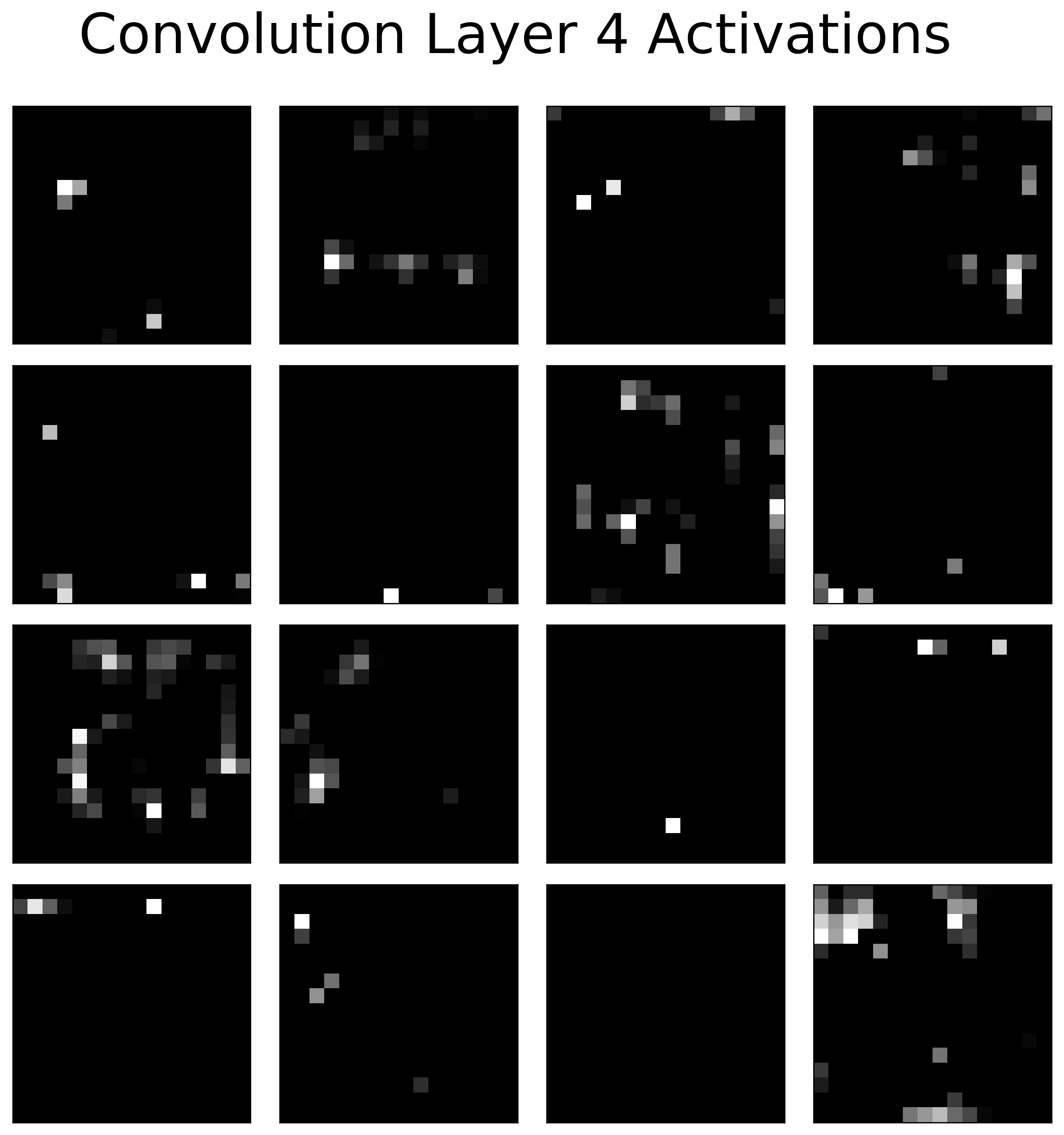}
    \caption{Intermediate layer activations.}
    \label{fig:feature map (b)}
    \end{subfigure}
    \caption{Feature map visualization for an image with a noisy saliency map.}
    \label{fig:feature map}
\end{figure}

\subsection{Supplementary Experiment for Noise Accumulation} \label{section:accum}

\begin{figure}[H]
	\raggedright
    \includegraphics[width=0.3\linewidth]{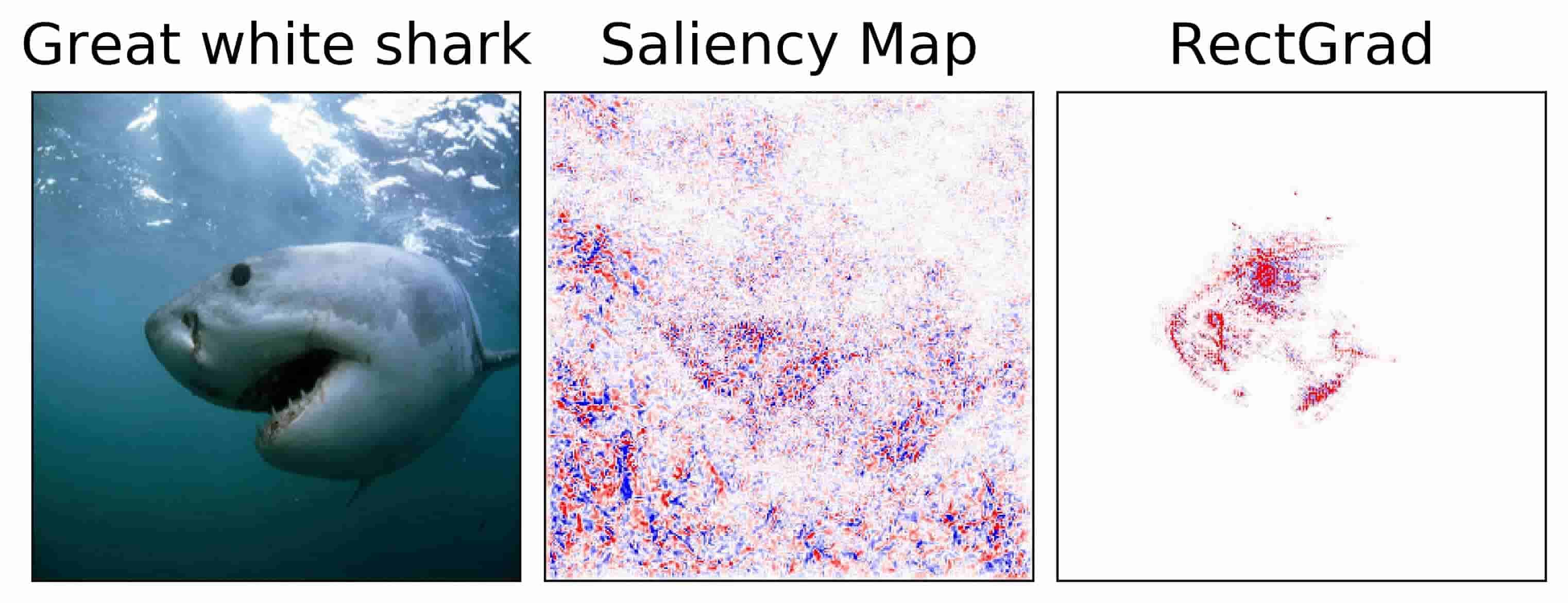}
    \includegraphics[width=\linewidth]{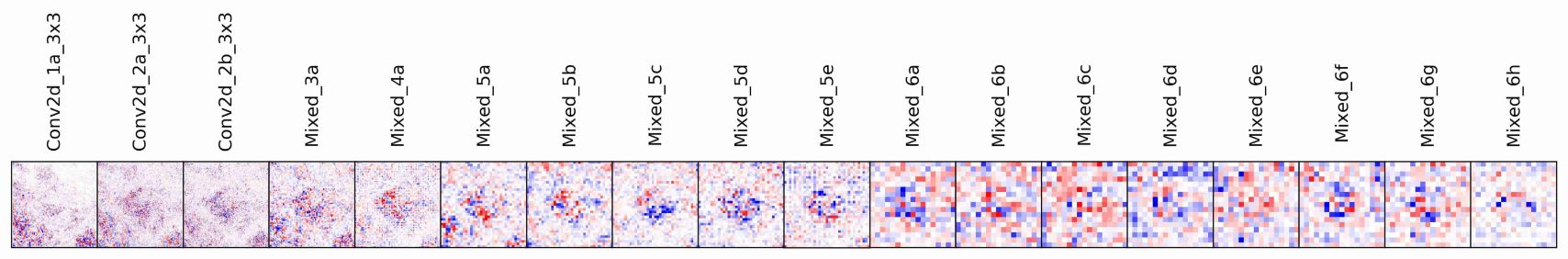}
    \includegraphics[width=\linewidth]{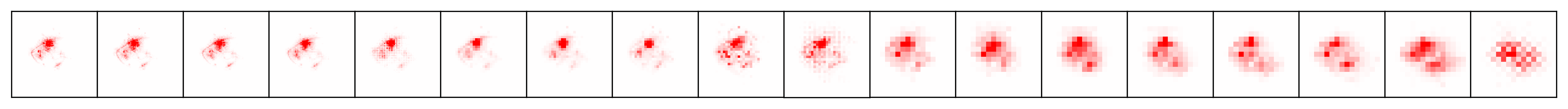} \\ \vspace{0.5em}
    
    \includegraphics[width=0.3\linewidth]{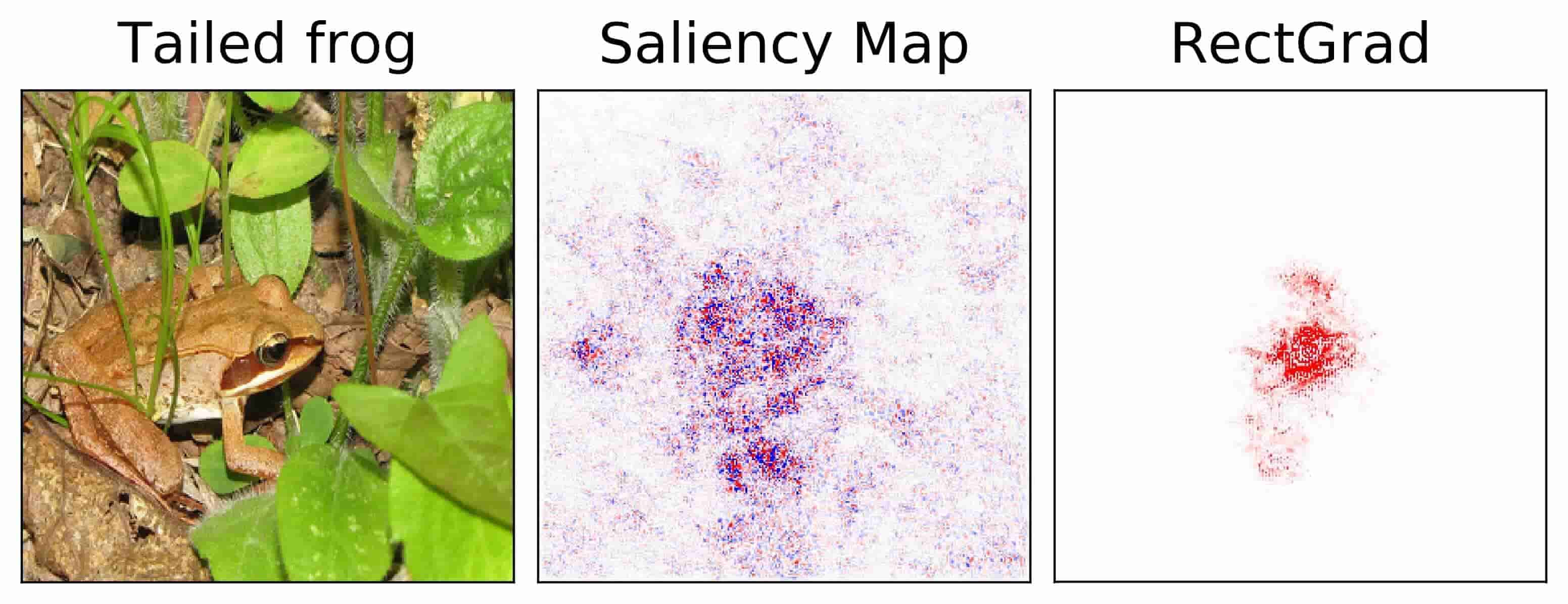}
    \includegraphics[width=\linewidth]{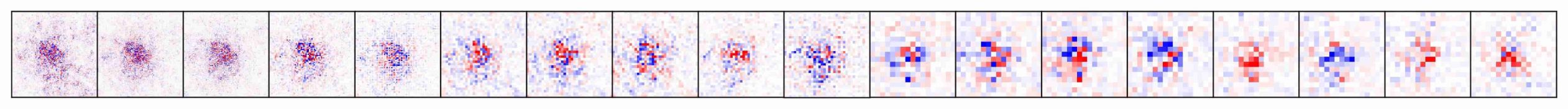}
    \includegraphics[width=\linewidth]{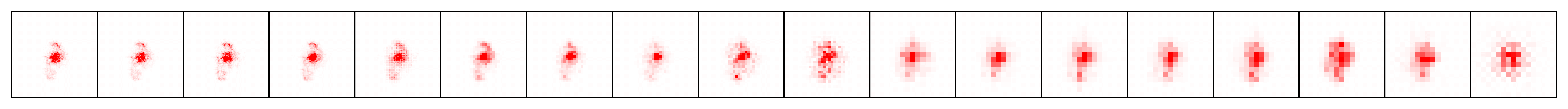} \\ \vspace{0.5em}
    
    \includegraphics[width=0.3\linewidth]{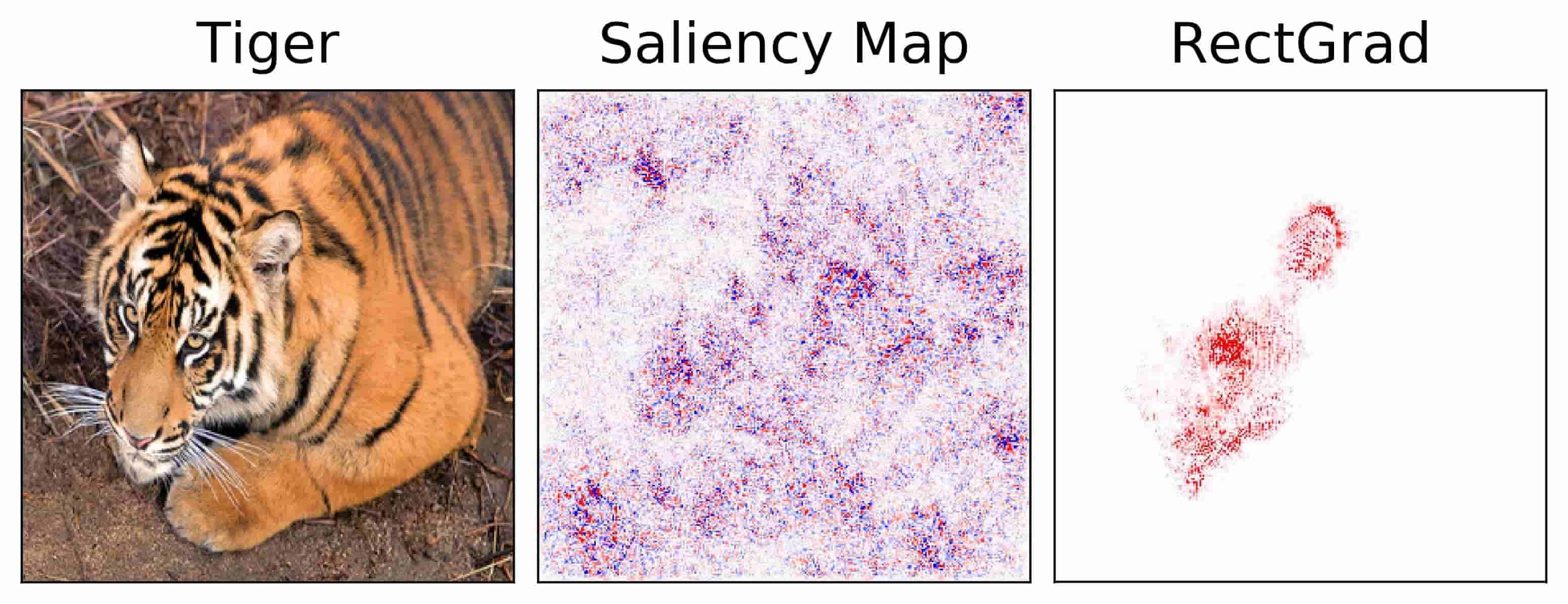}
    \includegraphics[width=\linewidth]{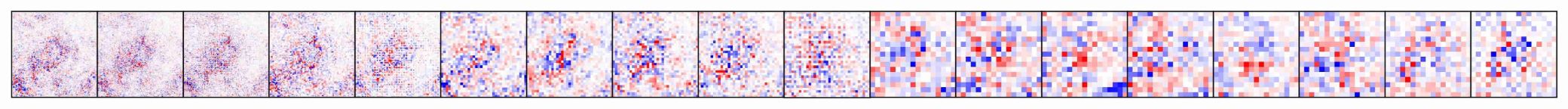}
    \includegraphics[width=\linewidth]{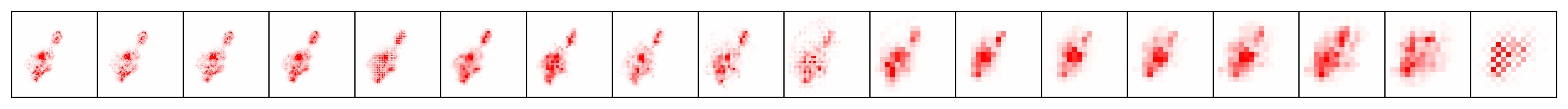} \\ \vspace{0.5em}
    
    \includegraphics[width=0.3\linewidth]{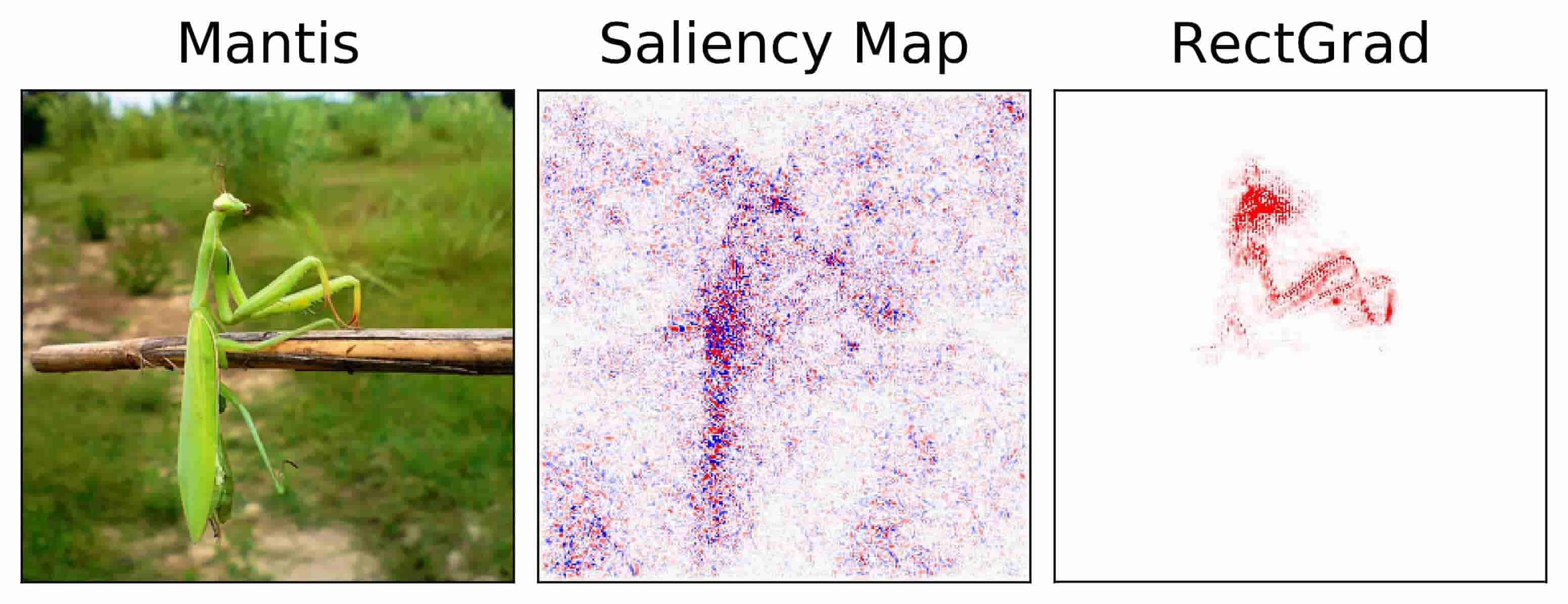}
    \includegraphics[width=\linewidth]{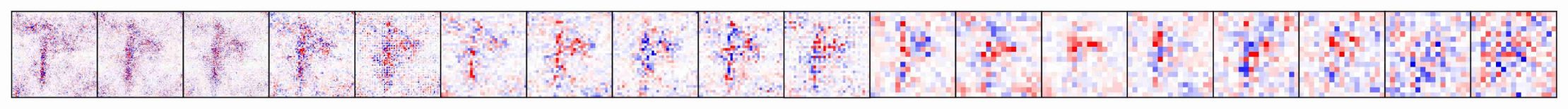}
    \includegraphics[width=\linewidth]{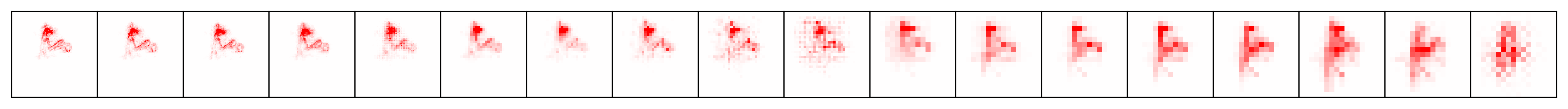} \\ \vspace{0.5em}
\end{figure}

\newpage

\begin{figure}[H]
	\raggedright
    \includegraphics[width=0.3\linewidth]{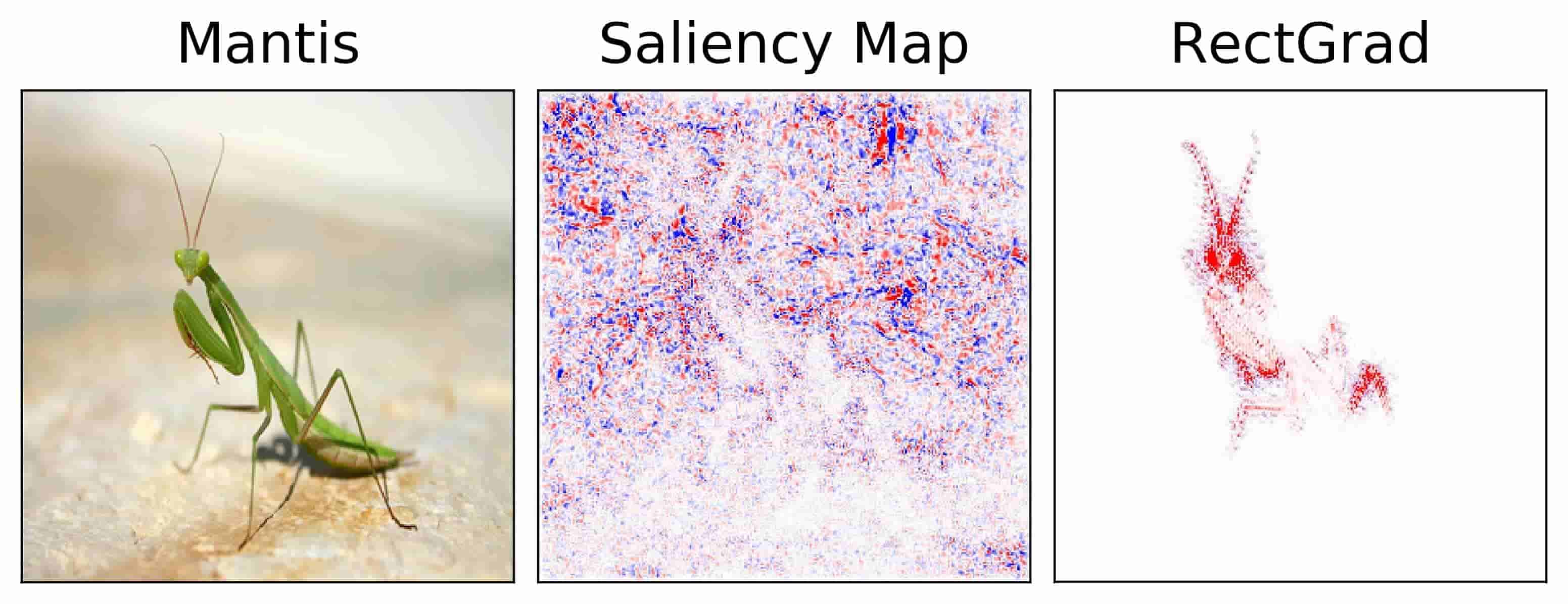}
    \includegraphics[width=\linewidth]{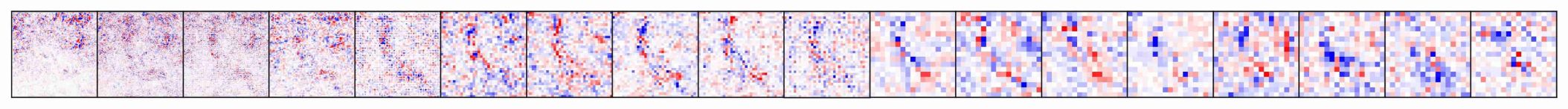}
    \includegraphics[width=\linewidth]{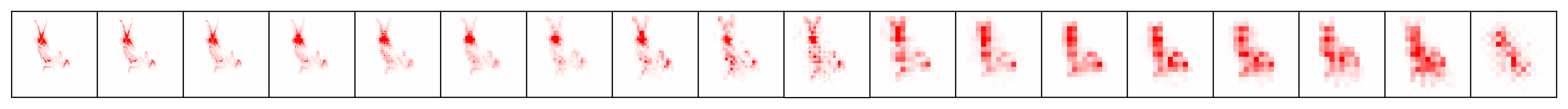} \\ \vspace{0.5em}
    
    \includegraphics[width=0.3\linewidth]{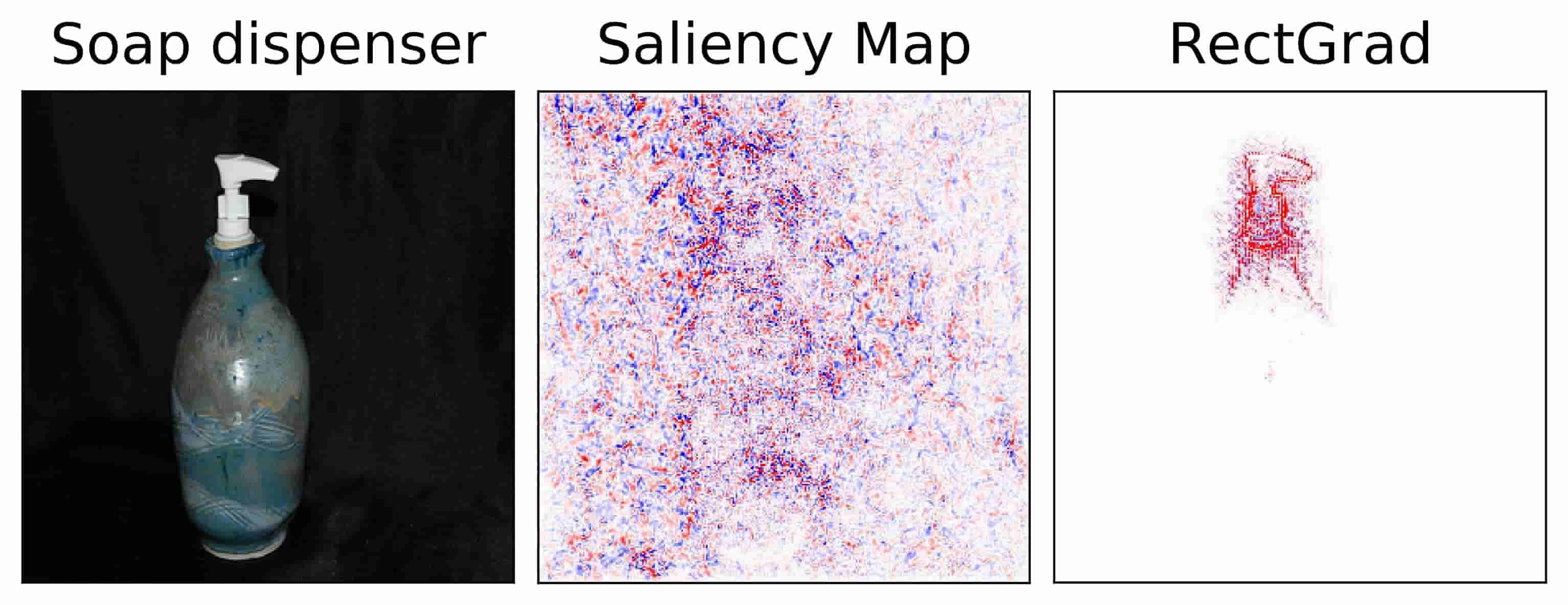}
    \includegraphics[width=\linewidth]{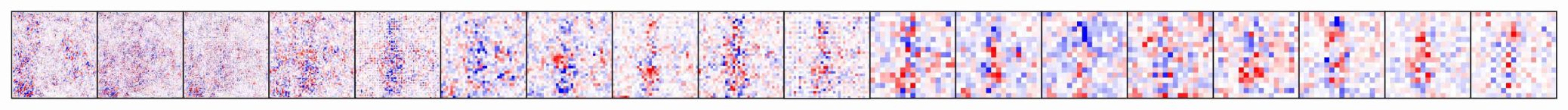}
    \includegraphics[width=\linewidth]{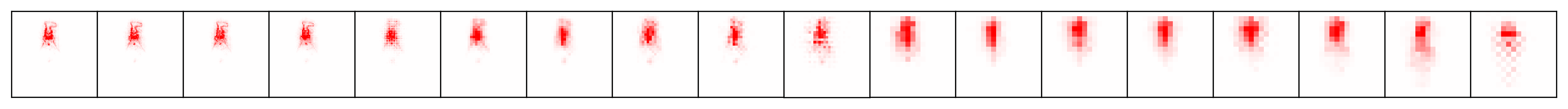} \\ \vspace{0.5em}
    
    \includegraphics[width=0.3\linewidth]{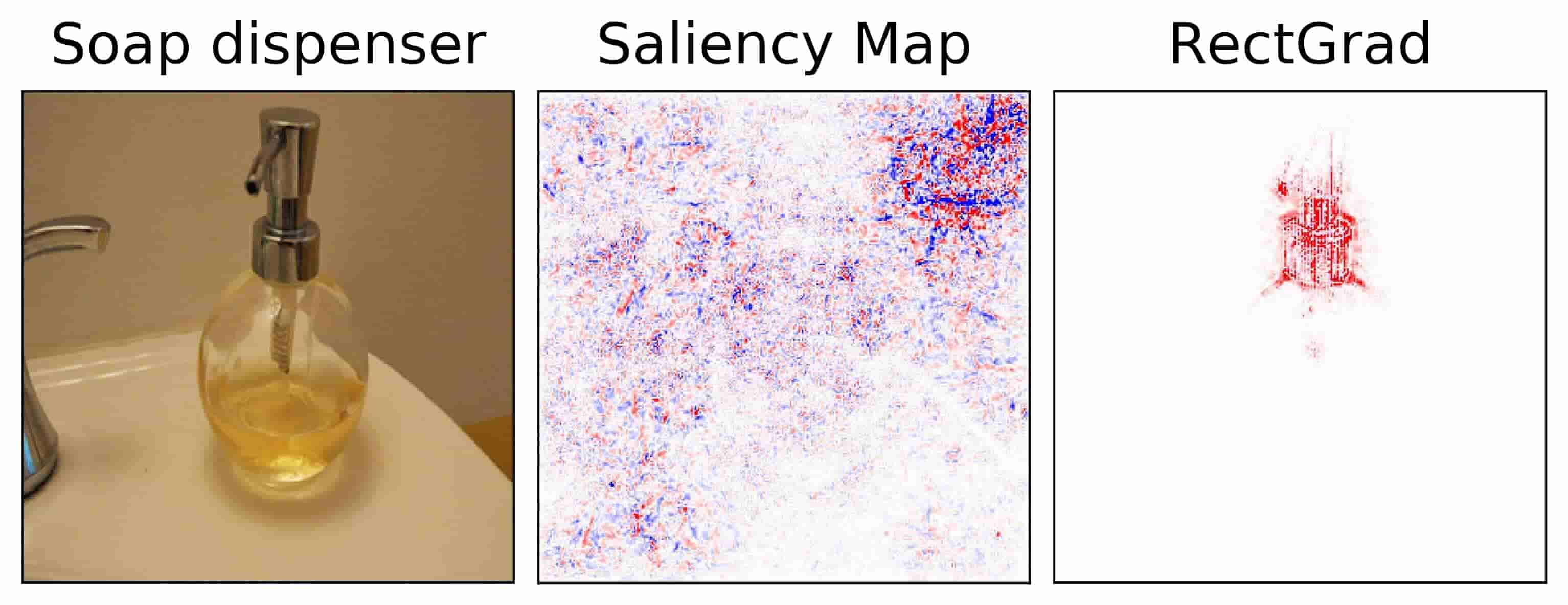}
    \includegraphics[width=\linewidth]{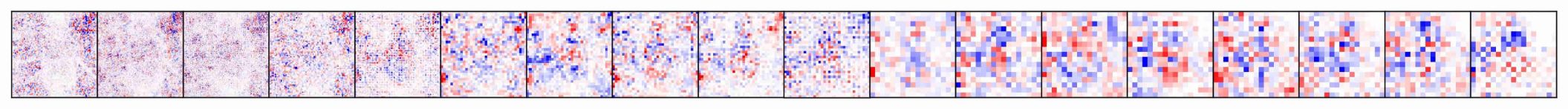}
    \includegraphics[width=\linewidth]{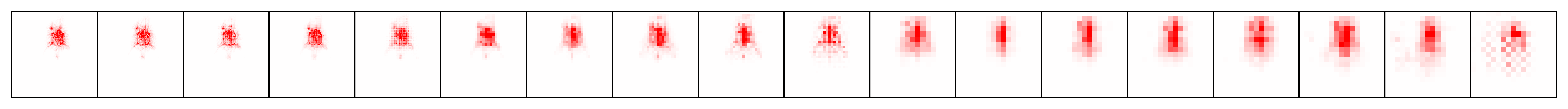}
    \caption{Saliency Map and RectGrad attributions at Inception V4 intermediate layers as they are propagated toward the input layer. We show channel-wise average attributions for hidden layer inputs with respect to the output layer. For each subfigure, first row shows the input image and Saliency Map and RectGrad attribution maps. Second and third rows show Saliency Map and RectGrad attributions at intermediate layers, respectively. An attribution map is closer to the output layer if it is closer to the right.}
    \label{fig:accum2}
\end{figure}

\newpage

\subsection{Qualitative Experiments} \label{section:qualitative figures}

\begin{figure}[H]
	\centering
    \includegraphics[width=0.49\linewidth]{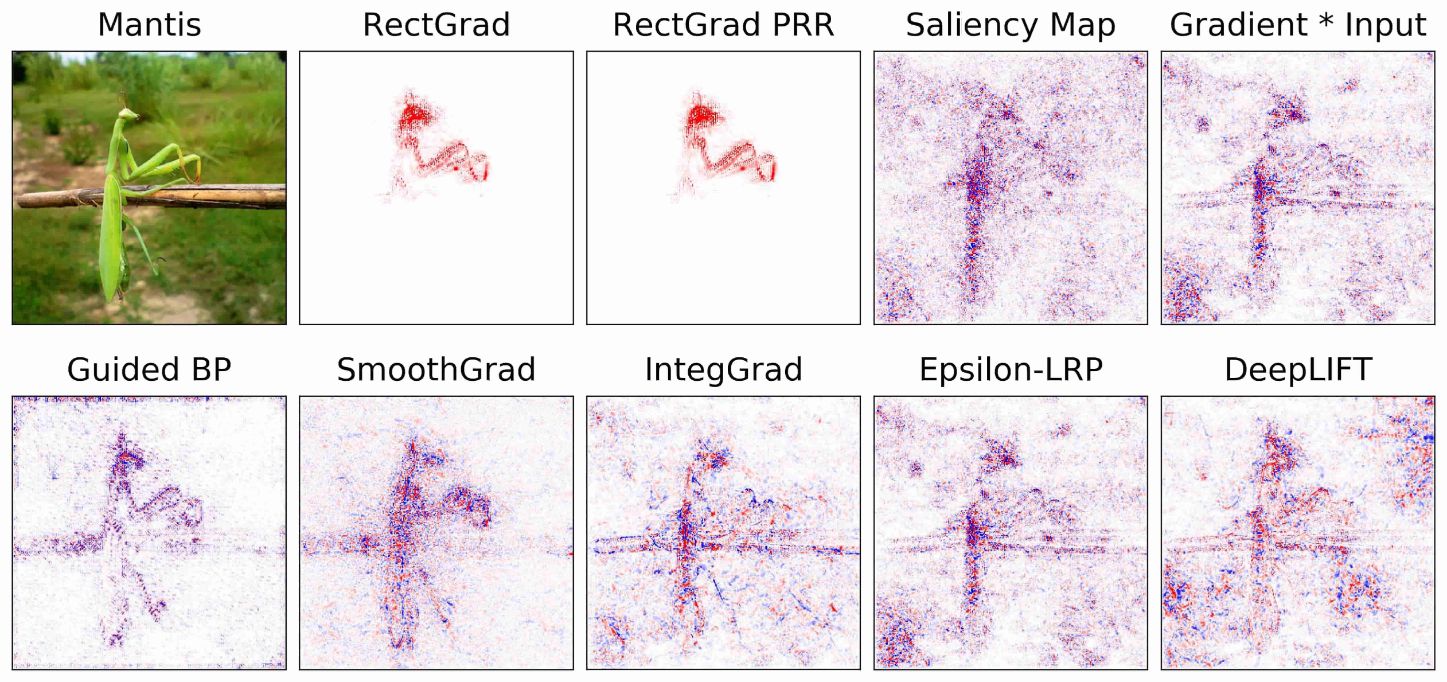}\hfill
    \includegraphics[width=0.49\linewidth]{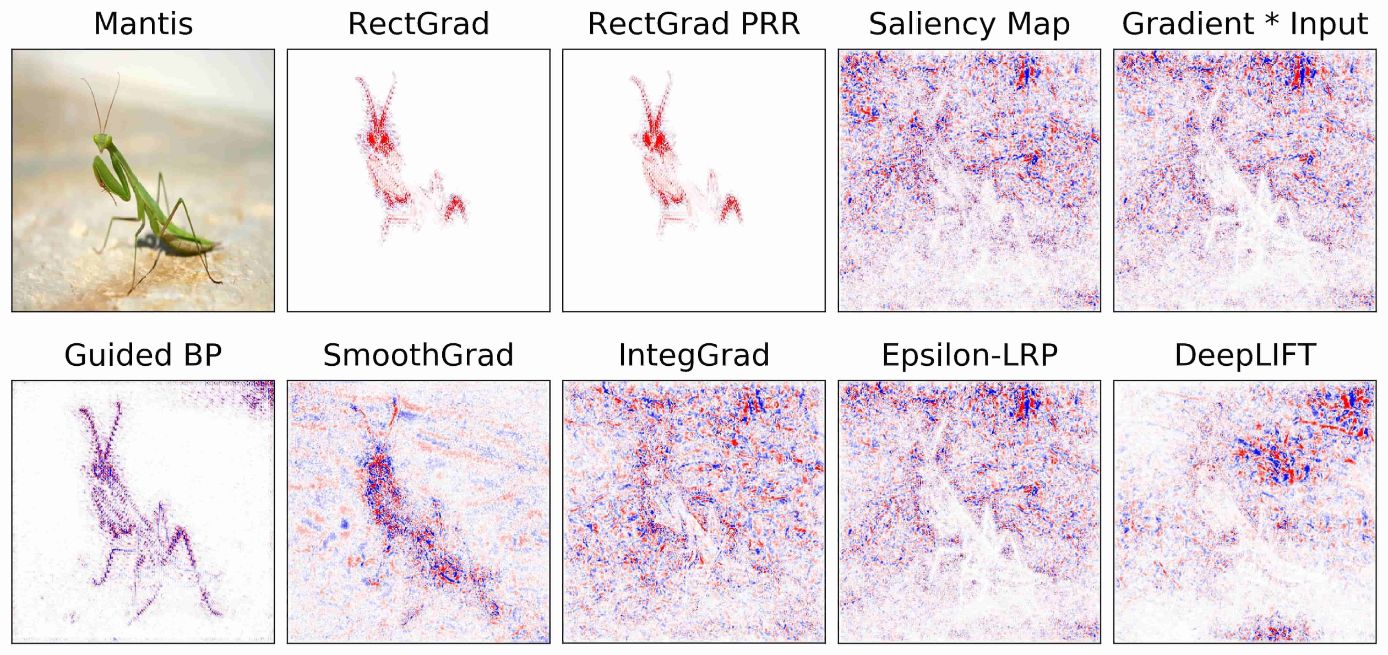}
    \par\medskip
    \includegraphics[width=0.49\linewidth]{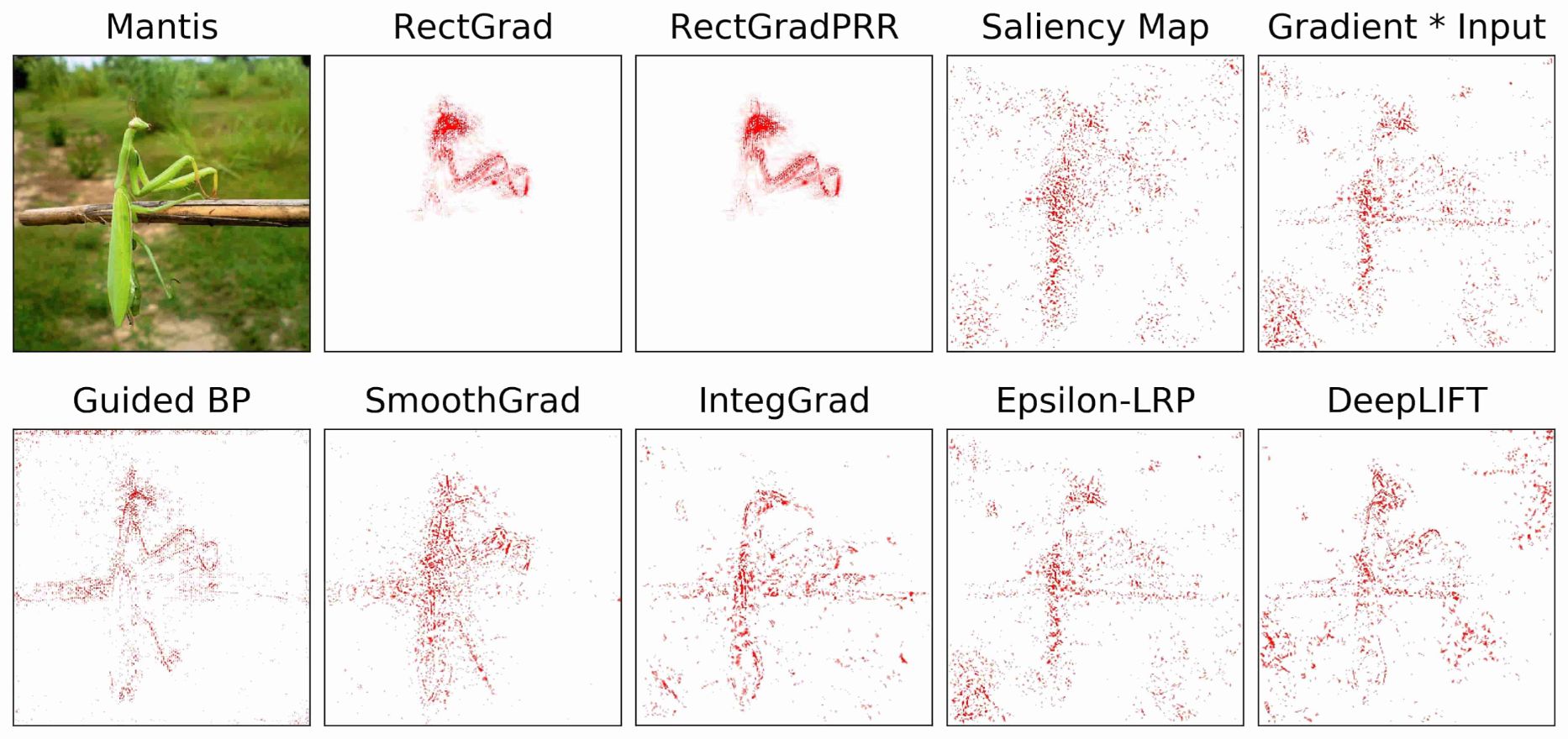}\hfill
    \includegraphics[width=0.49\linewidth]{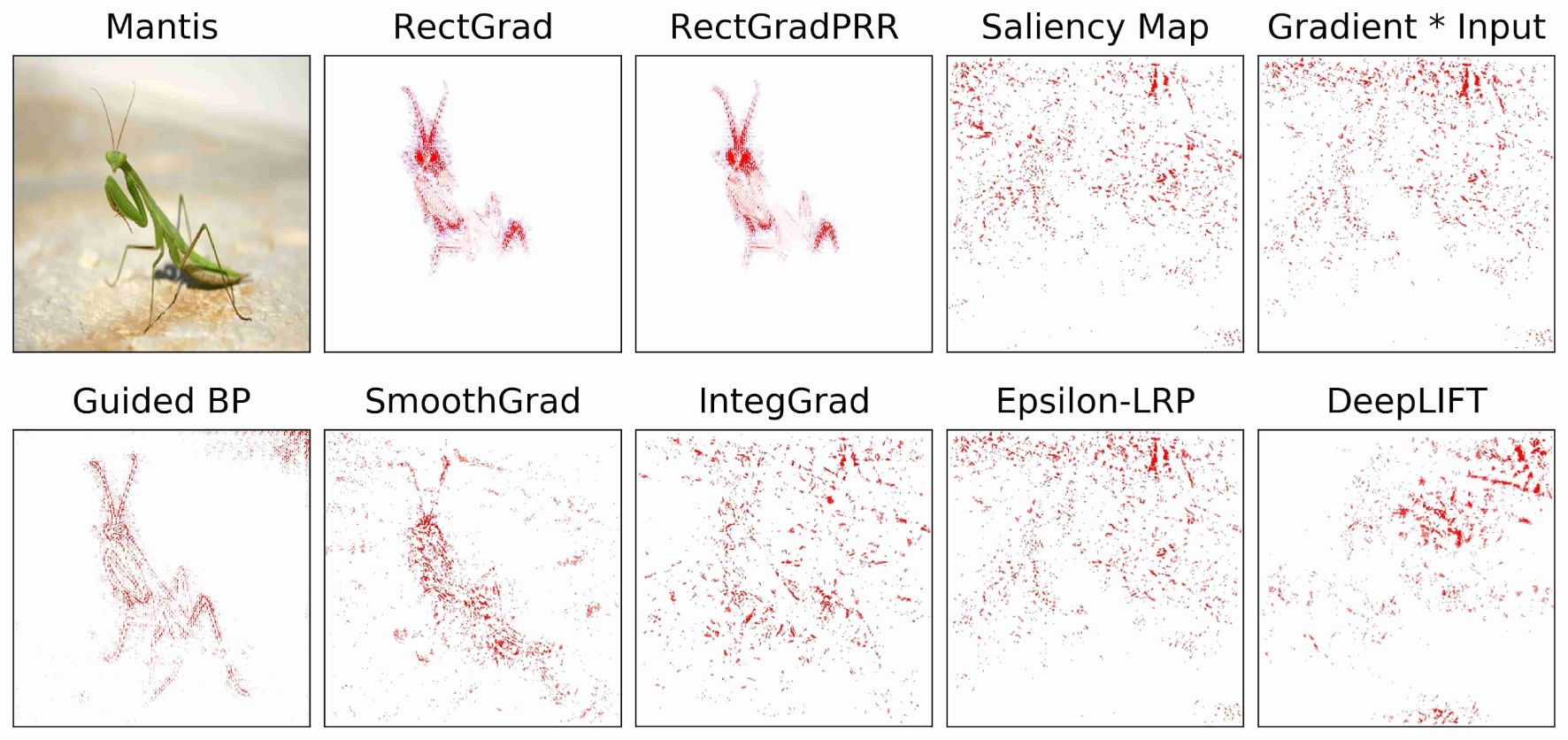}
    \caption{Evaluation of coherence within the same class (rows) without and with final thresholding.}
    \label{fig:coherence2}
\end{figure}

\begin{figure}[h]
	\centering
    \includegraphics[width=0.49\linewidth]{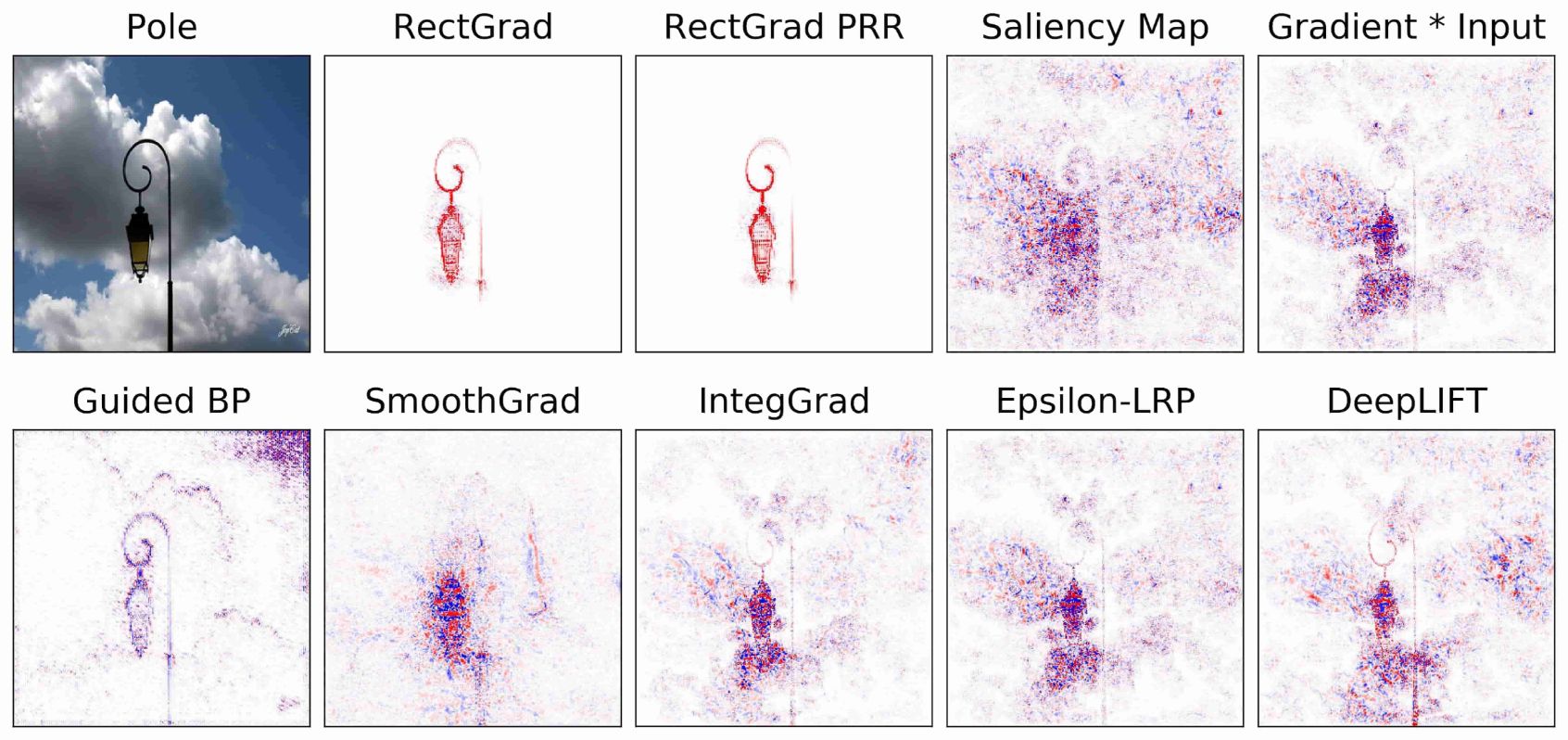}\hfill
    \includegraphics[width=0.49\linewidth]{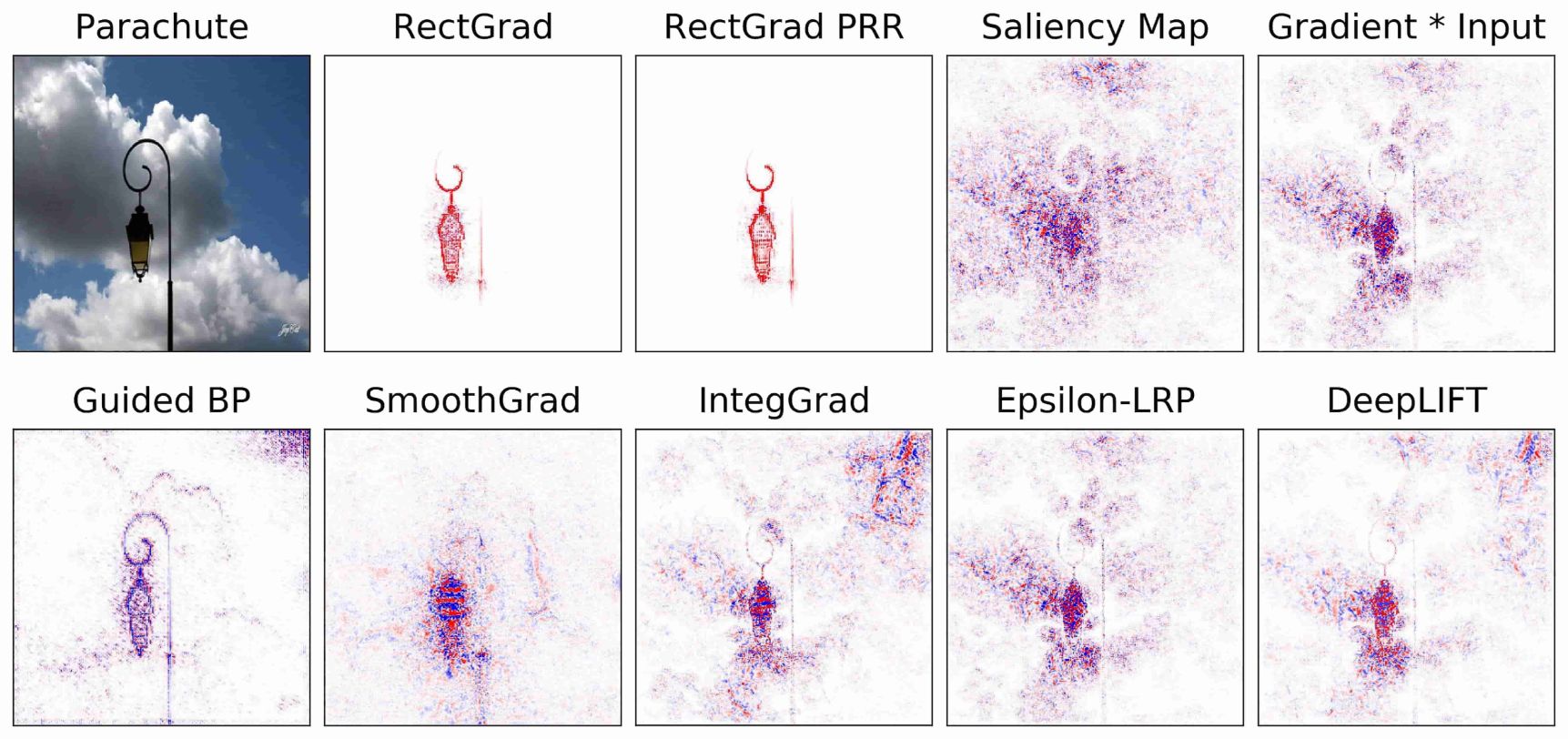}
    \par\medskip
    \includegraphics[width=0.49\linewidth]{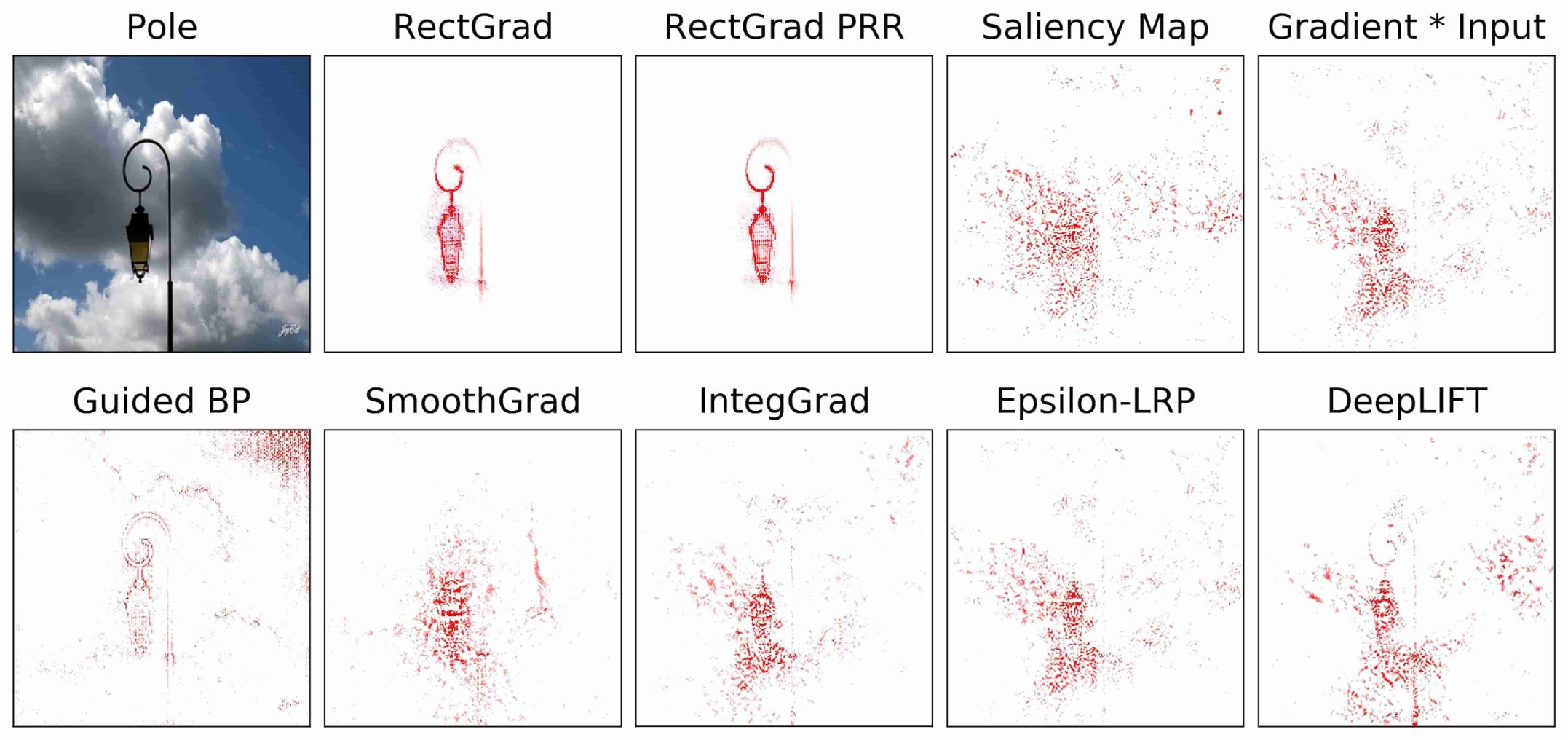}\hfill
    \includegraphics[width=0.49\linewidth]{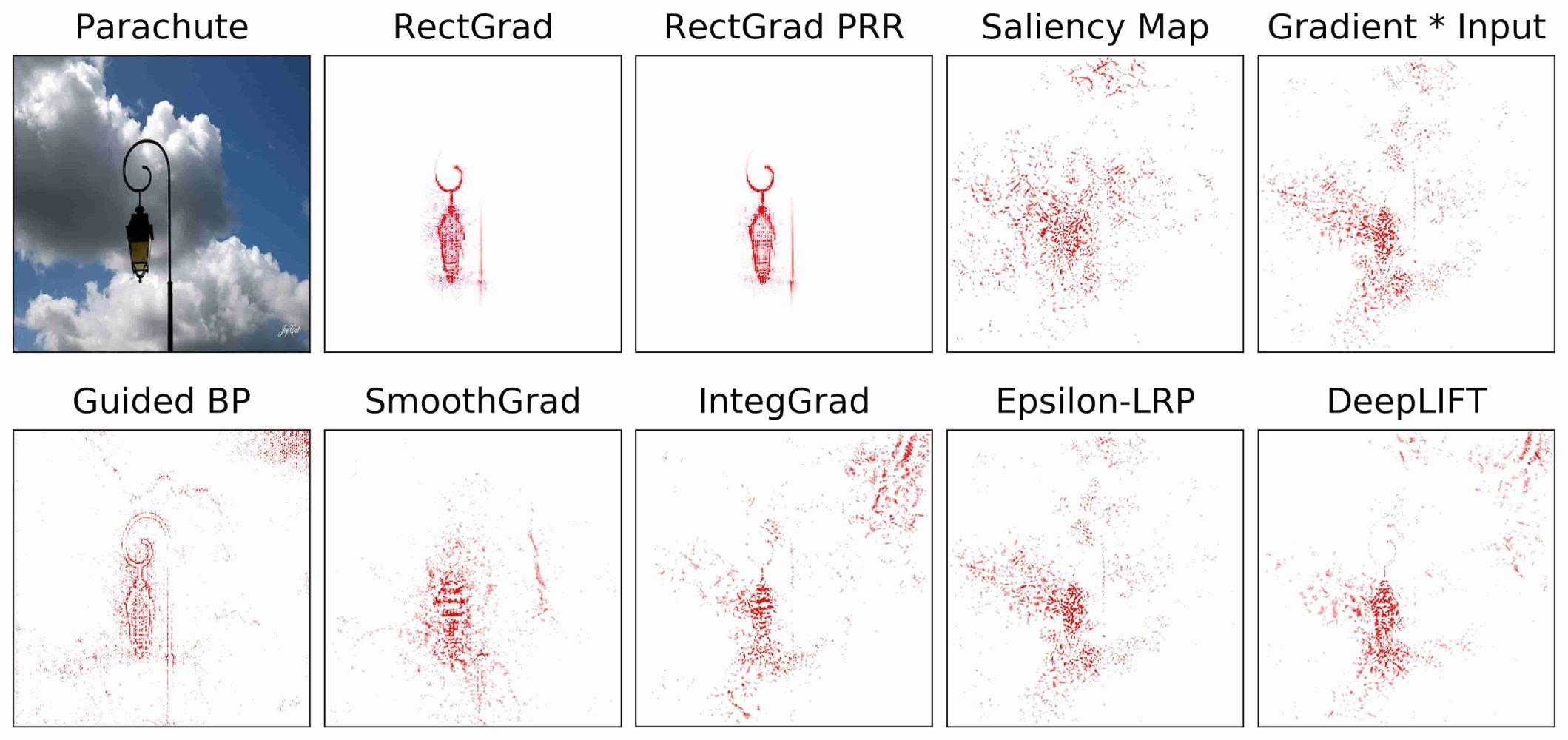}
    \caption{Comparison of attribution maps for images (left column) and their adversarial examples (right column) without and with final thresholding. This figure shows examples where attribution maps produced by RectGrad did not change significantly.}
    \label{fig:adv2}
\end{figure}

\newpage

\section{Rationale Behind the Propagation Rule for Rectified Gradient} \label{section:rationale}

\subsection{Propagation Rule Candidates}

We have the following propagation rule candidates:
\begin{align*}
\text{PR1: } R^{(l)}_i &= \mathbb{I}(a^{(l)}_i \cdot R^{(l + 1)}_i > \tau) \cdot R^{(l + 1)}_i, \\ \text{PR2: } R^{(l)}_i &= \mathbb{I}( |a^{(l)}_i \cdot R^{(l + 1)}_i | > \tau) \cdot R^{(l + 1)}_i, \\
\text{PR3: } R^{(l)}_i &= \mathbb{I}(a^{(l)}_i > \tau) \cdot R^{(l + 1)}_i, \\
\text{PR4: } R^{(l)}_i &= \mathbb{I}(R^{(l + 1)}_i > \tau) \cdot R^{(l + 1)}_i.
\end{align*}
We define the most influential units as those which dominate the function output. We claim that RectGrad propagation rule PR1 correctly identifies the most influential, or the dominating, units while others fail to do so. We start by demonstrating that PR3 and PR4 do not select the dominating unit even for single-layer networks.

\subsection{Single-layer Case}

Consider the affine model $f(a_1, a_2, a_3, a_4) = a_1 + 10 \cdot a_2 - 100 \cdot a_3 + 1000 \cdot a_4 + b$ with bias $b$. We have $\nabla f = (R_1, R_2, R_3, R_4) = (1, 10, -100, 1000)$. Suppose we feed $(a_1, a_2, a_3, a_4) = (3, 2, 1, 0)$ and apply the propagation rules with $q = 74$, i.e., we propagate the gradient only through the most influential unit. Here the function output is dominated by $-100 \cdot a_3$,  gradient should be propagated through $a_3$. However, since $\max a_i = a_1$, PR3 selects $a_1$ as the most influential unit and since $\max R_i = R_4$, PR4 selects $a_4$ as the most influential variable. In this example, we should use PR2 if we are to naively choose the dominating unit. In fact, this propagation rule selects the dominating unit in any affine model since $| a^{(l)}_i \cdot R^{(l + 1)}_i |$ is the absolute value of the input multiplied by its weight.

However, there is a problem with PR2. In a typical multi-class classification setting, the class with the \textit{largest} logit is selected as the decision of the network. Hence it is logical to define important units as those with the largest contribution $a^{(l)}_i \cdot R^{(l + 1)}_i$, not the largest absolute contribution $| a^{(l)}_i \cdot R^{(l + 1)}_i |$. For instance, in the above example, even though $-100 \cdot a_3$ dominates with the largest absolute contribution, it contributes least to the output due to its negative sign. Hence a reasonable propagation rule should first identify the units which have the largest contribution to the output and then select the dominating unit(s) among them. At this point, it is evident that the rule which satisfies this condition is PR2 without the absolute value. This is just PR1, which is the RectGrad propagation rule. Now we have two candidates left. We can apply PR1 to all layers or combine PR1 and PR2 by applying PR1 to the final layer and applying PR2 to lower layers. We show in the multi-layer case that the former works while the latter does not.

\subsection{Multi-layer Case}

Suppose we have an $N$-layer DNN ($N \geq 2$) mapping $\mathbb{R}^d$ to $\mathbb{R}$. Denote the ReLU activation function by $\sigma$, $l$-th layer by $L^{(l)}$, and alternating composition of $\sigma$ and layers $j$ to $k$ ($0 \leq j < k \leq N$) by
\begin{align*}
L^{(j,k)} = L^{(k)} \circ \sigma (L^{(k - 1)}) \circ \cdots \circ \sigma (L^{(j)}).
\end{align*}
Denote the $i$-th components of $L^{(l)}$ and $L^{(j,k)}$ by $L^{(l)}_i$ and $L^{(j,k)}_i$ respectively. Finally, denote the output dimension of $l$-th layer by $D_l$. Under this notation, $D_0 = d$, $D_N = 1$, $L^{(l)}$ is an affine function mapping $\mathbb{R}^{D_{l - 1}}$ to $\mathbb{R}^{D_l}$, $L^{(j,k)}$ is a nonlinear function mapping $\mathbb{R}^{D_{j - 1}}$ to $\mathbb{R}^{D_k}$, and the logit for $\mathbf{x} \in \mathbb{R}^d$ is $L^{(0,N)}(\mathbf{x})$.

Now we show by induction that PR1 identifies influential units at all layers. We have already verified the base case in the single-layer case. That is, given the logit $L^{(0,N)}(\mathbf{x})$, PR1 correctly identifies influential units in $L^{(N - 1)}$. In the inductive step, we show that if PR1 correctly identifies influential units in $L^{(l)}$, then we can again apply PR1 to identify influential units in $L^{(l - 1)}$. By induction hypothesis, PR1 identifies and assigns gradient $R_i^{(l)}{'}$ to units $L^{(l)}_i$ with the largest contribution and $0$ to others. Here, $R_i^{(l)}{'}$ is an approximate measure of how sensitive the output is to changes in $L^{(l)}_i$. Specifically,
\begin{align} \label{eq:sensitivity1}
L^{(0,N)}(\mathbf{x}) \approx R_i^{(l)}{'} \cdot L^{(l)}_i(\sigma(L^{(l - 1)}(\mathbf{v}))) + c
\end{align}
for $\mathbf{v} = \sigma(L^{(0,l-2)}(\mathbf{x}))$ and an appropriate constant $c$.

Now we show PR1 correctly identifies influential units in $L^{(l - 1)}(\mathbf{v})$. Suppose $L^{(l)}_1$ is identified as the unit with the largest contribution to the output. We reuse our toy example $f(a_1, a_2, a_3, a_4) = a_1 + 10 \cdot a_2 - 100 \cdot a_3 + 1000 \cdot a_4 + b$ with $\sigma(L^{(l - 1)}(\mathbf{v})) = (a^{(l - 1)}_1, a^{(l - 1)}_2, a^{(l - 1)}_3, a^{(l - 1)}_4) = (3, 2, 1, 0)$. Here we have $R_i^{(l)} = R_i \cdot R_1^{(l)}{'}$ and by Equation \ref{eq:sensitivity1},
\begin{align} \label{eq:sensitivity2}
L^{(0,N)}(\mathbf{x}) &\approx R_1^{(l)}{'} \cdot L^{(l)}_1(\sigma(L^{(l - 1)}(\mathbf{v}))) + c \nonumber \\
&= R_1^{(l)}{'} \cdot (a^{(l - 1)}_1 + 10 \cdot a^{(l - 1)}_2 - 100 \cdot a^{(l - 1)}_3 + 1000 \cdot a^{(l - 1)}_4 + b) + c \nonumber \\
&= R_1^{(l)} \cdot a^{(l - 1)}_1 + R_2^{(l)} \cdot a^{(l - 1)}_2 + R_3^{(l)} \cdot a^{(l - 1)}_3 + R_4^{(l)} \cdot a^{(l - 1)}_4 + c'.
\end{align}
where $c' = c + b$. PR1 correctly identifies influential units since $a^{(l)}_i \cdot R^{(l + 1)}_i$ is approximately the amount of the unit's contribution to the output. Clearly this reasoning applies even when multiple units $L^{(l)}_{i_k}$ ($k = 1, \ldots, n$) are identified as influential to the output, since a linear combination of affine functions is still affine:
\begin{align*}
L^{(0,N)}(\mathbf{x}) &\approx \sum_{k = 1}^n R_{i_k}^{(l)}{'} \cdot L^{(l)}_{i_k}(\sigma(L^{(l - 1)}(\mathbf{v}))) + c = \sum_{i = 1}^{D_l} R_i^{(l)} \cdot a^{(l - 1)}_i + c'.
\end{align*}
We have shown the base case the inductive step and hence our claim holds for all layers.

On the other hand, PR2 fails to select the influential units in $L^{(l - 1)}$ even when the influential units in $L^{(l)}$ are given. It suffers from the same problem that we pointed out in the one-layer case: in Equation \ref{eq:sensitivity2}, if $R_1^{(l)}{'} > 0$, $a^{(l - 1)}_3$ still contributes least to the output due to its negative sign. However, PR2 selects $a^{(l - 1)}_3$ since it has the largest absolute contribution. Hence using PR2 may cause the gradient to be propagated through units with the least contribution to the output.

\subsection{Experimental Justification of PR1}

\begin{figure}[H]
	\centering
    \includegraphics[width=0.33\linewidth]{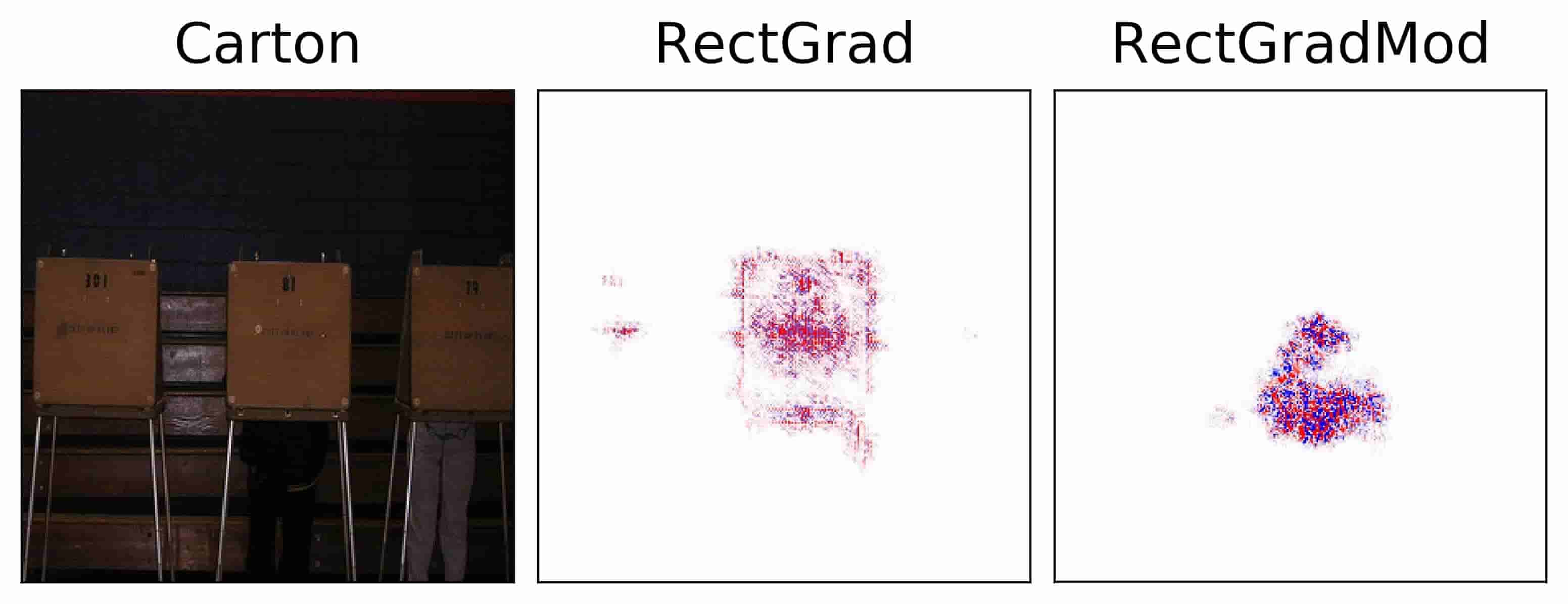}\hfill
    \includegraphics[width=0.33\linewidth]{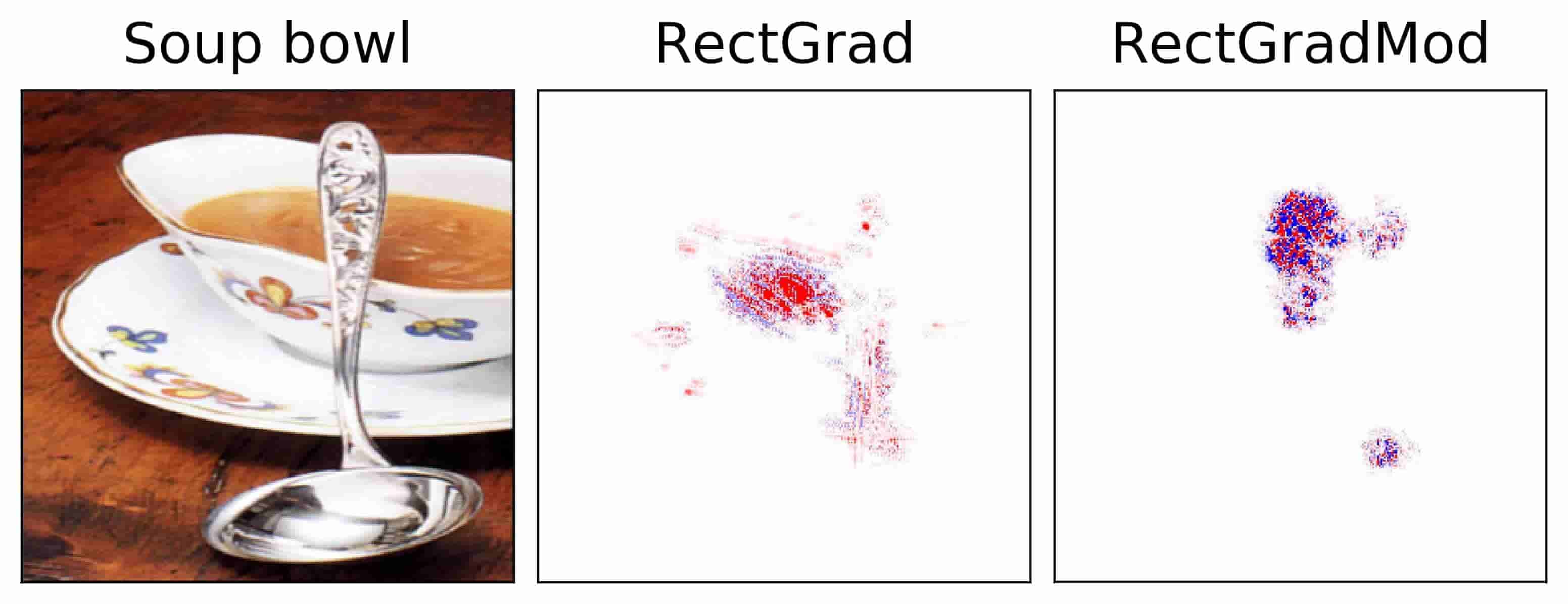}\hfill
    \includegraphics[width=0.33\linewidth]{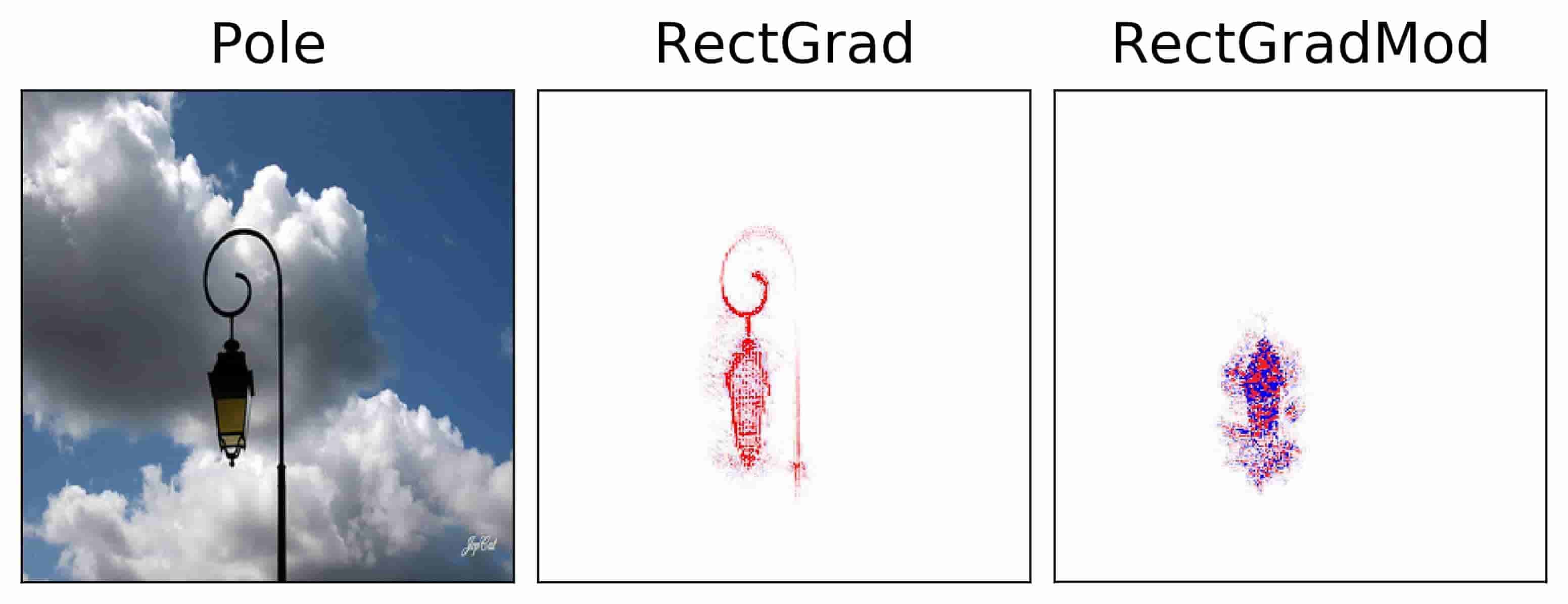}
    \par\medskip
    \includegraphics[width=0.33\linewidth]{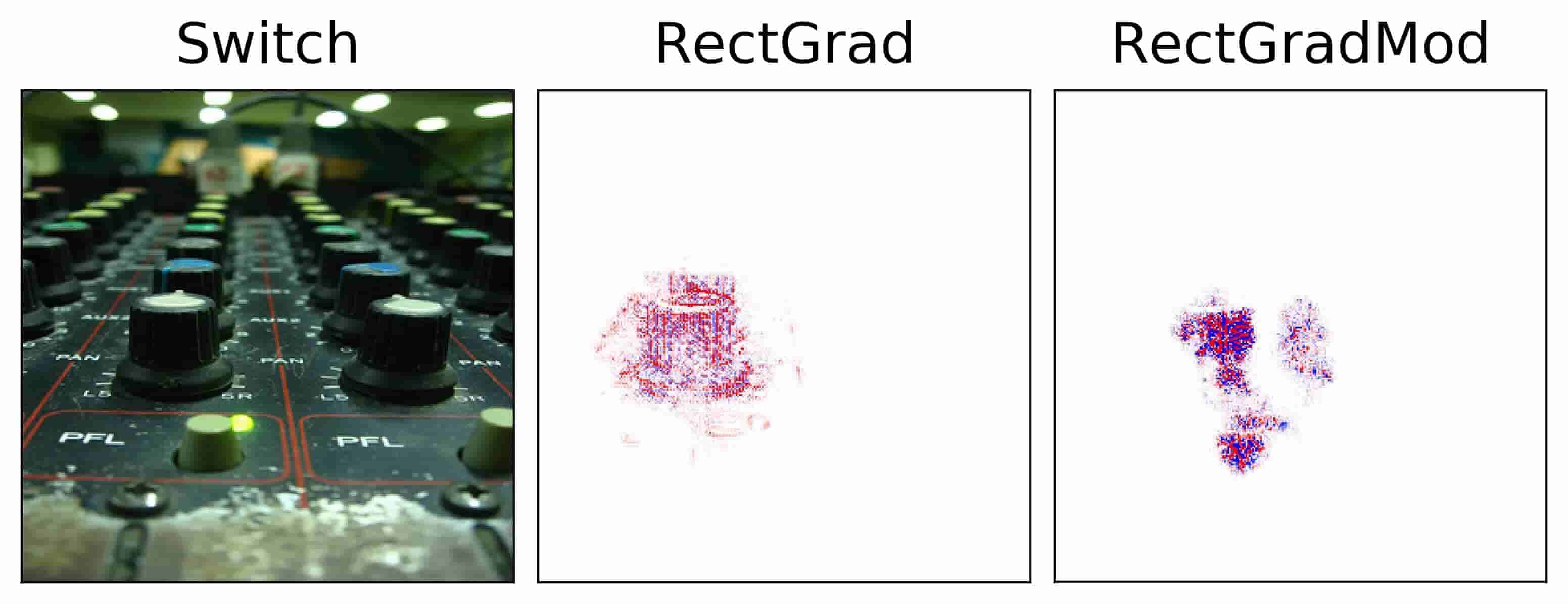}\hfill
    \includegraphics[width=0.33\linewidth]{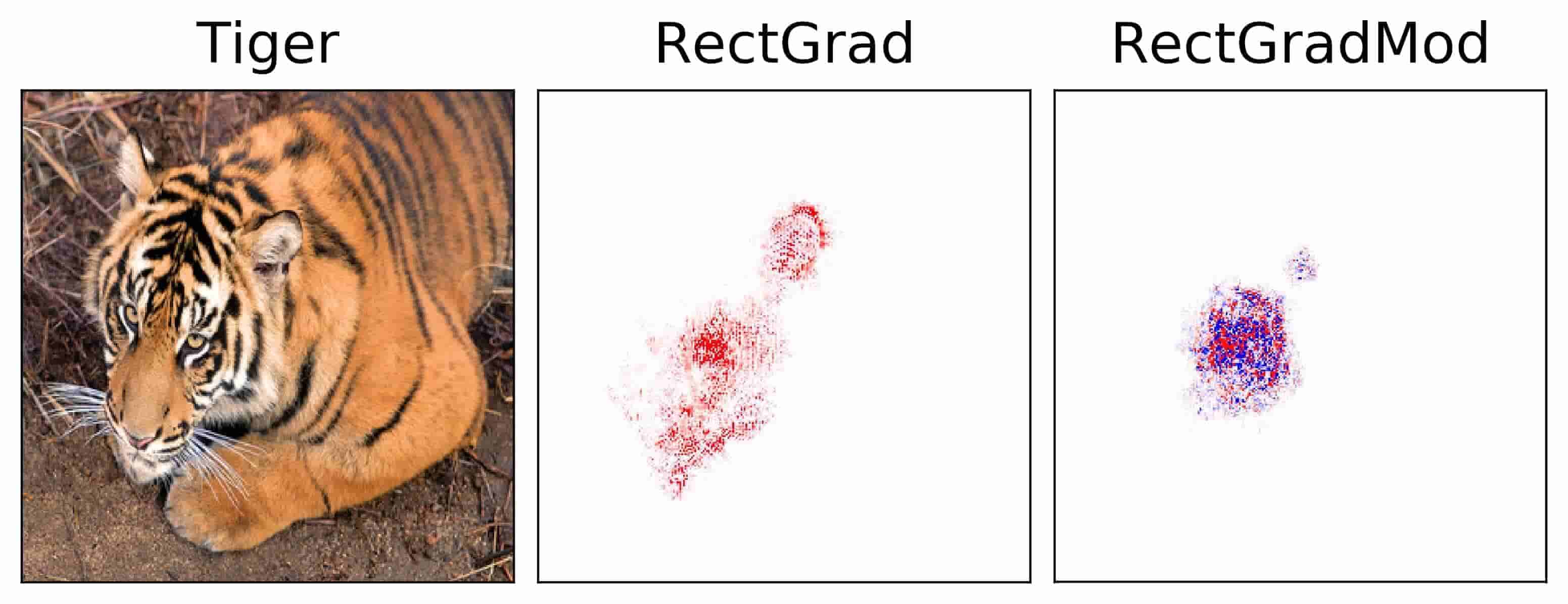}\hfill
    \includegraphics[width=0.33\linewidth]{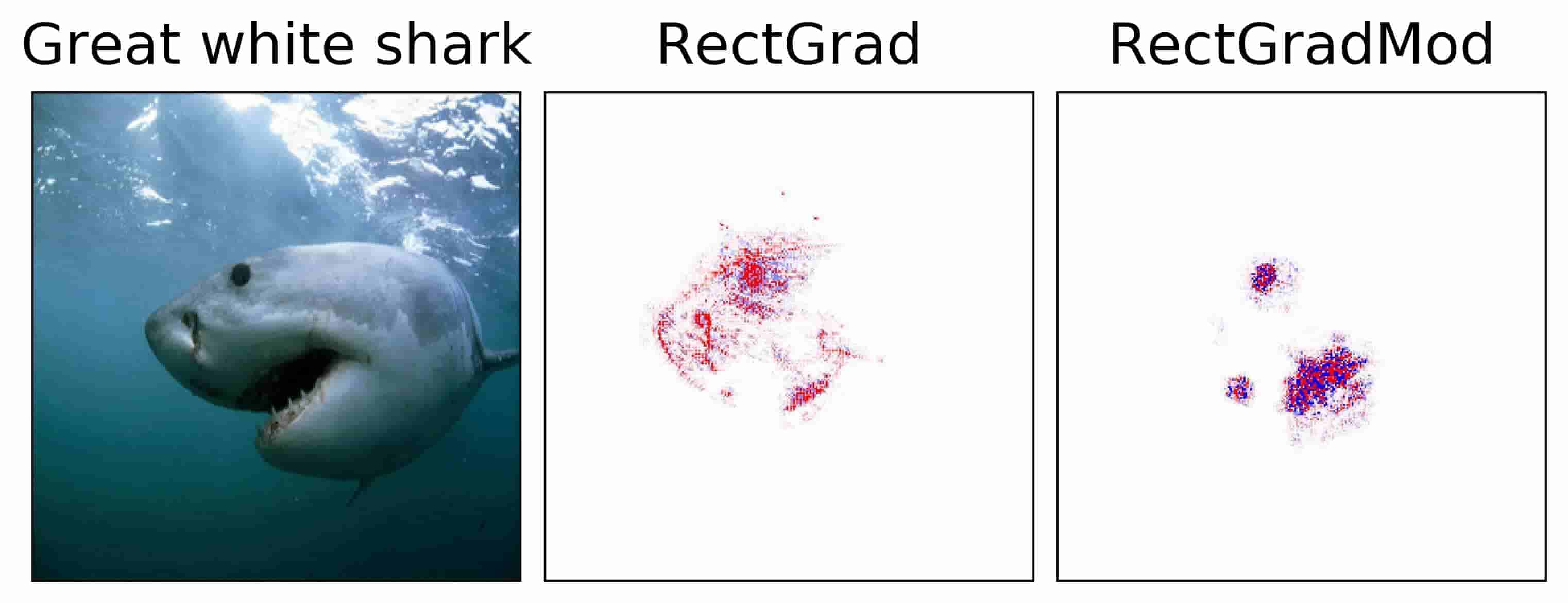}
    \par\medskip
    \includegraphics[width=0.33\linewidth]{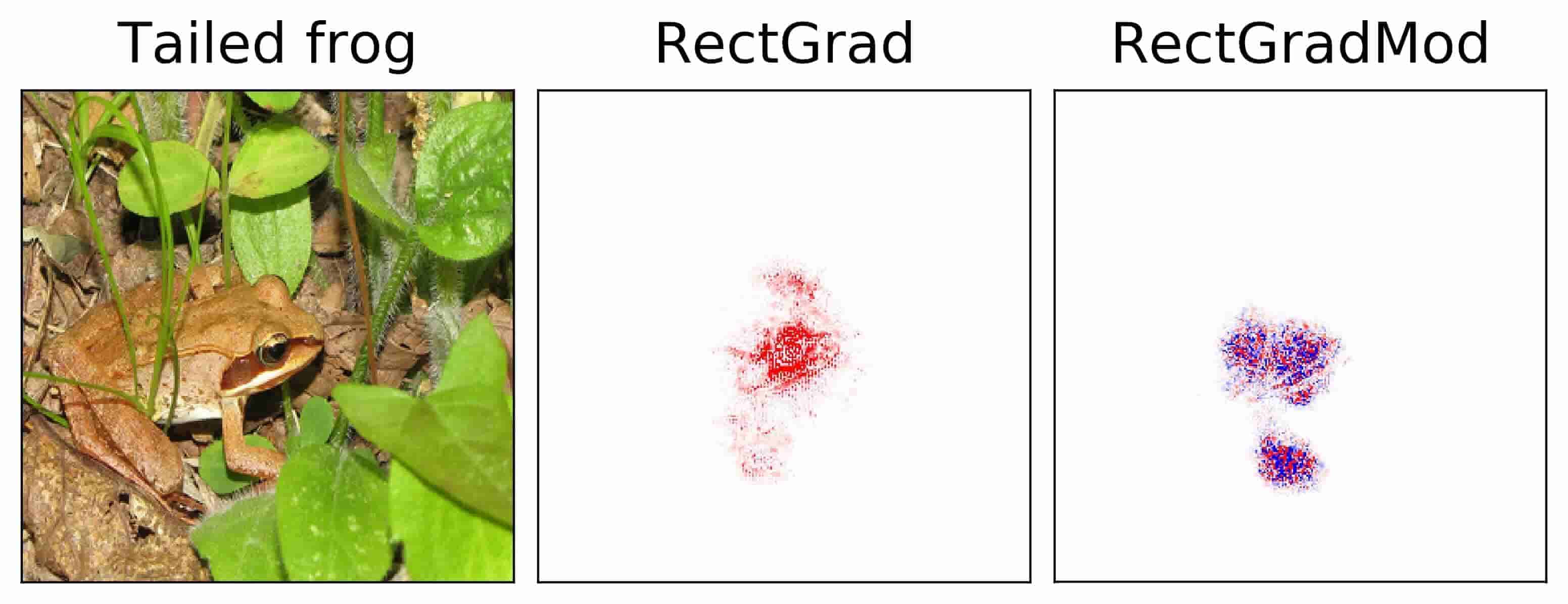}\hfill
    \includegraphics[width=0.33\linewidth]{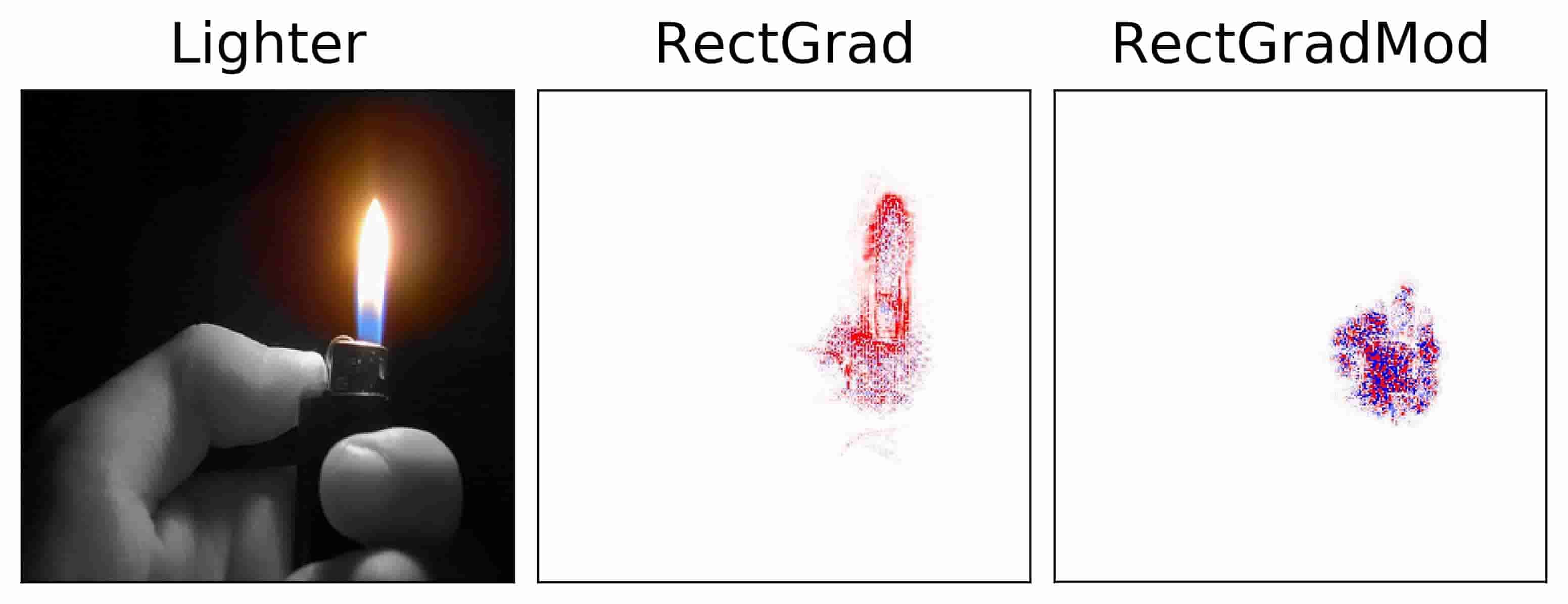}\hfill
    \includegraphics[width=0.33\linewidth]{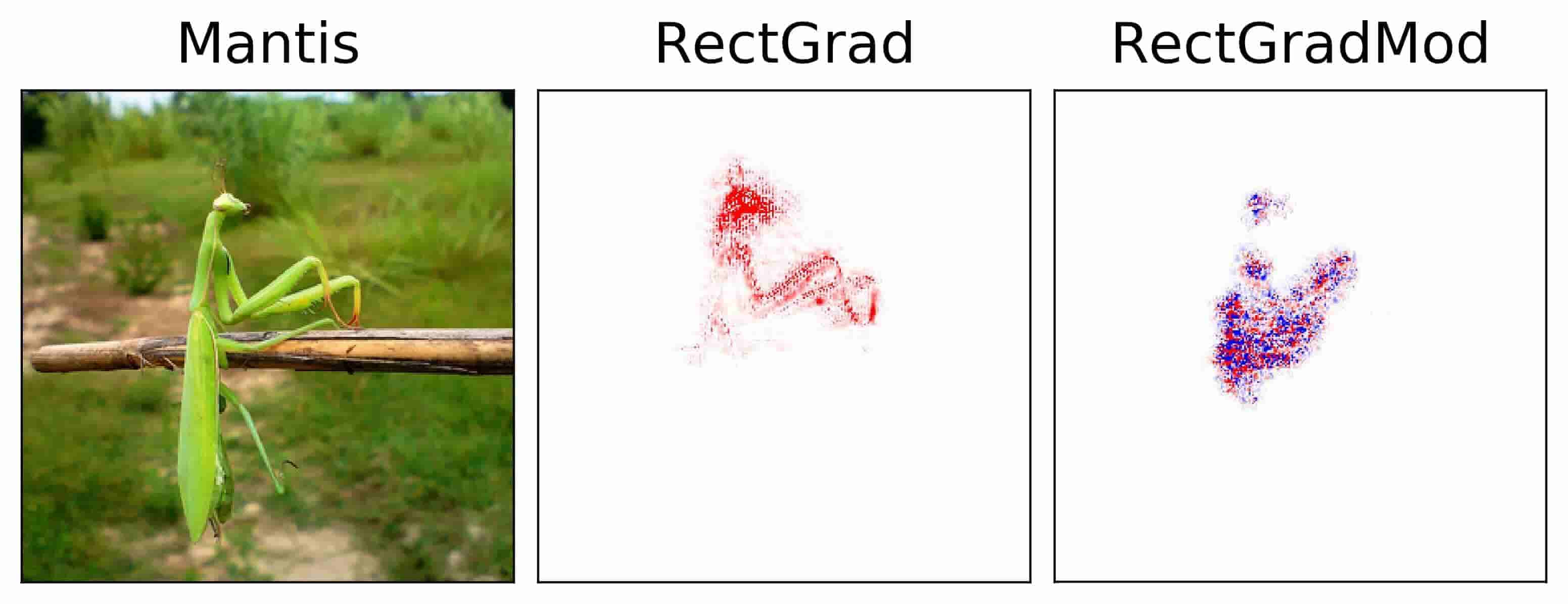}
    \caption{Comparison of samples generated by RectGrad and modified propagation rule.}
    \label{fig:mod}
\end{figure}

To corroborate our claim experimentally, we have generated several samples comparing PR1 (RectGrad) with the modified propagation rule which uses PR1 for the last logit layer and PR2 for other layers (RectGradMod). We show the results in Figure \ref{fig:mod}. We can observe that the modified propagation rule often fails to highlight discriminating features of the object of interest or highlights the background (e.g. \enquote{Carton} or \enquote{Soup bowl}) example. This corroborates our claim that since PR2 does not work for even the simplest examples, it is highly likely that this will not work for DNNs which are constructed by composing multiple affine layers.

\newpage

\section{Useful Techniques} \label{section:techniques}

Here, we present two useful techniques that can enhance the visual quality of attribution maps produced by RectGrad.

\subsection{Padding Trick}

Convolution inputs are typically zero padded along the border in order to preserve the spatial dimension of feature maps.\footnote{This corresponds to convolution with SAME padding in TensorFlow terminology.} This occasionally leads to high activation values along the border if zero is out of input distribution. Since importance scores are calculated by multiplying activation with gradient, outlying border activation can cause RectGrad to be propagated through the border instead of relevant features. To solve this problem, we masked the border of gradient to zero before the backward pass through convolutions with padding. One possible concern with the padding trick is that attributions may be faint for features adjacent to the border of the image. However, we did not find this to be a significantly problem experimentally. Listing \ref{lst:padding trick} in Appendix \ref{section:code2} shows how to implement the padding trick in TensorFlow.

\subsection{Proportional Redistribution Rule (PRR) for Pooling Layers.}

Attribution maps produced by RectGrad tend to be rough due to the discrete nature of thresholding. This discontinuity can be compensated by using the proportional redistribution rule proposed by \cite{Montavon2015} for the backward pass through max-pooling layers. Instead of propagating the gradient through only the most activated unit in the pool, gradient is redistributed proportional to unit activations. Since the redistribution operation is continuous, attribution maps generated with the proportional redistribution rule are smoother. Listing \ref{lst:prr} in Appendix \ref{section:code3} shows how to implement the proportional redistribution rule in TensorFlow.

\section{TensorFlow Codes} \label{section:codes}

\subsection{Implementation of Rectified Gradient} \label{section:code1}

\begin{lstlisting}[language=Python, caption={Implementation of Rectified Gradient in TensorFlow. After registering this function as the gradient for ReLU activation functions, call \texttt{tf.gradients()}, multiply with inputs, and threshold at $0$ to generate attributions.}, label={lst:rectified relu}]
import tensorflow as tf

from tensorflow.contrib.distributions import percentile

@tf.RegisterGradient("RectifiedRelu")
def _RectifiedReluGrad(op, grad):
    
    def threshold(x, q):
    
        if len(x.shape.as_list()) > 3:
            thresh = percentile(x, q, axis=[1,2,3], keep_dims=True)
        else:
            thresh = percentile(x, q, axis=1, keep_dims=True)
        
        return thresh
    
    activation_grad = op.outputs[0] * grad
    thresh = threshold(activation_grad, q)
    
    return tf.where(thresh < activation_grad, grad, tf.zeros_like(grad))
\end{lstlisting}

\newpage

\subsection{Implementation of the Padding Trick} \label{section:code2}

\begin{lstlisting}[language=Python, caption={Implementation of the padding trick in TensorFlow. Register this function as the gradient for convolution operations.}, label={lst:padding trick}]
import tensorflow as tf

@tf.RegisterGradient("RectifiedConv2D")
def _RectifiedConv2DGrad(op, grad):
    
    if op.get_attr('padding') == b'SAME':
        
        shape = tf.shape(grad)
        mask = tf.ones([shape[0], shape[1] - 2, shape[2] - 2, shape[3]])
        mask = tf.pad(mask, [[0,0],[1,1],[1,1],[0,0]])
        grad = grad * mask

    input_grad = tf.nn.conv2d_backprop_input(tf.shape(op.inputs[0]), op.inputs[1], grad, op.get_attr('strides'), op.get_attr('padding'))
    filter_grad = tf.nn.conv2d_backprop_filter(op.inputs[0], tf.shape(op.inputs[1]), grad, op.get_attr('strides'), op.get_attr('padding'))
    
    return input_grad, filter_grad
\end{lstlisting}

\subsection{Implementation of the Proportional Redistribution Rule} \label{section:code3}

\begin{lstlisting}[language=Python, caption={Implementation of the proportional redistribution rule in TensorFlow. Register this function as the gradient for max-pooling operations.}, label={lst:prr}]
import tensorflow as tf

from tensorflow.python.ops import gen_nn_ops

@tf.RegisterGradient("RectifiedMaxPool")
def _RectifiedMaxPoolGrad(op, grad):
    
    z = tf.nn.avg_pool(op.inputs[0], op.get_attr('ksize'), op.get_attr('strides'), op.get_attr('padding')) + 1e-10
    s = grad / z
    c = gen_nn_ops._avg_pool_grad(tf.shape(op.inputs[0]), s, op.get_attr('ksize'), op.get_attr('strides'), op.get_attr('padding'))
    
    return op.inputs[0] * c
\end{lstlisting}

\newpage

\section{Proofs of Claims} \label{section:proofs}

\subsection{Proof of Claim 1} \label{section:proof1}

\begin{proof}
Note that the backward propagation rule for Deconvolution through the ReLU nonlinearity is given by
\begin{equation}
R^{(l)}_i = \mathbb{I} \left(R^{(l + 1)}_i > 0 \right) \cdot R^{(l + 1)}_i.
\end{equation}
Since the DNN uses ReLU activation functions, $a^{(l)}_i + \epsilon > 0$ and therefore
\begin{equation} \label{eq:1}
\mathbb{I} \left[ \left( a^{(l)}_i + \epsilon \right) \cdot R^{(l + 1)}_i > 0 \right] = \mathbb{I} \left( R^{(l + 1)}_i > 0 \right)
\end{equation}
for all $l$ and $i$. The result follows from Equation \ref{eq:1}.
\end{proof}

\subsection{Proof of Claim 2} \label{section:proof2}

\begin{proof}
Note that the backward propagation rule for Guided Backpropagation through the ReLU nonlinearity is given by
\begin{equation}
R^{(l)}_i = \mathbb{I} \left(z^{(l)}_i > 0 \right) \cdot \mathbb{I} \left(R^{(l + 1)}_i > 0 \right) \cdot R^{(l + 1)}_i.
\end{equation}
Since the DNN uses ReLU activation functions, $a^{(l)}_i \geq 0$ and therefore
\begin{equation} \label{eq:2}
\mathbb{I} \left( a^{(l)}_i \cdot R^{(l + 1)}_i > 0 \right) = \mathbb{I} \left(z^{(l)}_i > 0 \right) \cdot \mathbb{I} \left(R^{(l + 1)}_i > 0 \right)
\end{equation}
for all $l$ and $i$. The result follows from Equation \ref{eq:2}.
\end{proof}

\newpage

\section{Experiment Setup} \label{section:setup}

\subsection{Attribution Map Visualization} \label{section:visualization}

To visualize the attributions, we summed up the attributions along the color channel and then capped low outlying values to $0.5$\textsuperscript{th} percentile and high outlying values to $99.5$\textsuperscript{th} percentile for RGB images. We only capped outlying values for grayscale images.




\subsection{CIFAR-10}

The CIFAR-10 dataset \cite{cifar2009} was pre-processed to normalize the input images into range $[-1;1]$. We trained a CNN using ReLU activation functions with Adam for 20 epochs to achieve $74.6\%$ test accuracy. For the dataset occluded with the random patch, we used the same settings to achieve $73.1\%$ test accuracy.

\begin{center}
\begin{tabular}{|c|}
\hline
\textbf{CIFAR-10 CNN} \\
\hline
Conv 2D ($3 \times 3$, $32$ kernels) \\
\hline
Conv 2D ($3 \times 3$, $32$ kernels) \\
\hline
Max-pooling ($2 \times 2$) \\
\hline
Conv 2D ($3 \times 3$, $64$ kernels) \\
\hline
Conv 2D ($3 \times 3$, $64$ kernels) \\
\hline
Max-pooling ($2 \times 2$) \\
\hline
Dense (256) \\
\hline
Dense (10) \\
\hline
\end{tabular}
\end{center}

\subsection{Inception V4}

We used a pre-trained Inception V4 network. The details of this architecture can be found in \cite{Szegedy2016}. For the adversarial attack, we used the fast gradient sign method with $\epsilon = 0.01$.

\newpage

\section{ROAR and KAR Curves}

\label{section:roar_kar_curves}

\begin{figure}[h]
	\centering
	\includegraphics[width=0.9\linewidth]{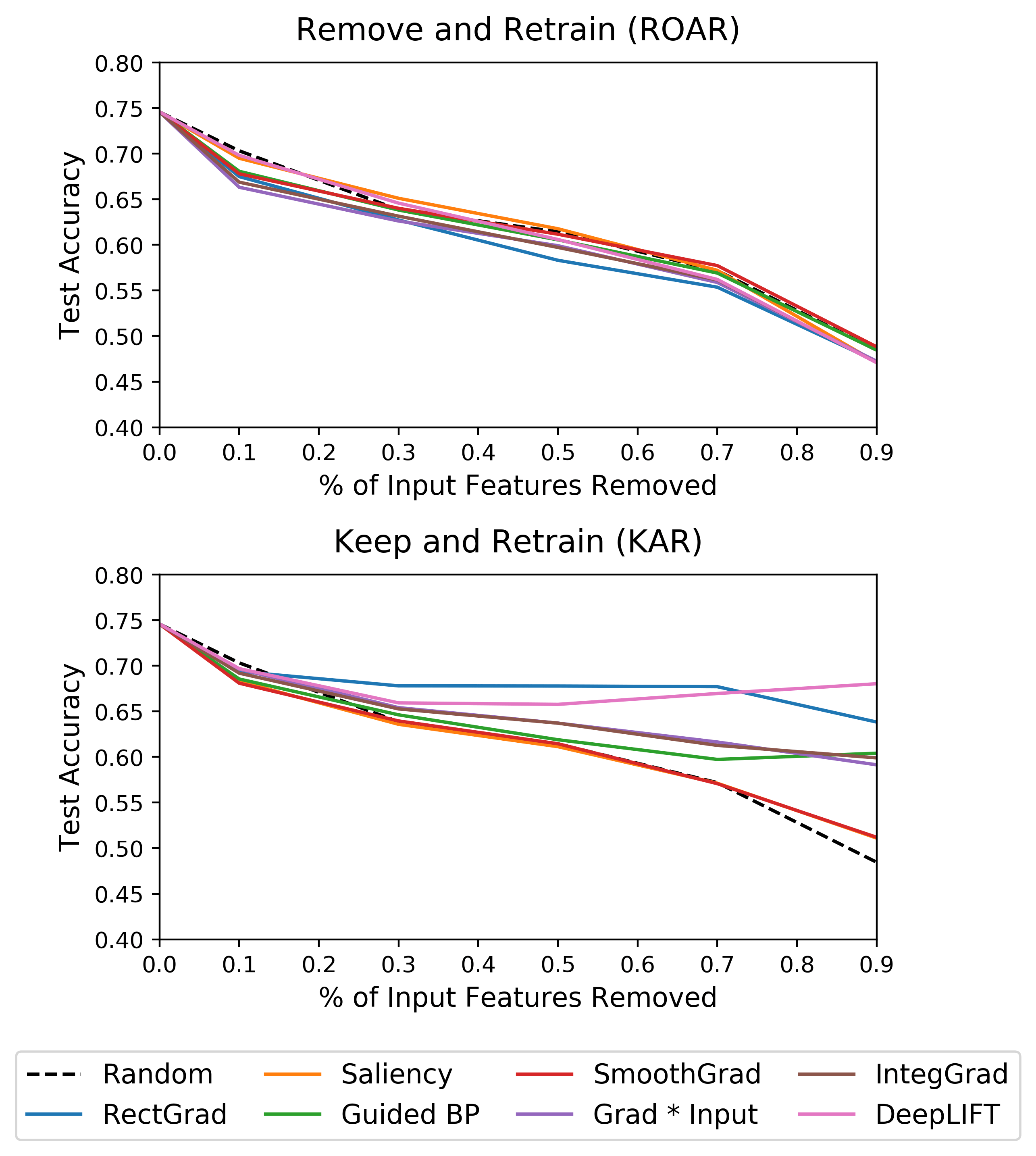}
    \caption{Comparison of ROAR and KAR curves. We include the random baseline (pixels are randomly removed) for reference.}
    \label{fig:roar kar}
\end{figure}

\end{document}